\documentclass{article} 
\usepackage{iclr2027_conference,times}
\usepackage{xcolor}
\usepackage[most]{tcolorbox}
\usepackage{titletoc}
\definecolor{RoboGreen}{HTML}{849D6A}   
\definecolor{revise}{HTML}{000000}
\usepackage{makecell}
\usepackage{colortbl}
\definecolor{RoboDark}{HTML}{4A544A}    
\definecolor{RoboGold}{HTML}{FFD96A}    
\definecolor{RoboOrange}{HTML}{FF974C}  
\definecolor{RoboBurnt}{HTML}{D74E09}   
\definecolor{RoboPaper}{HTML}{FCFAF1}   
\definecolor{RoboGate}{HTML}{FFC0AC}
\usepackage{multirow}
\newcommand{\FrameVLM}{\textcolor{RoboOrange}{\textbf{Frame-VLM}}}
\newcommand{\GateVLM}{\textcolor{RoboGreen}{\textbf{Gate-VLM}}}
\newcommand{\GateRule}{\textcolor{RoboGate}{\textbf{Gate-Rule}}}
\usepackage{booktabs}
\usepackage{tabularx}
\usepackage{booktabs}
\usepackage{longtable}
\usepackage{array}
\usepackage{ragged2e}
\usepackage{makecell}
\usepackage{booktabs}
\usepackage{xltabular}
\usepackage{array}
\usepackage{wrapfig}
\newcolumntype{L}[1]{>{\raggedright\arraybackslash}p{#1}}
\usepackage{graphicx}

\newcommand{\afflogo}[1]{%
  \raisebox{-0.25em}{%
    \includegraphics[height=1.2em]{#1}%
  }\hspace{0.3em}%
}

\newcommand{\coloredappendixletter}[1]{%
    \textcolor{RoboDark}{#1}%
}

\titlecontents{section}
  [0em]
  {\addvspace{0.10em}\large\bfseries\color{RoboDark}}
  {%
    \coloredappendixletter{\thecontentslabel}%
    \hspace{0.5em}%
  }
  {}
  {%
    \color{RoboDark}%
    \titlerule*[0.55pc]{.}\contentspage
  }

\titlecontents{subsection}
  [5em]
  {\small\color{RoboGreen}}
  {%
    \contentslabel{3.2em}%
  }
  {}
  {%
    \color{RoboDark}%
    \titlerule*[0.55pc]{.}\contentspage
  }

\newtcolorbox{expquestion}[2]{
    enhanced,
    breakable,
    arc=1.8mm,
    outer arc=1.8mm,
    boxrule=0.45pt,
    leftrule=1.8pt,
    colframe=#1,
    colback=RoboPaper,
    colbacktitle=#1!48!white,
    coltitle=RoboDark,
    title={#2},
    fonttitle=\bfseries,
    fontupper=\normalsize,
    left=5pt,
    right=5pt,
    top=3pt,
    bottom=3pt,
    toptitle=2.4pt,
    bottomtitle=2.4pt,
    titlerule=0pt,
    before skip=3.5pt,
    after skip=7pt,
}

\newtcolorbox{expfinding}[2]{
    enhanced,
    arc=1.6mm,
    outer arc=1.6mm,
    boxrule=0.35pt,
    leftrule=2.0pt,
    colframe=#1,
    colback=#1!5!white,
    colbacktitle=#1!22!white,
    coltitle=RoboDark,
    title={#2},
    fonttitle=\bfseries,
    fontupper=\normalsize,
    left=5pt,
    right=5pt,
    top=2.5pt,
    bottom=2.5pt,
    toptitle=2pt,
    bottomtitle=2pt,
    titlerule=0pt,
    before skip=3pt,
    after skip=7pt,
}

\usepackage{amsmath,amsfonts,bm}

\def\eqref#1{equation~\ref{#1}}

\def\1{\bm{1}}

\DeclareMathAlphabet{\mathsfit}{\encodingdefault}{\sfdefault}{m}{sl}
\SetMathAlphabet{\mathsfit}{bold}{\encodingdefault}{\sfdefault}{bx}{n}

\usepackage{hyperref}
\usepackage{url}
\usepackage{booktabs,array,tabularx,makecell,pifont}
\hypersetup{hidelinks}

\newcommand{\xmark}{\ding{55}}

\newcolumntype{L}[1]{>{\raggedright\arraybackslash}p{#1}}
\newcolumntype{Y}{>{\raggedright\arraybackslash}X}
\usepackage{xcolor}
\newcolumntype{C}{>{\centering\arraybackslash}X}
\definecolor{RoboLink}{HTML}{FF974C}

\hypersetup{
    colorlinks=true,
    citecolor=RoboLink,
    linkcolor=RoboLink,
    urlcolor=RoboLink
}

\title{RoboStep: Robot Rewards Benefit from Tracking Task Progress Step by Step}

\author{\normalfont Shujie Zhang$^{2,3}$\thanks{Equal contribution.
Author order decided via coin flip.}\hspace{0.6em}
Jingkun Yi$^{1,2,*}$\hspace{0.6em}
Zirui Zhou$^{4}$\hspace{0.6em}
Xudong Xu$^{2}$\hspace{0.6em}
Kailin Li$^{2}$ \\
\normalfont
Xing Shen$^{2}$\hspace{0.6em}
Xiaokang Yang$^{3}$\hspace{0.6em}
Yong Liu$^{1}$\hspace{0.6em}
Jiangning Zhang$^{1}$ \\[0.5em]
{\small
\afflogo{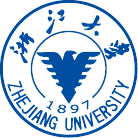}$^{1}$Zhejiang University
\hspace{0.8em}
\afflogo{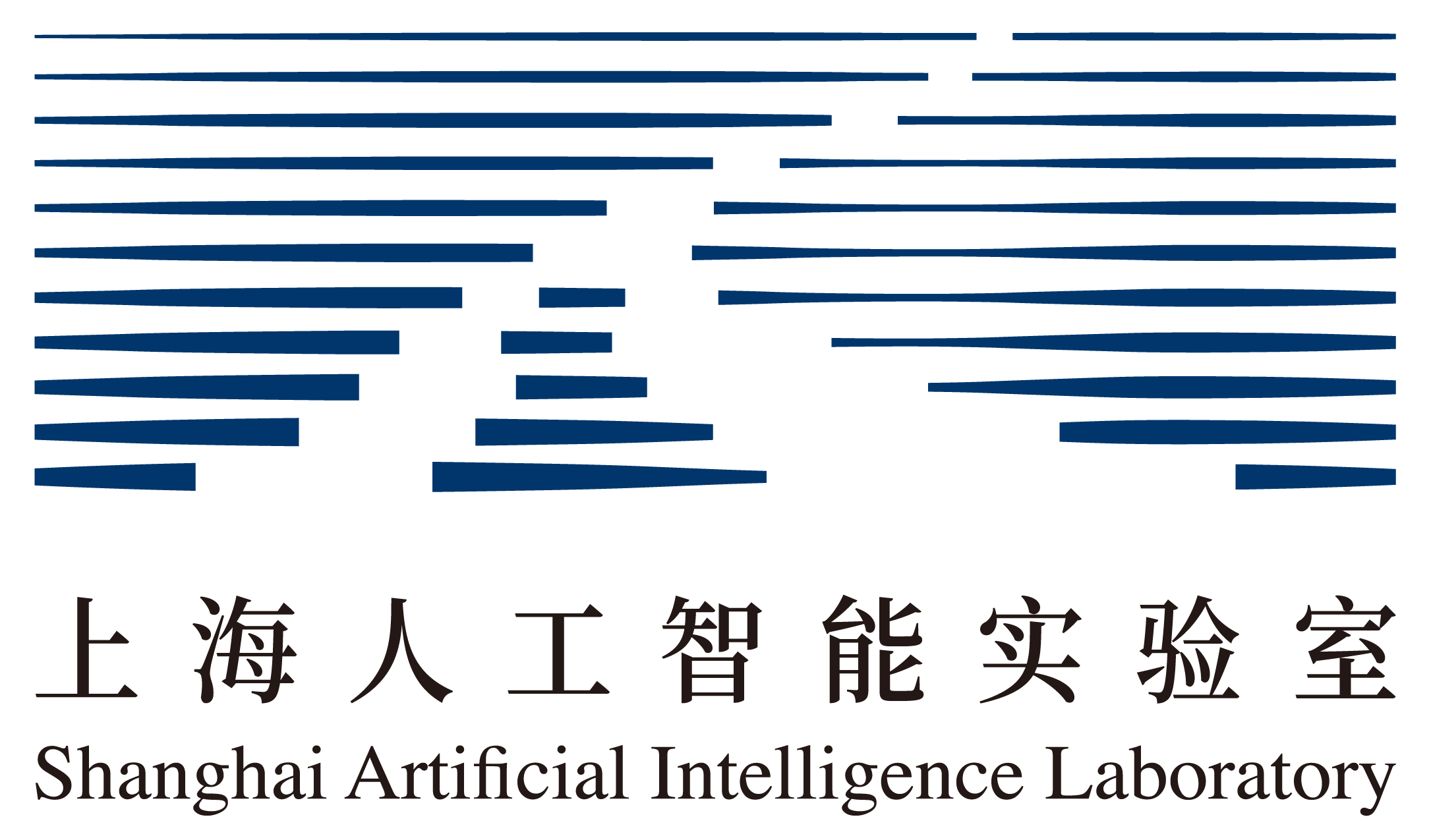}$^{2}$Shanghai Artificial Intelligence Laboratory
} \\[0.3em]
{\small
\afflogo{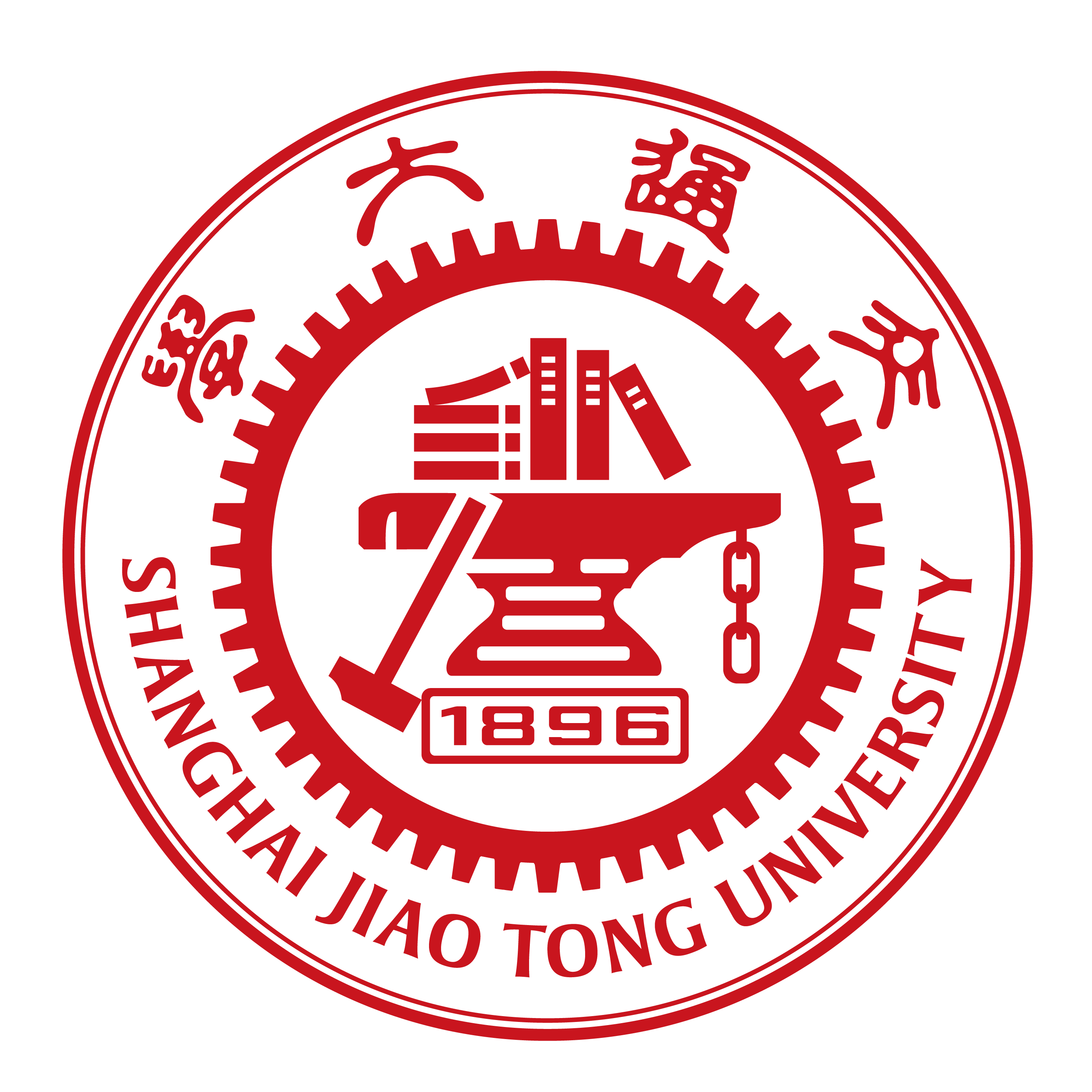}$^{3}$Shanghai Jiao Tong University
\hspace{0.8em}
\afflogo{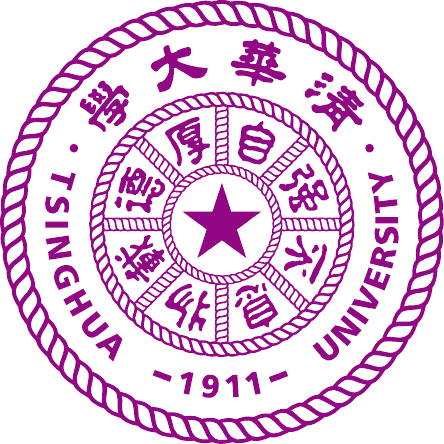}$^{4}$Tsinghua University
}
}

\usepackage{xspace}
\newcommand{\method}{RoboStep\xspace}
\iclrfinalcopy 
\begin{document}

\maketitle
\pagestyle{plain}
\thispagestyle{plain}

\begin{figure*}[htbp]
    \centering
    \includegraphics[width=\linewidth]{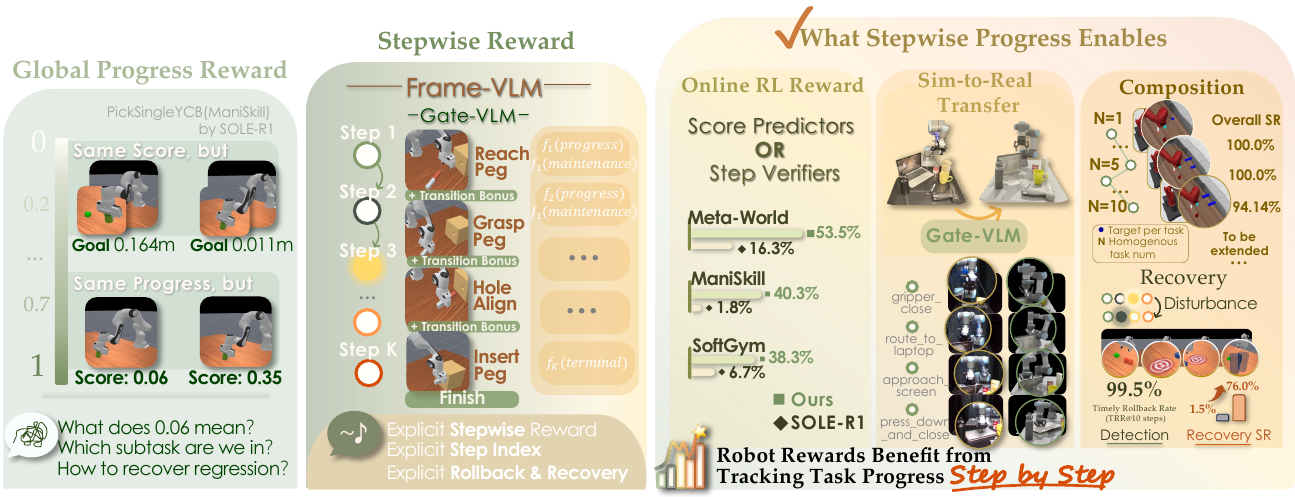}
    \vspace{-1.5em}
  \caption{
\textbf{RoboStep tracks task progress step by step instead of predicting a single global visual progress score.}
The persistent step state determines which reward is active, while Gate-VLM verifies semantic transitions, enabling explicit reward routing, composition, and efficient rollback after regressions.
}
    \label{fig:teaser}
\end{figure*}

\begin{abstract} Progress reward models aim to infer task progress from visual observations, especially when privileged task states are less accessible in the real world than in simulation. Yet progress is inherently task-specific and history-dependent, making a globally normalized score difficult to calibrate. Scalable supervision therefore often relies on frame positions in successful demonstrations, implicitly coupling progress with elapsed time and providing weak supervision for stagnation, regression, and recovery. In this work, we present RoboStep, a framework for constructing online robot rewards that explicitly tracks task progress step by step, using pretrained VLM semantics for step completion verification rather than global progress calibration. We use a \textit{Frame-VLM} to decompose tasks and generate step-local dense reward functions, while a \textit{Gate-VLM} decides when the reward advances to the next step and switches the active reward function. Across extensive simulated manipulation experiments, we systematically compare these two roles of VLMs under matched policy-learning settings. Combined with recent pipelines that reconstruct real-world scenes in simulation, RoboStep further enables policies trained with structured rewards in simulation to transfer back to corresponding real scenes. Overall, we show that robot learning benefits from tracking task progress step by step: for online reward, VLMs are better used as step-completion verifiers than as global scalar progress predictors. 
\end{abstract}

\section{Introduction}

Task progress provides an important signal for robot learning, yet privileged states needed to characterize semantic progress, like poses and distances, are often difficult to access in the real world. Therefore, visual reward models aim to infer task progress
directly from images or videos at
inference~\citep{liv2023,roboreward2026,soler12026} for downstream uses
such as policy learning and data annotation. However, a scalar score does not explicitly reveal which subgoals are completed, pending or invalidated, and dense human progress annotations are inevitably subjective and costly. Generally, a globally normalized task progress score is difficult to calibrate~\citep{progresssurvey2026}. Scalable alternatives often use normalized timestamps from successful demonstrations as  supervision~\citep{rlinfvla2025,rewind2025,zhanggeneralist,robometer2026,wvm2026,roboreward2026}, providing only weak supervision for stagnation and regression~\citep{robormbench2026,ji2026prmjudge,liu2026prmjudge15}.

\textcolor{revise}{Could semantic task steps provide progress signals easier to access for policy learning?} A related \textcolor{revise}{inspiration} from language-model reasoning is that supervision on intermediate
reasoning steps can be substantially more informative than final-outcome supervision
alone~\citep{lightman2023lets}. In robotic visual reward modeling,
several methods \textcolor{revise}{already incorporate} step awareness to improve semantic
grounding by inferring the current task stage from visual
observations~\citep{sarm2025,procvlm2026,stdr2026,sarm22026}. With task-specific expert demonstrations or boundary
annotations, \textcolor{revise}{these work efficiently inferred step and within-step progress} and recombined them into a global task-level scalar for downstream policy or data weighting.

\textcolor{revise}{With access to privileged simulator state,} programmatic and structured reward design can derive stepwise information directly from task and environment descriptions~\citep{text2reward2024,eureka2024}. These approaches range from one-shot subtask-completion bonuses to executable reward machines~\citep{rewardmachines2018,codeasreward2024,prm4rl2025,armfm2026}.
However, step transitions that fully rely on programmatic simulator states are difficult to access during real-world deployment. Besides, it can be difficult to express subgoal completion as a rule \textcolor{revise}{over particle positions even in simulation}, such as deciding whether a cloth fold or a fluid transfer is complete.

Pretrained VLMs have been widely used for task decomposition and extracting task-completion signals~\citep{moka2024,rekep2025,topreward2026}. \textit{Can VLMs use their native semantic abilities to provide step-level signals \textcolor{revise}{for online robot rewards and sim-to-real transfer}?}

To explore this question, in this work, we present \method{}, a framework for constructing online robot
rewards through persistent, stepwise tracking of task progress.
\method{} uses a \FrameVLM{} to decompose a task and generate step-local dense
reward functions from the task description and environment states. During interaction, a persistent step state determines which step-local rewards
are active, while a \GateVLM{} verifies the step completion and authorizes transitions. Upon a positive
verification, \method{} updates the active reward functions, exposes the updated
step index to the policy and issues a one-time transition bonus.
By representing task progression as an explicit reward state machine,
\method{} makes completion and regression auditable and provides a structured
interface for rollback and recovery when previously satisfied conditions are
invalidated.

Across controlled manipulation experiments, we show that robot learning benefits
from tracking task progress step by step. In \method{} framework, VLMs \textcolor{revise}{could serve as} step-completion verifiers for online rewards \textcolor{revise}{under from-scratch PPO protocol}. Combined with recent progress in real-to-sim scene
reconstruction~\citep{rialto2024,re3sim2026,robosnap2026,rola}, \GateVLM{} further
helps simulation-trained visual policies to transfer back to corresponding
real scenes by \textcolor{revise}{verifying the step index during real deployment}.

To summarize, our contributions are as follows:
\begin{itemize}

    \item We introduce \method{}, an online reward framework that constructs robot
rewards through stepwise tracking of task progress.

    \item Across controlled manipulation experiments, we demonstrate that VLMs \textcolor{revise}{can also serve} as step-completion verifiers for online rewards \textcolor{revise}{when training policies from scratch with PPO in simulation}.

    \item With real-to-sim scene reconstruction, we enable visual policies trained in simulation to transfer back to corresponding real scenes with \GateVLM{}-\textcolor{revise}{maintained step index}.

    \item We systematically study key dimensions of stepwise rewards including reward routing, policy step context, transition verification and
regression/recovery, with corresponding observations that may inform future
reward design.

\end{itemize}

\section{Related Work}

\paragraph{Visual Progress Rewards for Robot Learning.} Visual reward models estimate task-conditioned progress from visual observations for policy learning or data selection~\citep{vip2023,roboclip2023,vlmrm2024,decisionnce2024,rlvlmf2024,furl2024,gvl2025}. LIV~\citep{liv2023} learns a goal-conditioned value from temporally ordered videos, while ReWiND~\citep{rewind2025} augments temporal supervision with rewound trajectories and adds a one-time bonus from an external binary success signal. Large Reward Models~\citep{lrm2026} supports prediction of relative progress, discretized absolute progress and completion, whereas World Value Models~\citep{wvm2026} predict distributional value chunks for advantage-weighted regression or filtered behavior cloning. RoboReward~\citep{roboreward2026} and SOLE-R1~\citep{soler12026} predict episode-level and per-timestep progress respectively.
At the token level, RoboMeter~\citep{robometer2026} introduces learnable
progress and preference tokens for VLMs to learn, while TOPReward~\citep{topreward2026} directly
reads completion probabilities from frozen VLM output tokens. Stage-aware methods further introduce task structure into progress estimation
~\citep{sarm2025,procvlm2026,stdr2026,sarm22026,marvl2026,robodopamine2025,robodopamine22026,vllr2026}, typically predicting the current stage
and within-stage progress from visual observations before finally converting them into a task-level progress signal. Despite these differences, many scalable approaches still rely on temporal ordering from successful trajectories, while some further incorporate synthesized reversals, observed failure trajectories, or labor-intensive human labels. 

Our goal is to reconsider the role of visual progress models in online reward learning. Rather than repeatedly estimating a globally calibrated progress value, \method{} maintains the current step as persistent reward state, routes step-local dense rewards accordingly, and uses the visual reasoning capabilities of VLMs only to verify semantic transitions. This turns global progress prediction into a sequence of stepwise verification decisions.

\paragraph{Programmatic and Structured Reward Design.}
A complementary line of work directly encodes task structure into reward design~\citep{
language2reward2023}.
Reward Machines~\citep{rewardmachines2018} maintain persistent task state with a persistent automaton, while other works generate executable rewards with language models, both for providing reward~\citep{text2reward2024,eureka2024} and \textcolor{revise}{better sim-to-real transfer~\citep{dreureka2024}}. Code-as-Reward~\citep{codeasreward2024} further decomposes tasks with VLMs and persistently records subtask completion through one-shot auxiliary bonuses, and PRM4RL~\citep{prm4rl2025} generates subtask structure but scalarizes local progress into a task-level potential. 

ARM-FM~\citep{armfm2026} is particularly related with our method, using foundation models to generate explicit reward machines. Its manipulation transitions, however, rely on simulator-accessible predicates and require manual reward adjustment.
\method{} not only systematically studies how step state routes dense rewards,
conditions the policy, and handles transition errors and regression, but also further leverages pretrained VLM semantics to verify step transitions directly from visual observations. A detailed method comparison and task coverage is provided in Appendix~\ref{app:full-related-work}.


\paragraph{Real-to-Sim-to-Real Robot Learning.}
Real-world deployment often lacks simulator-side privileged quantities and may
require additional sensing or estimation to deploy simulation-trained
policies~\citep{phys2real2026}. Real-to-Sim-to-Real methods~\citep{rialto2024,jain2025polarisscalablerealtosimevaluations,re3sim2026,robosnap2026,rola,ranawaka2026simfoundry,twinrl2026,scalingsim2real2026} can be used to reconstruct target scenes in simulation for privileged-state training before transferring policies back to the real world.
Building on recent real-to-sim reconstruction advances~\citep{robosnap2026},
\method{} trains policies with stepwise rewards in simulation and transfers them
back to the corresponding real scene, where \GateVLM{} can still maintain the step state information from visual observations.
\section{Method}
\label{sec:method}

\paragraph{Overview.}
\method{} represents task progress as a persistent sequence of semantic steps.
Instead of predicting a globally calibrated scalar progress value, \method{}
maintains an explicit step index vector that determines which reward component should
be active. A \FrameVLM{} compiles the task description and available
environment interface into step-local dense reward functions and descriptions. During interaction, a \GateVLM{} verifies whether the current step has been completed from visual observations and authorizes transitions.
Upon an accepted transition, \method{} uses the updated step index to activate the corresponding reward function, to issue a one-time transition bonus and as the policy inputs. Full inplementation of \method{} is in Appendix~\ref{app:implementation}.

\begin{figure}[htbp]
    \centering
    \includegraphics[width=\linewidth]{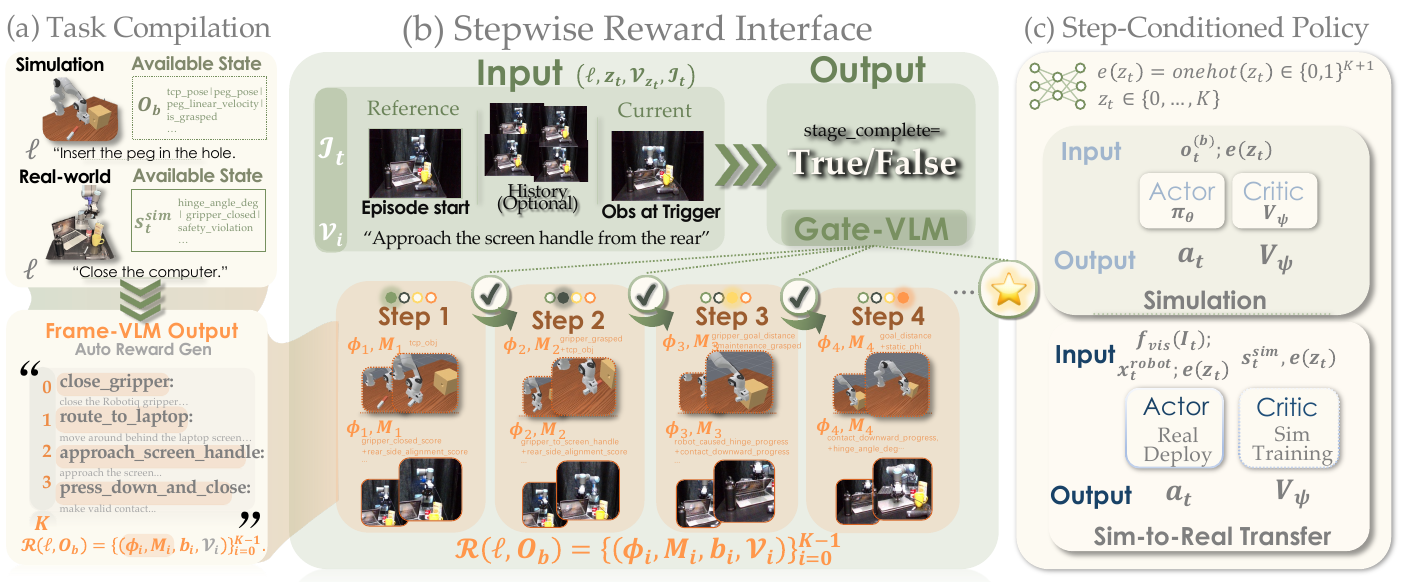}
    \vspace{-1.5em}
    \caption{
    \textbf{Overview of \method{}.}
    \FrameVLM{} compiles a task-conditioned stepwise reward program, while \GateVLM{} verifies semantic step transitions from RGB observations.
    The persistent step state routes step-local dense rewards and provides
    step context for policy learning.
    }
    \label{fig:method-overview}
\end{figure}

\subsection{Framework of Stepwise Task Compilation}
\label{sec:stepwise-reward}

\paragraph{Reward compilation.}
Let $\ell$ denote the task description, $O_b$ the observable-state interface
available in a simulation benchmark $b$, and
$o_t^{(b)}\in O_b$ its current observable state. The \FrameVLM{} compiles
the task into $K$ ordered non-terminal steps and generates the reward recipe
and invocation rule:
\begin{equation}
\mathcal{R}(\ell,O_b)
=
\left(
\left\{
(\phi_i,M_i,b_i,\mathcal{V}_i)
\right\}_{i=0}^{K-1},
\chi
\right).
\label{eq:reward-recipe}
\end{equation}
Here $\phi_i$ evaluates progress within step $i$ (e.g., approaching a target
pose). $M_i$ evaluates preserved conditions required for valid execution
(e.g., maintaining a grasp or stable contact). They together define the dense
reward sources for the active step. $b_i$ is a one-time transition bonus.
$\mathcal{V}_i$ is the visual completion standard description provided to
\GateVLM{} (e.g., whether the intended object is grasped). The invocation
rule $\chi$ produces the binary trigger $\chi_t\in\{0,1\}$ that determines
when semantic verification is requested. Instantiations of $\chi$ are
detailed in Appendix~\ref{app:formulation}. \FrameVLM{} does not use
benchmark-native rewards and success predicates to generate
$\mathcal{R}(\ell,O_b)$. The official success metrics are reserved only for
terminal success evaluation.

\paragraph{Persistent step state and routed reward.}
Let $z_t\in\{0,\ldots,K\}$ denote the persistent step state where $K$ is the terminal state. We define the activation indicator
\begin{equation}
A_i(t)=\mathbf{1}[z_t=i],
\qquad
\sum_{i=0}^{K-1}A_i(t)=1
\quad (z_t<K).
\label{eq:step-state}
\end{equation}
For each step, the local dense reward is computed as
\begin{equation}
D_i(o_t^{(b)},o_{t-1}^{(b)})
=
\Delta\Phi_i(o_t^{(b)},o_{t-1}^{(b)})
+
M_i(o_t^{(b)}),
\label{eq:dense-reward}
\end{equation}
\begin{equation}
\Delta\Phi_i(o_t^{(b)},o_{t-1}^{(b)})
=
s_i
C_i
\left(
\gamma_i\phi_i(o_t^{(b)})
-
\phi_i(o_{t-1}^{(b)})
\right)
\label{eq:potential-difference}
\end{equation}
is a bounded local progress change. Here $s_i$ is a step-specific scale
factor and $C_i$ is a clipping operator, both are fixed reward hyperparameters specified in the benchmark or task configuration.

Let $e_t=\mathbf{1}[\text{accepted step transition at }t]$ be a binary scalar. The stepwise reward is
\begin{equation}
r_t^{\mathrm{step}}
=
\sum_{i=0}^{K-1}
A_i(t)
D_i(o_t^{(b)},o_{t-1}^{(b)})
+
b_{z_t}e_t
+
P_t ,
\label{eq:stepwise-reward}
\end{equation}
where $P_t$ contains signed safety-related reward terms shared across steps. Once a transition is accepted, the previous step reward is deactivated and the next step reward becomes active. The transition bonus is paid only once at the transition boundary.

\subsection{\GateVLM{} Stepwise Reward Interface}
\label{sec:gate-verification}

\paragraph{Invocation and semantic verification.} The reward program provides a mechanism for deciding when the \GateVLM{} is invoked for verification. With $\chi_t\in\{0,1\}$ denoting the invocation
trigger, when $\chi_t=1$, the \GateVLM{} receives the task description $\ell$,
current step index $z_t$, visual completion description
$\mathcal{V}_{z_t}$, and RGB observations $\mathcal{I}_t$.
Its output is encoded as
\begin{equation}
v_t
=
\mathbf{1}
\left[
\GateVLM{}\!
\left(
\ell,z_t,\mathcal{V}_{z_t},\mathcal{I}_t
\right)
=
\texttt{success}
\right],
\label{eq:gate-decision}
\end{equation}
with $v_t=0$ when $\chi_t=0$. Here $\mathcal{V}_{z_t}$ is the
completion description generated by \FrameVLM{}, and the RGB input is
\begin{equation}
\mathcal{I}_t
=
\left[
I^{\mathrm{start}},
I_t^{\mathrm{history}},
I_t
\right],
\label{eq:gate-images}
\end{equation}
where $I^{\mathrm{start}}$ is the episode-start reference frame,
$I_t$ is the current frame captured at invocation, and
$I_t^{\mathrm{history}}$ is the optional ordered sequence of recent frames.
The history depends on the benchmark and is empty in the two-frame
implementation.

An accepted transition requires both invocation and successful semantic
verification:
\begin{equation}
e_t=\chi_t v_t,
\qquad
z_{t+1}=z_t+e_t.
\label{eq:transition}
\end{equation}

\subsection{Step-Conditioned Policy}
\label{sec:r2s2r}

Let $\mathbf{e}(z_t)$ denote the one-hot context of the persistent step state.
\begin{equation}
\mathbf{e}(z_t)=\operatorname{onehot}(z_t)
\in\{0,1\}^{K+1}
\label{eq:step-context}
\end{equation}
For the three
simulation benchmarks, the PPO actor--critic~\citep{ppo2017} receives the benchmark
observation together with this persistent step context:
\begin{equation}
a_t
\sim
\pi_\theta
\left(
\cdot
\mid
o_t^{(b)},\mathbf{e}(z_t)
\right),
\qquad
V_\psi
=
V_\psi
\left(
o_t^{(b)},\mathbf{e}(z_t)
\right).
\label{eq:policy-interface}
\end{equation}
For Real-to-Sim-to-Real transfer, we reconstruct the target real scene in simulation and train a deployment-compatible visual actor. The actor receives
\begin{equation}
x_t^{\mathrm{actor}}
=
\left[
f_{\mathrm{vis}}(I_t),
x_t^{\mathrm{robot}},
\mathbf{e}(z_t)
\right],
\end{equation}
where $f_{\mathrm{vis}}$ extracts visual features~\citep{dinov2_2023} and
$x_t^{\mathrm{robot}}$ contains robot proprioception.

The critic additionally receives privileged simulator information~\citep{asymmetricac2018}:
\begin{equation}
x_t^{\mathrm{critic}}
=
\left[
s_t^{\mathrm{sim}},
\mathbf{e}(z_t)
\right].
\end{equation}
At deployment, the actor relies on
deployment-compatible observations and the persistent step index provided through \GateVLM{} visual step verification.

\paragraph{Ablation Design.} We ablate stepwise structure, policy step context, \GateVLM{} invocation, and bidirectional progress (composition and regression). Detailed
formulations are in Appendix~\ref{app:additional-ablations}.
\section{Stepwise Rewards for Robot Learning}
\label{sec:experiments}

Within each benchmark, all reward methods use the same PPO architecture and evaluation protocol. \textit{\textbf{Unless noted otherwise}}, we evaluate policies trained with 3 independent seeds on 50 episodes each and report their
mean performance across task suites, and instantiate \FrameVLM{} with
\texttt{gpt-5.5-2026-04-23} using \texttt{xhigh} reasoning effort
~\citep{openai2026gpt55} and \GateVLM{} with \texttt{Qwen3.5-9B}~\citep{qwen35}.

\subsection{Stepwise Reward Structure}
\label{sec:exp-stepwise}

\begin{expquestion}{RoboGreen}{
    Q1. Does stepwise reward improve policy learning?
}
\phantomsection
\label{sec:q1}
Stepwise information benefits policy learning both as \textbf{reward structure}
and as \textbf{policy context}.
\end{expquestion}

To test the core premise of \method{} that
\textit{robot rewards benefit from tracking task progress step by step}, we evaluate a subset of tasks from Meta-World~\citep{metaworld2019},
ManiSkill~\citep{maniskill3}, and SoftGym~\citep{softgym2020}.
We independently vary two factors: whether the reward uses an explicit stepwise
structure and whether the actor and critic receive the current step index.

The \emph{stepwise} setting follows our main method in Sec.~\ref{sec:stepwise-reward}. In the \emph{all-dense} control, all step-local dense rewards are active simultaneously with the transition bonus removed, and the step index is used as an additional controlled policy input. All variants use the same \FrameVLM{}-generated reward components and task decomposition. 

\begin{table}[htbp]
  \centering
  \setlength{\belowcaptionskip}{2pt}
  \vspace{-0.8em}

  \caption{
  \textbf{Effect of reward-side stepwise structure and policy-side step-index input.}.
  }
  \label{tab:q1-structure}

  \small
  \setlength{\tabcolsep}{2.5pt}
  \renewcommand{\arraystretch}{0.82}

  \begin{tabularx}{\linewidth}{@{}lCCCcCCCc@{}}
  \toprule
  & \multicolumn{4}{c}{\textbf{w/ Step-Index Input}}
  & \multicolumn{4}{c}{\textbf{w/o Step-Index Input}} \\
  \cmidrule(lr){2-5}\cmidrule(lr){6-9}

  \textbf{Reward}
  & \mbox{Meta-World} & ManiSkill & SoftGym & Avg.
  & \mbox{Meta-World} & ManiSkill & SoftGym & Avg. \\
  \midrule

  Stepwise
  & \textbf{92.20}
  & \textbf{62.00}
  & \textbf{49.12}
  & \textbf{67.77}
  & 46.38
  & \textbf{44.40}
  & \textbf{49.83}
  & \textbf{46.87} \\

  All-dense
  & 62.85
  & 21.60
  & 44.00
  & 42.82
  & \textbf{52.98}
  & 20.00
  & 47.61
  & 40.19 \\

  \bottomrule
  \end{tabularx}
\end{table}

As shown in Table~\ref{tab:q1-structure}, both stepwise reward routing and
step-index policy context improve performance, and their combination provides even better improvement. Full task subsets, formulations and results are provided in Appendix~\ref{app:structure}.
\subsection{Roles of VLMs in Online Reward}
\label{sec:exp-verifier}

\begin{expquestion}{RoboGold}{
   Q2. For online rewards, should VLMs verify step completion or predict global progress?
}
\phantomsection
\label{sec:q2}
VLMs are more effective as step-completion verifiers than as global progress predictors
for online rewards.
\end{expquestion}


\hyperref[sec:q1]{Q1} establishes the benefit of stepwise rewards, but its
transitions rely on privileged \GateRule{} predicates that are unavailable in
the real world or difficult to specify for deformable and fluid tasks.
We therefore evaluate whether \GateVLM{} can serve as a practical
visual transition verifier.

\begin{table}[htbp]
\centering
\setlength{\belowcaptionskip}{2pt}
\vspace{-0.5em}
\caption{\textbf{Performance on VLM roles in online reward learning across 75 manipulation tasks.}
Bold indicates the best non-reference result in each column.}
\label{tab:q2-main}

\scriptsize
\setlength{\tabcolsep}{1.15pt}
\renewcommand{\arraystretch}{0.8}

\newcommand{\nativecell}[1]{%
  {\setlength{\fboxsep}{1pt}%
  \colorbox{RoboOrange!10}{%
    \makebox[\dimexpr\linewidth-1\fboxsep-0.1pt\relax][c]{\strut #1}%
  }}%
}

\begin{tabularx}{\linewidth}{@{}L{1.48in}*{10}{>{\centering\arraybackslash}X}@{}}
\toprule
& \multicolumn{4}{c}{\textbf{Benchmark Average}}
& \multicolumn{6}{c}{\textbf{Task-Type Average}} \\
\cmidrule(lr){2-5}
\cmidrule(lr){6-11}

\textbf{Reward Method}
& \makecell{\textbf{Meta-}\\\textbf{World}}
& \makecell{\textbf{Mani-}\\\textbf{Skill}}
& \textbf{SoftGym}
& \textbf{Avg.}
& \textbf{Rigid}
& \textbf{Artic.}
& \textbf{Precise}
& \textbf{Tool}
& \textbf{Deform.}
& \textbf{Fluid} \\
\midrule

\multicolumn{11}{@{}l}{\textit{Task-Solvability Reference}} \\[4pt]

\textbf{Bench-Native Reference}
& \nativecell{$\approx$81.8\%}
& \nativecell{31.0\%}
& \nativecell{57.7\%}
& \nativecell{67.0\%}
& \nativecell{77.7\%}
& \nativecell{82.1\%}
& \nativecell{42.5\%}
& \nativecell{52.4\%}
& \nativecell{50.0\%}
& \nativecell{73.0\%} \\

\midrule
\multicolumn{11}{@{}l}{\textit{Sparse Reward Controls}} \\[4pt]

\makecell[l]{\textbf{Terminal-only} Sparse reward}
& 7.5\% & 3.2\% & 37.2\% & 8.8\%
& 3.7\% & 11.9\% & 0.4\% & 12.0\%
& \textbf{35.2\%} & 41.1\% \\

\makecell[l]{\textbf{Per-subtask} Sparse reward}
& 16.5\% & 0.8\% & 7.3\% & 11.8\%
& 6.6\% & 24.9\% & 6.1\% & 1.9\%
& 11.0\% & 0.0\% \\

\midrule
\multicolumn{11}{@{}l}{\textit{Visual Progress Reward Baselines}} \\[4pt]

\textbf{SOLE-R1~\citep{soler12026}}
& 16.3\% & 1.6\% & 6.7\% & 11.8\%
& 6.7\% & 26.6\% & 3.9\% & 2.2\%
& 22.0\% & -23.9\% \\

\textbf{RoboMeter~\citep{robometer2026}}
& 5.5\% & 1.1\% & 6.1\% & 4.4\%
& 4.5\% & 8.6\% & 0.0\% & 0.0\%
& -0.2\% & 18.6\% \\

\textbf{RoboReward~\citep{roboreward2026}}
& 9.1\% & 5.9\% & 7.3\% & 8.1\%
& 4.5\% & 16.5\% & 0.9\% & 11.7\%
& 8.0\% & 6.0\% \\

\textbf{TOPReward~\citep{topreward2026}}
& 5.1\% & 0.7\% & 11.2\% & 4.5\%
& 1.2\% & 10.3\% & 0.2\% & 0.4\%
& 17.2\% & -0.7\% \\

\textbf{LRM-Progress~\citep{lrm2026}}
& 6.3\% & 1.8\% & 22.8\% & 6.5\%
& 1.4\% & 12.0\% & 0.6\% & 5.2\%
& 13.4\% & 41.6\% \\

\midrule
\multicolumn{11}{@{}l}{\textit{Stepwise Reward (Ours)}} \\[4pt]

\textbf{\method{}} (\GateRule{})
& \textbf{68.3\%} & \textbf{42.6\%}
& 31.2\% & \textbf{58.8\%}
& \textbf{72.1\%} & \textbf{77.1\%}
& \textbf{28.3\%} & \textbf{59.3\%}
& 27.0\% & 39.5\% \\

\textbf{\method{}} (\GateVLM{})
& 53.5\% & 40.3\%
& \textbf{38.3\%} & 49.0\%
& 56.6\% & 69.2\%
& 16.5\% & 51.4\%
& 31.5\% & \textbf{51.9\%} \\

\bottomrule
\end{tabularx}
\end{table}


\begin{figure}[t]
    \centering

    \includegraphics[width=\linewidth]{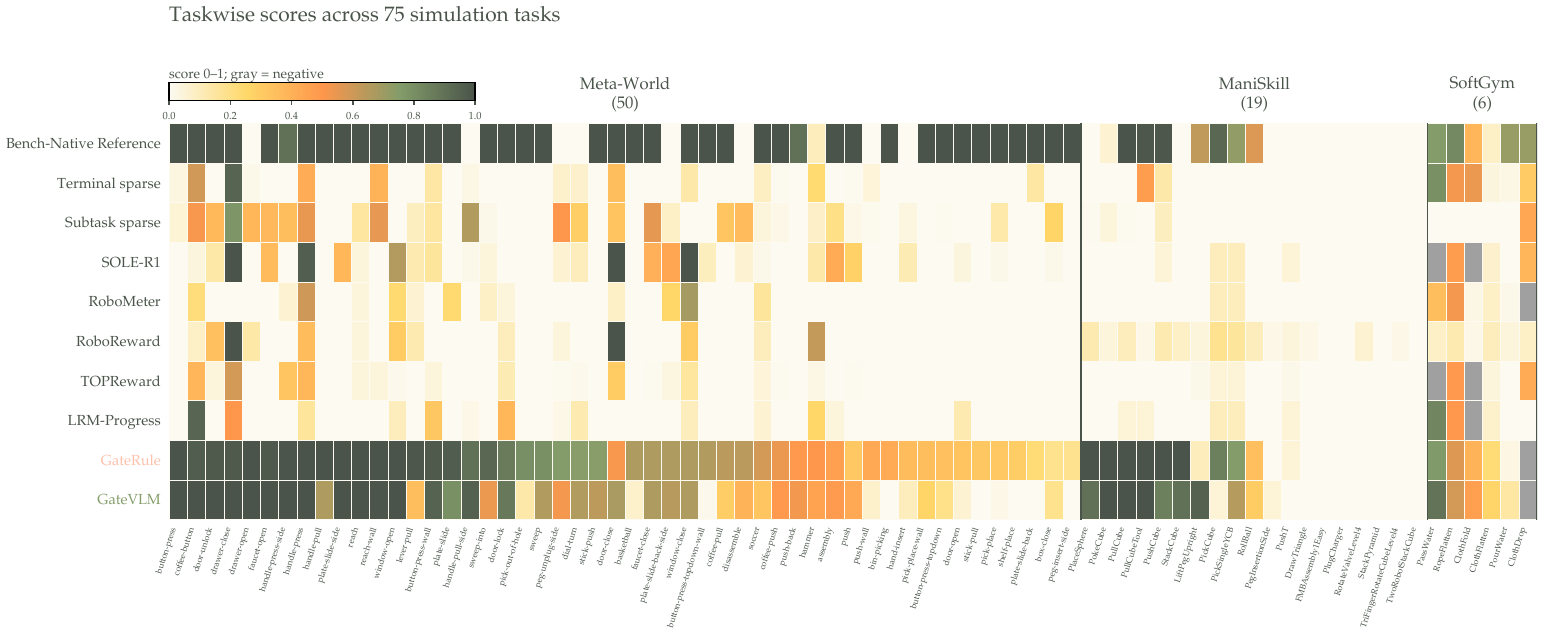}

    \vspace{-0.5em}
    \caption{
    \textbf{Per-task performance across 75 benchmark tasks. }
    }
    \label{fig:q2-task-heatmap}
\end{figure}

We train policies for 50 Meta-World, 19 ManiSkill and 6 SoftGym tasks and compare compatible VLM-based global progress methods under matched PPO settings. Results in Table~\ref{tab:q2-main} show that, across benchmarks and task categories, using VLMs as step-completion verifiers generally yields higher average success rates. Although \GateVLM{} generally trails the privileged \GateRule{} in simulation, policies trained with \GateVLM{} performs better on deformable and fluid tasks. This suggests that VLM verification may be particularly beneficial when subtask completion is difficult to express with distance- or pose-based rules.
Task types and definitions are provided in Appendix~\ref{app:task-definitions} with step decompositions in Table~\ref{tab:task_reward_spec}. Baseline implementation details are in Appendix~\ref{app:baselines}.

\subsection{Sim-to-Real Transfer}
\label{sec:exp-r2s2r}

\begin{expquestion}{RoboOrange}{
    Q3. Can stepwise rewards support sim-to-real transfer?
}
\phantomsection
\label{sec:q3}
With the persistent step index maintained by \GateVLM{} at deployment,
policies trained in reconstructed simulation can transfer back to the
corresponding real scenes.
\end{expquestion}

\hyperref[sec:q1]{Q1} and \hyperref[sec:q2]{Q2} show that the actor benefits
from explicit step-index context, which can be maintained by \GateVLM{}
through visual verification. This motivates maintaining the
policy step state from visual observations at
deployment. We reconstruct digital twins of the target real scenes using the
pipeline of prior work~\citep{robosnap2026}. During deployment, \GateVLM{} is
queried periodically to update the persistent step index supplied to the
actor. The \textit{w/o} control removes the step index only from the actor
during training and deployment.

\begin{figure}[htbp]
\centering

\begin{minipage}[c]{0.44\linewidth}
\centering

\includegraphics[width=\linewidth]{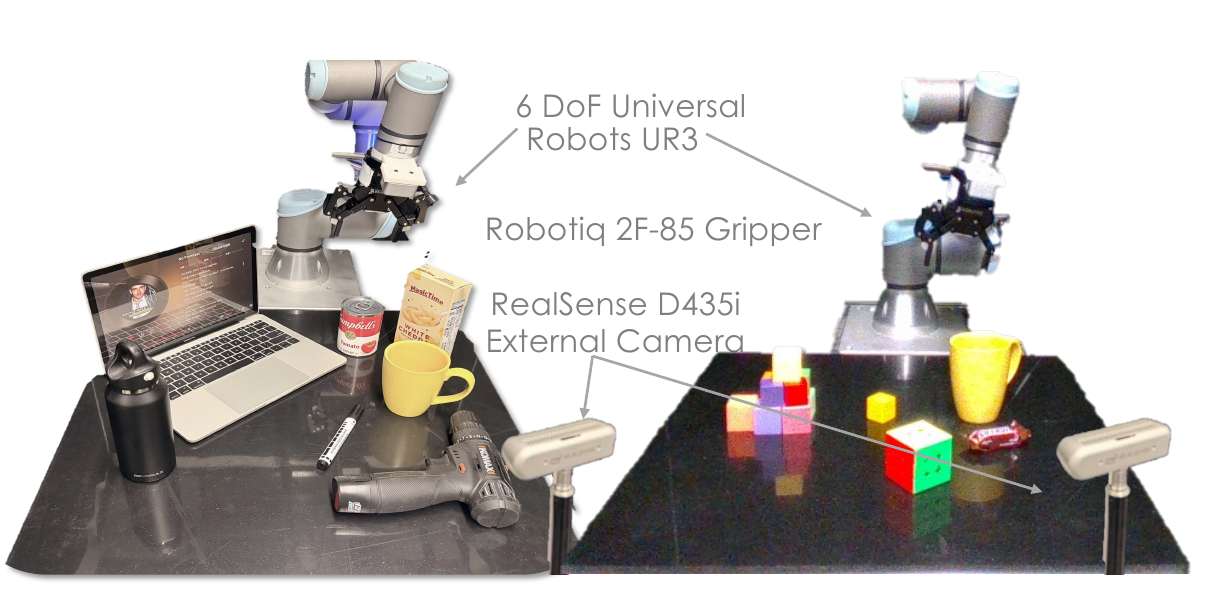}
\end{minipage}
\vspace{0.5em}
\hfill
\begin{minipage}[c]{0.54\linewidth}
\centering

\footnotesize
\setlength{\tabcolsep}{3pt}
\renewcommand{\arraystretch}{0.75}

\vspace{0.7em}
\begin{tabularx}{\linewidth}{
    @{}
    c
    c
    >{\raggedright\arraybackslash}X
    @{}
}
\toprule
\textbf{Task} & \textbf{Type} & \textbf{Task description} \\
\midrule

1.1 & Articulate
& \textit{Close the computer.} \\
\midrule

1.2 & Precise
& \textit{Place the pen in the cup.} \\
\midrule

2.1 & Rigid
& \textit{Put the cube on the right into the cup.} \\
\midrule

2.2 & Rigid
& \textit{Put the rubik near the cup.} \\

\bottomrule
\end{tabularx}

\end{minipage}

\vspace{-1.0em}
\caption{
\textbf{Real-world task setups for sim-to-real experiments.}
Detailed task definitions and success criteria are provided in Appendix~\ref{app:r2s2r-scenes}
}
\label{fig:real-tasks-transfer}
\end{figure}

\begin{table}[htbp]
\centering
\vspace{-0.3em}
\caption{
\textbf{Sim-to-real transfer success rates (\%).}
\textit{w/o} means \textbf{removing the step index only from the actor} in training and inference. Full statistics
are reported in Appendix Table~\ref{tab:q3-real-transfer}.
}
\label{tab:q3-transfer-summary}

\scriptsize
\setlength{\tabcolsep}{1.7pt}
\renewcommand{\arraystretch}{0.90}

\begin{tabularx}{\linewidth}{
@{}
c
*{4}{>{\centering\arraybackslash}X}
@{\hspace{4pt}}
c
*{4}{>{\centering\arraybackslash}X}
@{}
}
\toprule

\textbf{Task}
&
\makecell{\textbf{Sim. w/}}
&
\makecell{\textbf{Sim. w/o}}
&
\makecell{\textbf{Real w/}}
&
\makecell{\textbf{Real w/o}}
&
\textbf{Task}
&
\makecell{\textbf{Sim. w/}}
&
\makecell{\textbf{Sim. w/o}}
&
\makecell{\textbf{Real w/}}
&
\makecell{\textbf{Real w/o}}
\\
\midrule

1.1 & 45.0 & 33.0 & 33.3 & 26.7
&
2.1 & 36.0 & 25.0 & 23.3 & 16.7
\\

1.2 & 25.0 & 13.0 & 13.3 & 0.0
&
2.2 & 41.0 & 30.0 & 30.0 & 23.3
\\
\bottomrule
\end{tabularx}
\end{table}

Across the four tasks, the step-conditioned actor improves average success
from 25.3\% to 36.8\% in reconstructed simulation and from 16.7\% to
25.0\% in the real world, corresponding to gains of 11.5 and 8.3
percentage points, respectively. The precise insertion task remains the most
challenging, with the actor without step context failing in all real-world
trials. Both variants exhibit a clear sim-to-real drop, reflecting remaining
errors in perception, calibration, and physical dynamics. Nevertheless, the
consistent advantage across all four tasks suggests that stepwise reward
design and \GateVLM{}-maintained policy context can support direct transfer
without real-robot demonstrations or real-world policy updates. See
Figure~\ref{fig:method-overview} for the policy interfaces and
Appendix~\ref{app:r2s2r} for the detailed protocol.

\subsection{Bidirectional Task Progression}
\label{sec:exp-bidirectional}

\begin{expquestion}{RoboBurnt}{
    Q4. Can stepwise training support composition and recovery?
}
\phantomsection
\label{sec:q4}
Explicit step-index context support \textbf{task composition}
and \textbf{recovery from unseen regressions}.
\end{expquestion}
A valid task progress formulation should support the representation of both \textit{forward composition} and \textit{backward regression}.

\paragraph{Forward Composition.}
\label{par:exp-composition}

In our experiment, Homo$_n$ denotes $n$ \textbf{homogeneous} macro-subtasks, each moving one object from current position to the target with three atomic steps ($K=3n$). 

\begin{table}[htbp]
\centering
\caption{
\textbf{Homogeneous forward-composition success (\%).}
For each Homo$_n$, \textbf{one end-to-end policy} executes all $n$ \textbf{no-reset
transfers}. SR completes every gated atomic step.
}
\label{tab:forward-composition}

\scriptsize
\setlength{\tabcolsep}{2.2pt}
\renewcommand{\arraystretch}{0.90}

\resizebox{\linewidth}{!}{%
\begin{tabular}{@{}l*{10}{c}@{}}
\toprule
\textbf{Composition}
& Homo$_1$
& Homo$_2$
& Homo$_3$
& Homo$_4$
& Homo$_5$
& Homo$_6$
& Homo$_7$
& Homo$_8$
& Homo$_9$
& Homo$_{10}$
\\
\midrule

\textbf{$K$}
& 3
& 6
& 9
& 12
& 15
& 18
& 21
& 24
& 27
& 30
\\

\textbf{Success Rate}
& 100.00
& 100.00 
& 99.23 
& 100.00 
& 100.00
& 98.85
& 97.28
& 100.00
& 93.99
& 94.14
\\

\bottomrule
\end{tabular}%
}
\vspace{-0.3em}
\end{table}

Scratch policies maintain high success through Homo$_{10}$ with only a modest decrease after Homo$_9$. A clear failure boundary remains to be explored beyond Homo$_{10}$. \textbf{A non-stepwise control} that removes all stepwise design solves Homo$_1$ but fails on all substantial Homo$_{n}$. See Appendix~\ref{app:exp-composition} for detailed formulation, full experiment results and an additional fixed-length \textbf{heterogeneous} study.




\paragraph{Backward Regression.}
\label{par:exp-recovery}

We freeze clean-trained step-conditioned policies and inject controlled
regressions only at evaluation.

\begin{wraptable}{r}{0.4\columnwidth}
\vspace{-2.0em}
\centering
\caption{
\textbf{Recovery success(\%).}
}
\label{tab:backward-recovery}

\scriptsize
\setlength{\tabcolsep}{2.2pt}
\renewcommand{\arraystretch}{0.88}

\resizebox{\linewidth}{!}{%
\begin{tabular}{@{}lccccc@{}}
\toprule
&
\texttt{Pick}
&
\texttt{Push}
&
\texttt{Pull}
&
\texttt{YCB}
&
\textbf{Avg.}
\\
\midrule

Forward
& 0 & 4 & 2 & 0 & 1.5 \\

\GateVLM{}
& \textbf{64} & \textbf{92} & \textbf{92} & \textbf{56}
& \textbf{76.0} \\

Oracle
& 64 & 94 & 90 & 58 & 76.5 \\

\bottomrule
\end{tabular}%
}
\vspace{-1.0em}
\end{wraptable}

At test time, \GateVLM{} compares a valid RGB reference with the current
observation to re-estimate the step index. The policy
resumes execution using the corrected step-index input without any
recovery-specific training. Recovery Success Rate (Recovery SR) measures
task completion after disturbance. Timely Rollback Rate (TRR@10)
measures whether backward update is within 10 policy steps. \textit{Forward} uses the same actor but freezes the step index without rollback. Across our recovery tests, visual rollback achieves $76.0\%$ Recovery SR,
compared with $1.5\%$ for forward-only, and it is also timely with an average TRR@10 of $99.5\%$. This indicates
that with \method{} the policy can effciently recover from unseen
regressions once its persistent step context is corrected. See
Appendix~\ref{app:exp-recovery} for the full formulation.

\section{Diagnostic Ablations}
\label{sec:ablations}

\subsection{\GateVLM{} Invocation Cadence}
\label{sec:abl-transition}

The invocation schedule determines when semantic verification is requested. To test whether \method{} depends on task-conditioned query timing, we replace its default invocation
schedule $\chi_{state}$ with periodic \GateVLM{} queries $\chi_{freq}$ and vary only the \textbf{query interval} to test performance difference.

\begin{table*}[htbp]
\centering
\caption{
\textbf{\GateVLM{} invocation cadence.} Values are mean success rates (\%) $\pm$ standard deviation over three independently trained policies.
}
\label{tab:transition-frequency}

\small
\setlength{\tabcolsep}{4pt}
\renewcommand{\arraystretch}{0.60}

\begin{tabularx}{\textwidth}{@{}
>{\centering\arraybackslash}p{2.2cm}
>{\centering\arraybackslash\hsize=1.20\hsize}X
>{\centering\arraybackslash\hsize=1.00\hsize}X
>{\centering\arraybackslash\hsize=0.70\hsize}X
>{\centering\arraybackslash\hsize=1.10\hsize}X
@{}}

\toprule
\textbf{Benchmark}
&
\textbf{Cadence}
&
\textbf{Verifier}
&
\textbf{Interval}
&
\textbf{Success Rate} \\

\midrule

\multirow[c]{3}{*}{Meta-World}
& Frequent (fast)
& \multirow[c]{3}{*}{\GateVLM{}}
& 16
& 12.73$\pm$1.46 \\

& Frequent (mid)
&
& 64
& 23.60$\pm$2.16 \\

& Frequent (slow)
&
& 128
& 31.40$\pm$3.34 \\

\midrule

\multirow[c]{3}{*}{ManiSkill}
& Frequent (fast)
& \multirow[c]{3}{*}{\GateVLM{}}
& 8
& 31.25$\pm$3.66 \\

& Frequent (mid)
&
& 16
& 43.67$\pm$3.18 \\

& Frequent (slow)
&
& 32
& 38.33$\pm$1.20 \\

\midrule

\multirow[c]{3}{*}{SoftGym}
& Frequent (fast)
& \multirow[c]{3}{*}{\GateVLM{}}
& 8
& 28.56$\pm$5.74 \\

& Frequent (mid)
&
& 16
& 27.11$\pm$5.02 \\

& Frequent (slow)
&
& 32
& 28.22$\pm$2.34 \\

\bottomrule
\end{tabularx}

\end{table*}



Periodic invocation supports policy learning with benchmark-dependent preferred cadences: Meta-World favors slower queries, ManiSkill favors intermediate cadences, and SoftGym is relatively insensitive. Notably, the fastest cadence is substantially worse, especially on Meta-World. This is consistent with the conclusion in Sec.~\ref{sec:abl-verification-errors}: a delayed verification can be corrected later, whereas \textbf{a false acceptance} prematurely changes the step state and policy context. Thus, over-querying is more harmful than conservative querying. Full formulations are provided in Appendix~\ref{app:transition-variants}.

\subsection{Policy Context for Stepwise Reward Learning}
\label{sec:abl-stage-context}



Q1~\ref{sec:q1} studies whether exposing the step index to both policy
networks benefits learning. To further understand what enables effective
real-world deployment, we disentangle which policy components require this
context and whether privileged object and goal states can be replaced by
visual observations.

\begin{wraptable}[7]{r}{0.49\columnwidth}
\vspace{-2\baselineskip}
\centering

\caption{
\textbf{Policy-context ablations (\%).}
\href{q:step-state}{$z_t$}: Step index.
}
\label{tab:policy-context}

\scriptsize
\setlength{\tabcolsep}{2.2pt}
\renewcommand{\arraystretch}{0.88}

\begin{tabularx}{\linewidth}{
    @{}
    >{\centering\arraybackslash}p{0.05\linewidth}
    >{\raggedright\arraybackslash}X
    >{\centering\arraybackslash}p{0.17\linewidth}
    >{\centering\arraybackslash}p{0.17\linewidth}
    >{\centering\arraybackslash}p{0.25\linewidth}
    @{}
}
\toprule

&
\multicolumn{1}{c}{\textbf{Actor input}}
&
\textbf{Actor $z_t$}
&
\textbf{Critic $z_t$}
&
\textbf{SR}
\\

\midrule

1
& State obs.
& \xmark
& \checkmark
& 67.20$\pm$10.83
\\

2
& State obs.
& \xmark
& \xmark
& 58.67$\pm$2.49
\\

3
& DINOv2 + proprio
& \checkmark
& \checkmark
& 37.20$\pm$1.06
\\

4
& DINOv2 + proprio
& \xmark
& \checkmark
& 37.07$\pm$0.96
\\

\bottomrule
\end{tabularx}

\vspace{-0.9em}
\end{wraptable}

When the actor is step-blind, retaining the step index in the critic improves
mean success by around 10\%. Replacing object and goal states
with frozen visual features is substantially harder with success around 37\%. Within this visual setting, removing the actor's step input has
negligible effect, suggesting that the current
bottleneck is visual grounding rather than symbolic step context. The latter
variant further provides a deployment-compatible asymmetric actor using only
visual features and robot proprioception, while privileged information is
restricted to the training-time critic. Detailed formulations are provided
in Appendix~\ref{app:policy-context}.



\subsection{Sensitivity to Gate Errors}
\label{sec:abl-verification-errors}

A gate can fail by accepting an incomplete step
(\textbf{false-acceptance rate}, FAR) or rejecting a completed step
(\textbf{false-rejection rate}, FRR). For stepwise reward, We synthetically vary both error rates with the task decomposition and reward programs fixed.


\begin{wrapfigure}[14]{r}{0.48\columnwidth}
    \vspace{-1.0em}
    \centering
    \includegraphics[width=0.8\linewidth]{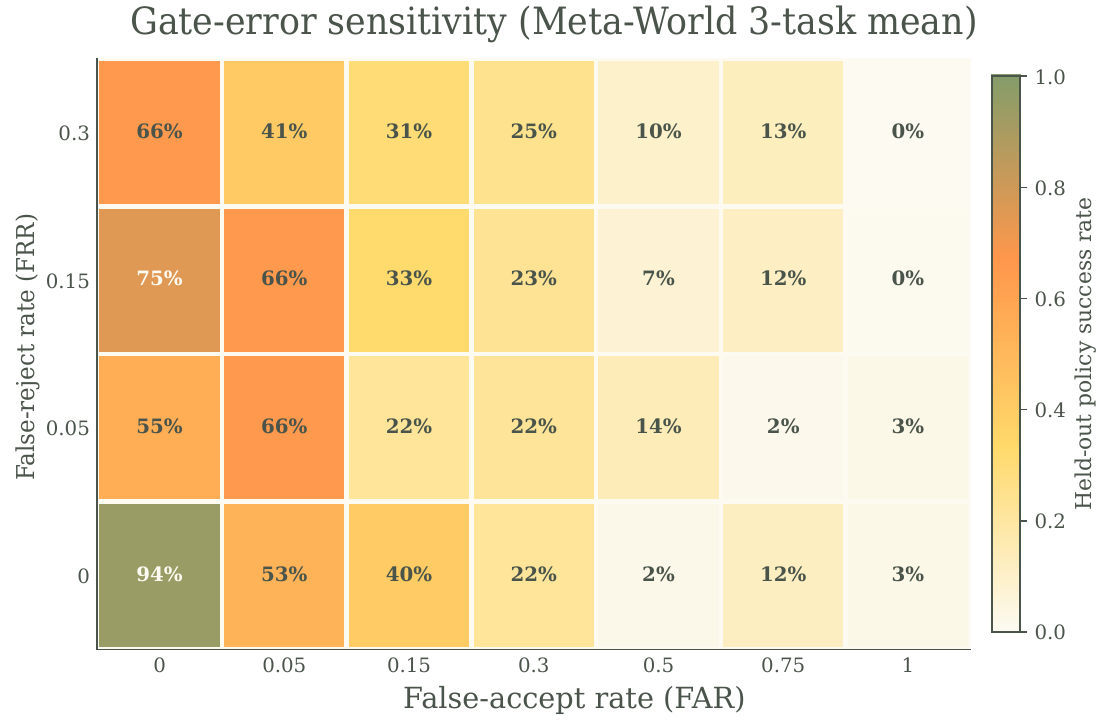}
    \vspace{-0.6em}
    \caption{
    \textbf{Sensitivity to gate errors.}
    Mean held-out policy success over Meta-World tasks and three
    training seeds. Columns vary FAR and rows vary FRR.
    }
    \label{fig:far-frr}
    \vspace{-1.0em}
\end{wrapfigure}

The results in Figure~\ref{fig:far-frr} show that FAR is substantially more damaging than FRR. With FRR fixed at zero,
increasing FAR from $0$ to $0.30$ reduces success from $94.0\%$ to
$21.8\%$; with FAR fixed at zero, increasing FRR to $0.30$ retains
$66.0\%$ success. A false rejection preserves the current step and permits
later retries, whereas a false acceptance prematurely advances the persistent
step state, issues a transition bonus, and reroutes both reward and policy
context. Gate operating points should therefore prioritize low on-policy FAR
rather than classification accuracy alone. See Appendix~\ref{app:gate-error} for the full protocol.

\section{Limitations and Future Work}

\paragraph{Visual Progress Data Engine.} The task structure from visual input to progress score of visual progress models is particularly important for
large-scale data annotation and curation, ranking action or trajectory
candidates, online data-quality assessment, and reward-weighted behavior
cloning. For onlin robot RL, \method shows that using VLMs as step verifiers could provide a generalizable and transferable step index, and its explicit transition conditions and controllable failures also offer an option for generating controllable process supervision. Together with simulation scene datasets and real-to-sim-to-real reconstruction, this may provide useful task-specific data for future progress reward models.

\paragraph{Limitations and open questions.} Our real-world experiments do not yet cover deformable object or fluid manipulation tasks. Our composition experiments also cover a limited range of task lengths and subtask combinations, leaving open how performance of stepwise reward framework scales with longer and
more semantically diverse sequences. Future work could also test whether adapting
\GateVLM{} to benchmark-specific visual domains improves policy learning.
Besides, additional training seeds and evaluation episodes would help quantify how the false acceptance rate and false
rejection rate of VLMs affect downstream policy performance as semantic subtask gates.








\newpage

\bibliography{iclr2027_conference}

@misc{jain2025polarisscalablerealtosimevaluations,
      title={PolaRiS: Scalable Real-to-Sim Evaluations for Generalist Robot Policies}, 
      author={Arhan Jain and Mingtong Zhang and Kanav Arora and William Chen and Marcel Torne and Muhammad Zubair Irshad and Sergey Zakharov and Yue Wang and Sergey Levine and Chelsea Finn and Wei-Chiu Ma and Dhruv Shah and Abhishek Gupta and Karl Pertsch},
      year={2025},
      eprint={2512.16881},
      archivePrefix={arXiv},
      primaryClass={cs.RO},
      url={https://arxiv.org/abs/2512.16881}, 
}

@misc{openai2026gpt55,
  author       = {{OpenAI}},
  title        = {Introducing GPT-5.5},
  year         = {2026},
  howpublished = {OpenAI},
  note         = {Accessed 2026}
}

@misc{qwen35,
  title  = {{Qwen3.5}: Towards Native Multimodal Agents},
  author = {{Qwen Team}},
  month  = {February},
  year   = {2026},
  url    = {https://qwen.ai/blog?id=qwen3.5}
}

@inproceedings{zhanggeneralist,
  title={A Generalist Pair-wise Progress Critic Model for Vision-Language-Action Robots},
  author={Zhang, Qi and Zhai, Shaopeng and Zhang, Shengzhe and Liu, Litao and Zhang, Tianyi and Huang, Fuxian and Zhou, Ming},
  booktitle={Forty-third International Conference on Machine Learning}
}

@article{progresssurvey2026,
  title        = {Progress Reward Modeling for Robotic Learning: A Comprehensive Survey},
  author       = {Zhang, Jianshu and Wu, Keliang and Lu, Haoran and Liu, Anbang and Zhang, Ce and Yin, Weijie and Qian, Chengxuan and Yang, Xiyuan and Pan, Zhenyu and Ye, Guo and Liu, Han},
  journal      = {arXiv preprint arXiv:2607.21655},
  year         = {2026},
  eprint       = {2607.21655},
  archivePrefix= {arXiv},
  primaryClass = {cs.RO},
  doi          = {10.48550/arXiv.2607.21655}
}

@inproceedings{moka2024,
  author    = {Fang, Kuan and Liu, Fangchen and Abbeel, Pieter and Levine, Sergey},
  title     = {MOKA: Open-World Robotic Manipulation through Mark-Based Visual Prompting},
  booktitle = {Proceedings of Robotics: Science and Systems},
  year      = {2024},
  address   = {Delft, Netherlands},
  month     = jul,
  doi       = {10.15607/RSS.2024.XX.062}
}

@inproceedings{dreureka2024,
  author    = {Ma, Yecheng Jason and Liang, William and Wang, Hung-Ju and
               Wang, Sam and Zhu, Yuke and Fan, Linxi and Bastani, Osbert and
               Jayaraman, Dinesh},
  title     = {{DrEureka}: Language Model Guided {Sim-to-Real} Transfer},
  booktitle = {Proceedings of Robotics: Science and Systems},
  year      = {2024},
  address   = {Delft, Netherlands},
  month     = jul,
  doi       = {10.15607/RSS.2024.XX.094}
}

@inproceedings{rekep2025,
  title     = {ReKep: Spatio-Temporal Reasoning of Relational Keypoint Constraints for Robotic Manipulation},
  author    = {Huang, Wenlong and Wang, Chen and Li, Yunzhu and Zhang, Ruohan and Fei-Fei, Li},
  booktitle = {Proceedings of The 8th Conference on Robot Learning},
  pages     = {4573--4602},
  year      = {2025},
  editor    = {Agrawal, Pulkit and Kroemer, Oliver and Burgard, Wolfram},
  volume    = {270},
  series    = {Proceedings of Machine Learning Research},
  month     = {06--09 Nov},
  publisher = {PMLR}
}

@inproceedings{rialto2024,
  author    = {Torne Villasevil, Marcel and Simeonov, Anthony and Li, Zechu and Chan, April and Chen, Tao and Gupta, Abhishek and Agrawal, Pulkit},
  title     = {Reconciling Reality through Simulation: A Real-To-Sim-to-Real Approach for Robust Manipulation},
  booktitle = {Proceedings of Robotics: Science and Systems},
  year      = {2024},
  address   = {Delft, Netherlands},
  month     = jul,
  doi       = {10.15607/RSS.2024.XX.015}
}

@inproceedings{re3sim2026,
  title     = {RE$^3$SIM: Generating High-Fidelity Simulation Data via 3D-Photorealistic Real-to-Sim for Robotic Manipulation},
  author    = {Han, Xiaoshen and Yu, Junqiu and Liu, Minghuan and Chen, Yilun and Lyu, Xiaoyang and Tian, Yang and Wang, Bolun and Zhang, Weinan and Pang, Jiangmiao},
  booktitle = {IEEE International Conference on Robotics and Automation (ICRA)},
  year      = {2026}
}

@inproceedings{rola,
  title={Robot learning from any images},
  author={Zhao, Siheng and Mao, Jiageng and Chow, Wei and Shangguan, Zeyu and Shi, Tianheng and Xue, Rong and Zheng, Yuxi and Weng, Yijia and You, Yang and Seita, Daniel and others},
  booktitle={Conference on Robot Learning},
  pages={4226--4245},
  year={2025},
  organization={PMLR}
}

@misc{lightman2023lets,
   title={Let's Verify Step by Step}, 
   author={Hunter Lightman and Vineet Kosaraju and Yura Burda and Harri Edwards and Bowen Baker and Teddy Lee and Jan Leike and John Schulman and Ilya Sutskever and Karl Cobbe},
   year={2023},
   eprint={2305.20050},
   archivePrefix={arXiv},
   primaryClass={cs.LG}
}

@article{ranawaka2026simfoundry,
  title   = {SimFoundry: Modular and Automated Scene Generation for Policy Learning and Evaluation},
  author  = {Ranawaka, Nadun and Wong, Josiah and Pai, Wei-Lin and Chu, Wei-Teng and
             Dai, Tianyuan and Moghani, Masoud and Yin, Hang and Jiang, Yunfan and
             Durbano, Wesley and Huynh, Brandon and Fang, Yu and Xu, Danfei and
             Zhang, Ruohan and {Fei-Fei}, Li and Fan, Linxi and Wen, Bowen and
             Mandlekar, Ajay and Zhu, Yuke},
  journal = {arXiv preprint arXiv:2606.28276},
  year    = {2026},
}

@InProceedings{liv2023,
  title = {{LIV}: Language-Image Representations and Rewards for Robotic Control},
  author = {Ma, Yecheng Jason and Kumar, Vikash and Zhang, Amy and Bastani, Osbert and Jayaraman, Dinesh},
  booktitle = {Proceedings of the 40th International Conference on Machine Learning},
  pages = {23301--23320},
  year = {2023},
  volume = {202},
  series = {Proceedings of Machine Learning Research},
  publisher = {PMLR},
  url = {https://proceedings.mlr.press/v202/ma23b.html}
}

@inproceedings{rewind2025,
  title={ReWi{ND}: Language-Guided Rewards Teach Robot Policies without New Demonstrations},
  author={Jiahui Zhang and Yusen Luo and Abrar Anwar and Sumedh Anand Sontakke and Joseph J Lim and Jesse Thomason and Erdem Biyik and Jesse Zhang},
  booktitle={9th Annual Conference on Robot Learning},
  year={2025},
  url={https://openreview.net/forum?id=XjjXLxfPou}
}

@inproceedings{sarm2025,
  title={{SARM}: Stage-Aware Reward Modeling for Long Horizon Robot Manipulation},
  author={Qianzhong Chen and Justin Yu and Mac Schwager and Pieter Abbeel and Fred Shentu and Philipp Wu},
  booktitle={The Fourteenth International Conference on Learning Representations},
  year={2026}
}

@misc{roboreward2026,
  title={RoboReward: General-Purpose Vision-Language Reward Models for Robotics},
  author={Tony Lee and Andrew Wagenmaker and Karl Pertsch and Percy Liang and Sergey Levine and Chelsea Finn},
  year={2026},
  eprint={2601.00675},
  archivePrefix={arXiv},
  primaryClass={cs.RO},
  url={https://arxiv.org/abs/2601.00675}
}

@article{topreward2026,
  title={TOPReward: Token Probabilities as Hidden Zero-Shot Rewards for Robotics},
  author={Chen, Shirui and Harrison, Cole and Lee, Ying-Chun and Yang, Angela Jin and Ren, Zhongzheng and Ratliff, Lillian J and Duan, Jiafei and Fox, Dieter and Krishna, Ranjay},
  journal={arXiv preprint arXiv:2602.19313},
  year={2026}
}

@article{lrm2026,
  title={Large Reward Models: Generalizable Online Robot Reward Generation with Vision-Language Models},
  author={Wu, Yanru and Yuan, Weiduo and Qi, Ang and Guizilini, Vitor and Mao, Jiageng and Wang, Yue},
  journal={arXiv preprint arXiv:2603.16065},
  year={2026}
}

@inproceedings{robometer2026,
  title={Robometer: Scaling General-Purpose Robotic Reward Models via Trajectory Comparisons},
  author={Anthony Liang and Yigit Korkmaz and Jiahui Zhang and Minyoung Hwang and Abrar Anwar and Sidhant Kaushik and Aditya Shah and Alex S. Huang and Luke Zettlemoyer and Dieter Fox and Yu Xiang and Anqi Li and Andreea Bobu and Abhishek Gupta and Stephen Tu and Erdem Biyik and Jesse Zhang},
  year={2026},
  booktitle={Robotics: Science and Systems 2026}
}

@article{soler12026,
  title={SOLE-R1: Video-Language Reasoning as the Sole Reward for On-Robot Reinforcement Learning},
  author={Schroeder, Philip and Weng, Thomas and Schmeckpeper, Karl and Rosen, Eric and Hart, Stephen and Biza, Ondrej},
  journal={arXiv preprint arXiv:2603.28730},
  year={2026},
  eprint={2603.28730},
  archivePrefix={arXiv},
  primaryClass={cs.RO},
  doi={10.48550/arXiv.2603.28730},
  url={https://arxiv.org/abs/2603.28730}
}

@misc{procvlm2026,
  title={ProcVLM: Learning Procedure-Grounded Progress Rewards for Robotic Manipulation},
  author={Youhe Feng and Hansen Shi and Haoyang Li and Xinlei Guo and Yang Wang and Chengyang Zhang and Jinkai Zhang and Xiaohan Zhang and Jie Tang and Jing Zhang},
  year={2026},
  eprint={2605.08774},
  archivePrefix={arXiv},
  primaryClass={cs.RO},
  url={https://arxiv.org/abs/2605.08774}
}

@misc{stdr2026,
  author={Yang Yang and Bingjie Chen and Zihan Wang and Yizhe Li and Guoping Pan and Yi Cheng and Houde Liu},
  title={{S}tage-{T}ransition {D}ense {R}eward {M}odeling for {R}einforcement {L}earning},
  year={2026},
  eprint={2606.31377},
  note={arXiv:2606.31377v1}
}

@misc{wvm2026,
  title={World Value Models for Robotic Manipulation},
  author={Zhihao Wang and Jianxiong Li and Yu Cui and Yuan Gao and Xianyuan Zhan and Junzhi Yu and Xiao Ma},
  year={2026},
  eprint={2606.24742},
  archivePrefix={arXiv},
  primaryClass={cs.RO},
  url={https://arxiv.org/abs/2606.24742}
}

@article{sarm22026,
  title={SARM2: Multi-Task Stage Aware Reward Modeling for Self Improving Robotic Manipulation},
  author={Chen, Qianzhong and Zheng, Hau and Yu, Justin and Huang, Suning and Sun, Jiankai and Goldberg, Ken and Wen, Chuan and Abbeel, Pieter and Shentu, Yide and Wu, Philipp and others},
  journal={arXiv preprint arXiv:2606.10305},
  year={2026}
}

@InProceedings{arm2026,
  author={Mao, Yiming and Yu, Zixi and Mao, Weixin and Li, Yinhao and Hu, Qirui and Lan, Zihan and Zhu, Minzhao and Chen, Hua},
  title={ARM: Advantage Reward Modeling for Long-Horizon Manipulation},
  booktitle={Proceedings of the IEEE/CVF Conference on Computer Vision and Pattern Recognition (CVPR) Workshops},
  month={June},
  year={2026},
  pages={4468-4477}
}

@article{robodopamine2025,
  title={Robo-Dopamine: General Process Reward Modeling for High-Precision Robotic Manipulation},
  author={Tan, Huajie and Chen, Sixiang and Xu, Yijie and Wang, Zixiao and Ji, Yuheng and Chi, Cheng and Lyu, Yaoxu and Zhao, Zhongxia and Chen, Xiansheng and Co, Peterson and Xie, Shaoxuan and Yao, Guocai and Wang, Pengwei and Wang, Zhongyuan and Zhang, Shanghang},
  journal={arXiv preprint arXiv:2512.23703},
  year={2025}
}

@InProceedings{rlvlmf2024,
  title={RL-VLM-F: Reinforcement Learning from Vision Language Foundation Model Feedback},
  author={Wang, Yufei and Sun, Zhanyi and Zhang, Jesse and Xian, Zhou and Biyik, Erdem and Held, David and Erickson, Zackory},
  booktitle={Proceedings of the 41st International Conference on Machine Learning},
  year={2024}
}

@InProceedings{furl2024,
  title={{F}u{RL}: Visual-Language Models as Fuzzy Rewards for Reinforcement Learning},
  author={Fu, Yuwei and Zhang, Haichao and Wu, Di and Xu, Wei and Boulet, Benoit},
  booktitle={Proceedings of the 41st International Conference on Machine Learning},
  pages={14256--14274},
  year={2024},
  volume={235},
  series={Proceedings of Machine Learning Research},
  publisher={PMLR},
  url={https://proceedings.mlr.press/v235/fu24j.html}
}

@inproceedings{gvl2025,
  title={{Vision Language Models Are In-Context Value Learners}},
  author={Ma, Yecheng Jason and Hejna, Joey and Fu, Chuyuan and Shah, Dhruv and Liang, Jacky and Xu, Zhuo and Kirmani, Sean and Xu, Peng and Driess, Danny and Xiao, Ted and Bastani, Osbert and Jayaraman, Dinesh and Yu, Wenhao and Zhang, Tingnan and Sadigh, Dorsa and Xia, Fei},
  booktitle={International Conference on Learning Representations},
  year={2025},
  url={https://mlanthology.org/iclr/2025/ma2025iclr-vision/}
}

@inproceedings{rewardmachines2018,
  author={Toro Icarte, Rodrigo and Klassen, Toryn Q. and Valenzano, Richard and McIlraith, Sheila A.},
  title={Using Reward Machines for High-Level Task Specification and Decomposition in Reinforcement Learning},
  booktitle={Proceedings of the 35th International Conference on Machine Learning (ICML)},
  year={2018},
  pages={2112--2121}
}

@inproceedings{text2reward2024,
  title={Text2Reward: Reward Shaping with Language Models for Reinforcement Learning},
  author={Tianbao Xie and Siheng Zhao and Chen Henry Wu and Yitao Liu and Qian Luo and Victor Zhong and Yanchao Yang and Tao Yu},
  booktitle={The Twelfth International Conference on Learning Representations},
  year={2024},
  url={https://openreview.net/forum?id=tUM39YTRxH}
}

@article{eureka2024,
  title={Eureka: Human-Level Reward Design via Coding Large Language Models},
  author={Yecheng Jason Ma and William Liang and Guanzhi Wang and De-An Huang and Osbert Bastani and Dinesh Jayaraman and Yuke Zhu and Linxi Fan and Anima Anandkumar},
  year={2023},
  journal={arXiv preprint arXiv: Arxiv-2310.12931}
}

@InProceedings{codeasreward2024,
  title={Code as Reward: Empowering Reinforcement Learning with {VLM}s},
  author={Venuto, David and Islam, Mohammad Sami Nur and Klissarov, Martin and Precup, Doina and Yang, Sherry and Anand, Ankit},
  booktitle={Proceedings of the 41st International Conference on Machine Learning},
  pages={49368--49387},
  year={2024},
  editor={Salakhutdinov, Ruslan and Kolter, Zico and Heller, Katherine and Weller, Adrian and Oliver, Nuria and Scarlett, Jonathan and Berkenkamp, Felix},
  volume={235},
  series={Proceedings of Machine Learning Research},
  month={21--27 Jul},
  publisher={PMLR},
  url={https://proceedings.mlr.press/v235/venuto24a.html}
}

@inproceedings{prm4rl2025,
  title={{Progress Reward Model for Reinforcement Learning via Large Language Models}},
  author={Zhang, Xiuhui and Gao, Ning and Jiang, Xingyu and Chen, Yihui and Pan, Yuheng and Zhang, Mohan and Deng, Yue},
  booktitle={Advances in Neural Information Processing Systems},
  year={2025},
  url={https://proceedings.nips.cc/paper_files/paper/2025/hash/91b047c5f5bd41ef56bfaf4ad0bd19e3-Abstract-Conference.html}
}

@inproceedings{armfm2026,
  author={Castanyer, Roger Creus and Mohamed, Faisal and Castro, Pablo Samuel and Neary, Cyrus and Berseth, Glen},
  title={ARM-FM: Automated Reward Machines via Foundation Models for Compositional Reinforcement Learning},
  booktitle={The Fourteenth International Conference on Learning Representations},
  year={2026}
}

@misc{robosnap2026,
  title={RoboSnap: One-Shot Real-to-Sim Scene Generation for Generalizable Robot Learning and Evaluation},
  author={Shujie Zhang and Jingkun Yi and Weipeng Zhong and Zirui Zhou and Yangkun Zhu and Hanqing Wang and Xudong Xu and Weinan Zhang and Chunhua Shen},
  year={2026},
  eprint={2607.06699},
  archivePrefix={arXiv},
  primaryClass={cs.RO},
  url={https://arxiv.org/abs/2607.06699}
}

@inproceedings{phys2real2026,
  title={Phys2Real: Fusing VLM Priors with Interactive Online Adaptation for Uncertainty-Aware Sim-to-Real Manipulation},
  author={Wang, Maggie and Tian, Stephen and Swann, Aiden and Shorinwa, Ola and Wu, Jiajun and Schwager, Mac},
  booktitle={IEEE International Conference on Robotics and Automation (ICRA)},
  year={2026}
}

@article{twinrl2026,
  title={TwinRL-VLA: Digital Twin-Driven Reinforcement Learning for Real-World Robotic Manipulation},
  author={Xu, Qinwen and Liu, Jiaming and Zhou, Rui and Shi, Shaojun and Han, Nuowei and Liu, Zhuoyang and Gu, Chenyang and Gu, Shuo and Yue, Yang and Huang, Gao and others},
  journal={arXiv preprint arXiv:2602.09023},
  year={2026}
}

@inproceedings{vip2023,
  title     = {{VIP}: Towards Universal Visual Reward and Representation via
               Value-Implicit Pre-Training},
  author    = {Ma, Yecheng Jason and Sodhani, Shagun and Jayaraman, Dinesh
               and Bastani, Osbert and Kumar, Vikash and Zhang, Amy},
  booktitle = {International Conference on Learning Representations},
  year      = {2023},
  url       = {https://mlanthology.org/iclr/2023/ma2023iclr-vip/}
}

@inproceedings{roboclip2023,
  title     = {{RoboCLIP}: One Demonstration is Enough to Learn Robot Policies},
  author    = {Sontakke, Sumedh A. and Zhang, Jesse and Arnold, S{\'e}bastien M. R.
               and Pertsch, Karl and B{\i}y{\i}k, Erdem and Sadigh, Dorsa
               and Finn, Chelsea and Itti, Laurent},
  booktitle = {Advances in Neural Information Processing Systems},
  volume    = {36},
  pages     = {55681--55693},
  year      = {2023},
  url       = {https://papers.nips.cc/paper_files/paper/2023/hash/ae54ce310476218f26dd48c1626d5187-Abstract-Conference.html}
}

@InProceedings{language2reward2023,
  title     = {Language to Rewards for Robotic Skill Synthesis},
  author    = {Yu, Wenhao and Gileadi, Nimrod and Fu, Chuyuan and Kirmani, Sean
               and Lee, Kuang-Huei and Arenas, Montserrat Gonzalez
               and Chiang, Hao-Tien Lewis and Erez, Tom and Hasenclever, Leonard
               and Humplik, Jan and Ichter, Brian and Xiao, Ted and Xu, Peng
               and Zeng, Andy and Zhang, Tingnan and Heess, Nicolas
               and Sadigh, Dorsa and Tan, Jie and Tassa, Yuval and Xia, Fei},
  booktitle = {Proceedings of The 7th Conference on Robot Learning},
  pages     = {374--404},
  year      = {2023},
  volume    = {229},
  series    = {Proceedings of Machine Learning Research},
  publisher = {PMLR},
  url       = {https://proceedings.mlr.press/v229/yu23a.html}
}

@misc{robodopamine22026,
  title         = {{Robo-Dopamine 2.0}: History-Conditioned and OOD-Aware
                   Process Reward Modeling for Robotic Manipulation},
  author        = {Xu, Yijie and Jin, Haopeng and Zhou, Run and Liu, Shengbang
                   and Chen, Sixiang and Cheng, Hongyang and Hu, Sicheng
                   and Co, Peterson and Luo, Jinwen and Tan, Huajie
                   and Zhang, Shanghang},
  year          = {2026},
  eprint        = {2608.15680},
  archivePrefix = {arXiv},
  primaryClass  = {cs.RO},
  url           = {https://arxiv.org/abs/2608.15680}
}

@misc{robormbench2026,
  title         = {Same Trajectory, Contradictory Rewards ({ROBORMBENCH}):
                   Paraphrase Fragility in Vision Language Reward Models},
  author        = {Jeung, Wonje and Yoon, Sangyeon and Hong, Hyesoo
                   and Cho, Yoonjun and Jeon, Dongjae and Kim, Bumjun
                   and Oh, Jean and Yu, Youngjae and No, Albert},
  year          = {2026},
  eprint        = {2609.05401},
  archivePrefix = {arXiv},
  primaryClass  = {cs.RO},
  url           = {https://arxiv.org/abs/2609.05401}
}

@inproceedings{vlmrm2024,
  title     = {Vision-Language Models are Zero-Shot Reward Models for
               Reinforcement Learning},
  author    = {Rocamonde, Juan and Montesinos, Victoriano and Nava, Elvis
               and Perez, Ethan and Lindner, David},
  booktitle = {The Twelfth International Conference on Learning Representations},
  year      = {2024},
  url       = {https://openreview.net/forum?id=N0I2RtD8je}
}

@InProceedings{decisionnce2024,
  title     = {{DecisionNCE}: Embodied Multimodal Representations via
               Implicit Preference Learning},
  author    = {Li, Jianxiong and Zheng, Jinliang and Zheng, Yinan and Mao, Liyuan
               and Hu, Xiao and Cheng, Sijie and Niu, Haoyi and Liu, Jihao
               and Liu, Yu and Liu, Jingjing and Zhang, Ya-Qin and Zhan, Xianyuan},
  booktitle = {Proceedings of the 41st International Conference on Machine Learning},
  pages     = {29461--29488},
  year      = {2024},
  volume    = {235},
  series    = {Proceedings of Machine Learning Research},
  publisher = {PMLR},
  url       = {https://proceedings.mlr.press/v235/li24cr.html}
}

@misc{scalingsim2real2026,
  title={Scaling Sim-to-Real Reinforcement Learning for Robot VLAs with Generative 3D Worlds},
  author={Andrew Choi and Xinjie Wang and Zhizhong Su and Wei Xu},
  year={2026},
  eprint={2603.18532},
  archivePrefix={arXiv},
  primaryClass={cs.RO},
  url={https://arxiv.org/abs/2603.18532}
}

@misc{rlinfvla2025,
  title={RLinf-VLA: A Unified and Efficient Framework for VLA+RL Training},
  author={Hongzhi Zang and Mingjie Wei and Si Xu and Yongji Wu and Zhen Guo and Yuanqing Wang and Hao Lin and Liangzhi Shi and Yuqing Xie and Zhexuan Xu and Zhihao Liu and Kang Chen and Wenhao Tang and Quanlu Zhang and Weinan Zhang and Chao Yu and Yu Wang},
  year={2025},
  eprint={2510.06710},
  archivePrefix={arXiv},
  primaryClass={cs.RO},
  url={https://arxiv.org/abs/2510.06710}
}

@misc{marvl2026,
  title={MARVL: Multi-Stage Guidance for Robotic Manipulation via Vision-Language Models},
  author={Xunlan Zhou and Xuanlin Chen and Shaowei Zhang and Xiangkun Li and ShengHua Wan and Xiaohai Hu and Yuan Lei and Le Gan and De-chuan Zhan},
  year={2026},
  eprint={2602.15872},
  archivePrefix={arXiv},
  primaryClass={cs.RO},
  url={https://arxiv.org/abs/2602.15872}
}

@misc{core2026,
  title={CoRe: Combined Rewards with Vision-Language Model Feedback for Preference-Aligned Reinforcement Learning},
  author={Hexian Ni and Tao Lu and Yinghao Cai},
  year={2026},
  eprint={2607.01721},
  archivePrefix={arXiv},
  primaryClass={cs.RO},
  url={https://arxiv.org/abs/2607.01721}
}

@article{vllr2026,
  title={Generalizable Dense Reward for Long-Horizon Robotic Tasks},
  author={Silong Yong and Stephen Sheng and Carl Qi and Xiaojie Wang and Evan Sheehan and Anurag Shivaprasad and Yaqi Xie and Katia Sycara and Yesh Dattatreya},
  year={2026},
  eprint={2604.00055},
  archivePrefix={arXiv},
  url={https://arxiv.org/abs/2604.00055}
}

@inproceedings{metaworld2019,
  title={Meta-World: A Benchmark and Evaluation for Multi-Task and Meta Reinforcement Learning},
  author={Tianhe Yu and Deirdre Quillen and Zhanpeng He and Ryan Julian and Karol Hausman and Chelsea Finn and Sergey Levine},
  booktitle={Conference on Robot Learning (CoRL)},
  year={2019},
  eprint={1910.10897},
  archivePrefix={arXiv},
  primaryClass={cs.LG},
  url={https://arxiv.org/abs/1910.10897}
}

@article{maniskill3,
  title={ManiSkill3: GPU Parallelized Robotics Simulation and Rendering for Generalizable Embodied AI},
  author={Stone Tao and Fanbo Xiang and Arth Shukla and Yuzhe Qin and Xander Hinrichsen and Xiaodi Yuan and Chen Bao and Xinsong Lin and Yulin Liu and Tse-kai Chan and Yuan Gao and Xuanlin Li and Tongzhou Mu and Nan Xiao and Arnav Gurha and Viswesh Nagaswamy Rajesh and Yong Woo Choi and Yen-Ru Chen and Zhiao Huang and Roberto Calandra and Rui Chen and Shan Luo and Hao Su},
  journal={Robotics: Science and Systems},
  year={2025}
}

@inproceedings{softgym2020,
  title={SoftGym: Benchmarking Deep Reinforcement Learning for Deformable Object Manipulation},
  author={Lin, Xingyu and Wang, Yufei and Olkin, Jake and Held, David},
  booktitle={Conference on Robot Learning},
  year={2020}
}

@misc{ppo2017,
  title={Proximal Policy Optimization Algorithms},
  author={John Schulman and Filip Wolski and Prafulla Dhariwal and Alec Radford and Oleg Klimov},
  year={2017},
  eprint={1707.06347},
  archivePrefix={arXiv},
  primaryClass={cs.LG},
  url={https://arxiv.org/abs/1707.06347}
}

@misc{dinov2_2023,
  title={DINOv2: Learning Robust Visual Features without Supervision},
  author={Oquab, Maxime and Darcet, Timoth\'ee and Moutakanni, Theo and Vo, Huy V. and Szafraniec, Marc and Khalidov, Vasil and others},
  journal={arXiv:2304.07193},
  year={2023}
}

@INPROCEEDINGS{asymmetricac2018,
  AUTHOR={Lerrel Pinto and Marcin Andrychowicz and Peter Welinder and Wojciech Zaremba and Pieter Abbeel},
  TITLE={Asymmetric Actor Critic for Image-Based Robot Learning},
  BOOKTITLE={Proceedings of Robotics: Science and Systems},
  YEAR={2018},
  ADDRESS={Pittsburgh, Pennsylvania},
  MONTH={June},
  DOI={10.15607/RSS.2018.XIV.008}
}

@article{ji2026prmjudge,
  title   = {PRM-as-a-Judge: A Dense Evaluation Paradigm for Fine-Grained Robotic Auditing},
  author  = {Ji, Yuheng and Liu, Yuyang and Tan, Huajie and Huang, Xuchuan and Huang, Fanding and others},
  journal = {arXiv preprint arXiv:2603.21669},
  year    = {2026},
  url     = {https://arxiv.org/abs/2603.21669}
}

@article{liu2026prmjudge15,
  title   = {PRM-as-a-Judge 1.5: A Toolkit for Robot Process Assessment},
  author  = {Liu, Yuyang and Shen, Yanqing and Chen, Ruike and Zhao, Jifan and Tian, Yuxuan and others},
  journal = {arXiv preprint arXiv:2608.14284},
  year    = {2026},
  url     = {https://arxiv.org/abs/2608.14284}
}
\bibliographystyle{iclr2027_conference}

\clearpage
\appendix

\renewcommand{\thesubsection}{\thesection-\arabic{subsection}}
\renewcommand{\thesubsubsection}{\thesubsection.\arabic{subsubsection}}

\startcontents[appendix]

\phantomsection
\begin{center}
    {\Large\scshape\bfseries\color{RoboDark}
     Appendix Table of Contents}
\end{center}

\vspace{-0.5em}

\begingroup

\hypersetup{hidelinks}

\setcounter{tocdepth}{2}
\printcontents[appendix]{}{1}{}

\endgroup

\section{RoboStep Method Details}
\label{app:implementation}

\subsection{Reflective Reward Compilation}
\label{app:reward-compilation}

For each task across the benchmarks we tested, if the development validation success remains below \(50\%\), \FrameVLM{} receives the task specification, simulator-state interface, current reward program, and training diagnostics, and proposes one reward/stage repair while the PPO policy-learning protocol is kept fixed. \textbf{The procedure is limited to 5 compilation rounds}, after which the round-five program is frozen for the main-table evaluation.

\subsection{Detailed Formulation}
\label{app:formulation}

\paragraph{State-side invocation trigger.}
The invocation rule $\chi$ in Eq.~\ref{eq:reward-recipe} determines only
when semantic verification is requested. In the reported simulation
experiments, $\chi_t$ is instantiated using state/API-based transition
candidates. For this simulation-specific instantiation, \FrameVLM{}
additionally generates a candidate predicate $c_i:O_b\rightarrow\{0,1\}$
and a consecutive-hold requirement $H_i$ for each step $i$.

For the active step, $c_i(o_t^{(b)})$ evaluates whether the current state
satisfies the physical conditions for requesting a possible transition.
To suppress transient fluctuations, we maintain

\begin{equation}
q_i(t+1)
=
\begin{cases}
q_i(t)+1, & c_i(o_t^{(b)})=1,\\
0,        & c_i(o_t^{(b)})=0.
\end{cases}
\label{eq:sim-trigger-counter}
\end{equation}

The corresponding invocation trigger is

\begin{equation}
\chi_t^{\mathrm{state}}
=
\mathbf{1}[z_t<K]\,
\mathbf{1}
\left[
q_{z_t}(t)\ge H_{z_t}
\right],
\label{eq:sim-trigger}
\end{equation}

and we set $\chi_t=\chi_t^{\mathrm{state}}$ for the reported simulation
experiments. When $\chi_t=1$, the current RGB packet $\mathcal{I}_t$ is sent
to \GateVLM{} for semantic verification:

\begin{equation}
v_t
=
\mathbf{1}
\left[
\GateVLM{}
\left(
\ell,
z_t,
\mathcal{V}_{z_t},
\mathcal{I}_t
\right)
=
\texttt{success}
\right],
\qquad
e_t=\chi_t v_t,
\qquad
z_{t+1}=z_t+e_t.
\label{eq:sim-transition}
\end{equation}

When no query is issued, or when \GateVLM{} returns failure or uncertainty,
$v_t=0$. Thus, the state/API trigger determines only when visual
verification is requested, while \GateVLM{} provides the semantic
transition authority.

\paragraph{Deployment-compatible triggers.}
The state-side trigger above relies on simulator-accessible quantities, but
the verification interface itself is trigger-agnostic. When such state/API
candidates are unavailable, $\chi_t$ can instead be instantiated as a
periodic trigger:

\begin{equation}
\chi_t^{\mathrm{freq}}
=
\mathbf{1}[z_t<K]\,
\mathbf{1}[t\bmod F_b=0],
\label{eq:frequent-trigger}
\end{equation}

where $F_b$ is the query interval measured in the native policy steps of
benchmark $b$. We evaluate this frequency-based instantiation and its
benchmark-specific query cadences in
Appendix~\ref{app:transition-variants}.

A possible fully visual alternative is to invoke verification after a
sufficient change in a frozen RGB representation:

\begin{equation}
\chi_t^{\mathrm{vis}}
=
\mathbf{1}[z_t<K]\,
\mathbf{1}
\left[
\left\|
f(I_t)-f(I_{t-\Delta})
\right\|_2>\eta
\right],
\label{eq:visual-change-trigger}
\end{equation}

where $f(\cdot)$ is a frozen visual encoder. We include this only as a
deployment-compatible trigger option; it is not evaluated in the reported
experiments.

\subsection{Frame-VLM and Gate-VLM Inputs}
\label{app:vlm-prompts}

We use two VLM components with distinct computational roles.
\FrameVLM{} is used only during offline reward compilation and repair, where
it must decompose the task, reason over the declared state/API interface, and
generate executable stepwise reward logic. Since its inference cost is
incurred before policy optimization and amortized over the entire training
run, we prioritize reasoning and code-generation capability over latency.
Unless otherwise stated, we instantiate \FrameVLM{} with
\texttt{gpt-5.5-2026-04-23} using \texttt{xhigh} reasoning effort
~\citep{openai2026gpt55}.

\GateVLM{}, in contrast, lies on the online transition path. In the reported
simulation experiments, it is queried only when the state-side candidate and
its dwell requirement trigger a verification request; when simulator state is
unavailable, the same interface can instead use periodic or visual-change
invocation. The trigger controls only when verification is requested and does
not itself authorize a transition. Unless otherwise stated, we instantiate
\GateVLM{} with \texttt{Qwen/Qwen3.5-9B}~\citep{qwen35}, prioritizing
multimodal inference efficiency over the more expensive reasoning used for
offline reward compilation.

Because the latest verified step state is held between queries, \GateVLM{}
does not need to operate at the policy or low-level controller frequency.
Instead, each sparse online query must process the prescribed RGB packet and
return exactly one parseable decision from
\{\texttt{success}, \texttt{failure}, \texttt{unknown}\} before the
corresponding request becomes stale. Invalid, unparsable, and
\texttt{unknown} outputs fail closed and never advance the persistent step
state. The verifier receives only the specified visual-language evidence,
rather than simulator state, reward values, or evaluation labels.

Except for the experiments where the VLMs become part of the variant factors, all of \method{} experiments follows the above selection.

The prompts below show the reusable templates; task-specific content is
inserted into the bracketed fields.

\begin{expquestion}{RoboOrange}{
    Frame-VLM Prompt
}
\phantomsection
\label{sec:frame-vlm-prompt}

\small
\begin{verbatim}
You are designing a stepwise reward program 
for a robot manipulation task.

Task description: {task_description}
Scene and goal conventions: {scene_and_goal_description}
Available state APIs: {available_state_apis}
Design constraints: {design_constraints}

Decompose the task into feasible, 
ordered stages. For each stage, specify:
- its semantic objective and prerequisites;
- a bounded dense progress potential using only the available states;
- a persistent candidate-completion predicate and dwell requirement;
- a bounded, one-time transition bonus;
- maintenance and safety conditions that prevent shortcut solutions;
- visible evidence that would allow an RGB-only verifier to 
check completion.

Do not invent unavailable observations. Distinguish progress toward
a stage from completion of that stage. 

Return the stage definitions and executable
reward logic in the required format: {output_schema}.
\end{verbatim}

\normalsize

\end{expquestion}

\FrameVLM{} receives the task specification, scene conventions, available
state/API interface, and task-level design constraints. Its outputs 
\begin{equation}
\mathcal{R}(\ell,O_b)
=
\left(
\left\{
(\phi_i,M_i,b_i,\mathcal{V}_i)
\right\}_{i=0}^{K-1},
\chi
\right).
\label{eq:reward-recipe}
\end{equation}

as defined in Sec.~\ref{sec:method}.

\begin{expquestion}{RoboGreen}{
    Gate-VLM Prompt
}
\phantomsection
\label{sec:gate-vlm-prompt}

\small
\begin{verbatim}
You are a visual verifier of robot manipulation stage completion.

Task: {task_description}
Current stage: {stage_id_and_name}
Visible completion condition: {visible_completion_contract}
Evidence: {episode_start_image}, {ordered_recent_frames},
          {current_frozen_image}

A fast state-side predicate has proposed a transition candidate.
This is not a completion label. Judge only whether the CURRENT stage 
is visibly complete in the current image, using earlier images to 
establish identity and motion. 
Do not require a later stage. Reject a visibly wrong object, 
wrong target, robot-only motion when object motion is required, 
loss of a required relation, or other contradictory evidence. 

Do not infer precise hidden distances, forces, contacts,
or benchmark scores from RGB.

Return exactly one decision from {allowed_outputs}.
\end{verbatim}
\end{expquestion}

At each candidate boundary, \GateVLM{} receives the task description, current
stage, its visible completion contract $v_i$, and RGB evidence consisting of
the episode-start image, optional temporally ordered recent frames, and the
current frozen image. In the two-frame implementation,
\texttt{\{ordered\_recent\_frames\}} is empty.

The response is parsed as exactly one element of
$$
\hat g_i \in
\{\texttt{success},\texttt{failure},\texttt{unknown}\}.
$$
No scalar progress value is parsed from \GateVLM{}. A forward transition is
authorized only when the persistent state-side candidate and visual
verification agree:
$$
c_i \text{ sustained for } H_i
\quad\land\quad
\hat g_i=\texttt{success}
\quad\Longrightarrow\quad
z_t \rightarrow z_t+1.
$$
Outputs \texttt{failure} and \texttt{unknown} do not authorize the transition.
\GateVLM{} therefore verifies stage completion without modifying the dense
potential generated by \FrameVLM{}.

 \subsection{Reward and Subtask Example}
  \label{app:reward-programs}

  \noindent
  We show one representative reward program from each benchmark. Each program
  is generated from the task description and the benchmark state interface,
  and follows
  \[
  \mathcal{R}(\ell,O_b)
  =
  \left\{
  (\phi_i,M_i,c_i,H_i,b_i,\mathcal{V}_i)
  \right\}_{i=0}^{K-1}.
  \]

  \begin{expquestion}{RoboGold}{Meta-World}
  \phantomsection
  \label{sec:q2-reward-metaworld}

  \textbf{Task: \texttt{stick-push-v3}.}

  The generated subtasks are:
\[
\begin{aligned}
\text{approach the stick}
&\rightarrow
\text{capture and lift the stick}
\rightarrow
\text{contact the secondary object} \\
&\rightarrow
\text{push the object to the target}.
\end{aligned}
\]

  For \(K=4\), the stage potentials are summarized as
  \[
  \begin{aligned}
  \phi_0&=0.9q_{\mathrm{pre}}+0.1q_{\mathrm{open}},\\
  \phi_1&=0.25q_{\mathrm{grasp}}+0.30q_{\mathrm{held}}+0.45q_{\mathrm{lift}},\\
  \phi_2&=\sqrt{q_{\mathrm{contact}}q_{\mathrm{held}}},\\
  \phi_3&=0.7q_{\mathrm{outcome}}+0.2q_{\mathrm{contact}}
  +0.1q_{\mathrm{held}}.
  \end{aligned}
  \]
  The candidates \(c_i\) test pre-grasp, held-and-lifted, tool--object contact,
  and object--target completion. The program uses
  \[
  H=(2,2,2,1),\qquad
  b=(16,70,230,320),
  \]
  with \(\mathcal{V}_i\) containing TCP, stick, object, target, gripper, and
  relative-distance features. \(M_i\) retains the relevant grasp, contact, and
  outcome conditions.

  \end{expquestion}

  \begin{expquestion}{RoboGold}{ManiSkill}
  \phantomsection
  \label{sec:q2-reward-maniskill}

  \textbf{Task: \texttt{PegInsertionSide-v1}.}

  The generated subtasks are:
  \[
  \text{reach the peg}
  \rightarrow
  \text{grasp the peg}
  \rightarrow
  \text{align the peg with the hole}
  \rightarrow
  \text{insert the peg}.
  \]

  For \(K=4\), the state-based program uses
  \[
  \begin{aligned}
  \phi_0&=q_{\mathrm{reach}},\\
  \phi_1&=0.65g+0.35q_{\mathrm{reach}},\\
  \phi_2&=0.75q_{\mathrm{align}}+0.25g,\\
  \phi_3&=0.80q_{\mathrm{insert}}+0.15q_{\mathrm{align}}
  +0.05\,\mathrm{success}.
  \end{aligned}
  \]
  The candidates \(c_i\) test TCP--peg proximity, grasping and hole alignment. The program uses
  \[
  H=(2,2,2,2),\qquad
  b=(0.20,0.47,0.73,1.00),
  \]
  with \(\mathcal{V}_i\) containing TCP, peg and peg-head poses, hole and goal
  poses, grasp state, alignment error, and success. \(M_i\) maintains grasp,
  alignment, and insertion quality.

  \end{expquestion}

  \begin{expquestion}{RoboGold}{SoftGym}
  \phantomsection
  \label{sec:q2-reward-softgym}

  \textbf{Task: \texttt{PassWater}.}

  The generated subtasks are:
  \[
  \text{transport the cup while retaining water}
  \rightarrow
  \text{deliver water near the target}.
  \]

  For \(K=2\), let \(m\) denote normalized cup transport, \(q\) normalized
  distance to the target, and \(\rho\) retained-water fraction. The generated
  potentials are
  \[
  \phi_0=[0.80m\rho+0.20\rho]_{[0,1]},
  \qquad
  \phi_1=[0.70q\rho+0.30\rho]_{[0,1]}.
  \]
  The candidates are
  \[
  c_0:\phi_0\geq0.35,
  \qquad
  c_1:\phi_1\geq0.75,
  \]
  with
  \[
  H=(2,2),\qquad b=(1.5,1.5).
  \]
  The state set \(\mathcal{V}_i\) contains cup and target positions, water
  particle positions, glass geometry, and containment indicators. The
  retention term is included directly in the potentials, so \(M_i=0\) in the
  clean recipe. The later spill-aware version preserves this structure while
  adding stricter rejection of empty cups and major spills.

  \end{expquestion}

\section{Benchmark Experimental Protocols}
\label{app:benchmarks}

Q1 in Sec.~\ref{sec:exp-stepwise} shows that an explicit stepwise reward structure and step-index context substantially improve policy learning over conventional dense rewards.
However, the step transitions in Q1 are provided by privileged
\GateRule{} predicates. Such predicates can be difficult to specify outside
simulation, particularly for tasks whose intermediate completion cannot be
reliably characterized by simple state, distance, or pose thresholds, such as
deformable-object manipulation. We therefore replace rule-based transition
detection with visual-language verification, \GateVLM{}, and investigate
whether VLMs are better used as global visual progress predictors or as
semantic step-completion verifiers for online reward.

\subsection{Benchmark Configurations, Task Definitions and Success Criteria}
\label{app:task-definitions}

Across the 75 tasks, the number of active steps ranges from 2 to 7.
Meta-World tasks use 3--5 steps, while all six SoftGym tasks use two.
The largest decomposition is \texttt{StackPyramid-v1} in ManiSkill, with
seven active steps and therefore an eight-dimensional step encoding
including the terminal state. Full task definitions and reward decompositions
are provided in Appendix~\ref{app:benchmarks}.

\subsubsection{Benchmark Comparison Configurations}

For all benchmark comparisons, we match the policy-learning setup within each task and vary only the reward and transition-verification mechanism under study. Specifically, the environment version, reset distribution, action and controller interface, actor--critic architecture, PPO optimization protocol, rollout budget, random seeds, checkpoint-selection rule, and evaluation protocol are held fixed across reward methods. The benchmark-specific network architectures and PPO hyperparameters used in these matched comparisons are summarized in Table~\ref{tab:q2-ppo-config}.
\begin{table}[htbp]
\centering

\caption{
\textbf{Matched PPO policy-learning configurations.}
Configurations are fixed across reward methods within each benchmark.
Per task-specific training budgets and according reward functions are provided in Appendix~\ref{app:observations}.}
\label{tab:q2-ppo-config}

\scriptsize
\setlength{\tabcolsep}{2.5pt}
\renewcommand{\arraystretch}{0.88}

\begin{tabularx}{\linewidth}{
    @{}
    l
    >{\centering\arraybackslash}X
    >{\centering\arraybackslash}X
    >{\centering\arraybackslash}X
    >{\centering\arraybackslash}X
    c c c
    @{}
}
\toprule
\textbf{Benchmark}
& \textbf{Actor/Critic}
& \textbf{Rollout}
& \textbf{Epochs / Mini.}
& \textbf{LR / Schedule}
& $\boldsymbol{\gamma}$
& $\boldsymbol{\lambda}$
& \textbf{Clip}
\\
\midrule

Meta-World
& 256--256--128, ELU
& $32\times64$
& $8/8$
& Adaptive KL
& .99
& .95
& .20
\\

ManiSkill
& $3\times256$, Tanh
& $16\times50$
& $4/32$
& $3{\times}10^{-4}$
& .80
& .90
& .20
\\

SoftGym
& 256--256--128, ELU
& 256 steps/update
& $4/4$
& $3{\times}10^{-4}$
& .99
& .95
& .20
\\

\bottomrule
\end{tabularx}
\end{table}

These policy backbones are benchmark-specific controls rather than components
  optimized for our reward method. For ManiSkill, we follow the three-layer,
  256-unit Tanh actor--critic and core PPO hyperparameters of its official PPO
  example. For Meta-World and SoftGym, we use a 256--256--128 ELU actor--critic
  within each benchmark. We hold policy capacity fixed within a task to isolate the effect of reward design and transition verification.

 As detailed in Table~\ref{tab:policy-input-settings}, the main comparisons use each benchmark's native state observation together with the active-step code for both actor and critic, while RGB observations are provided only to visual reward or transition-verification modules and are never concatenated with the policy input. For a fixed task, the policy observation layout and step-code width are therefore shared across compared reward methods; variants that remove or hide the step code, or replace the state-based actor with visual features, are evaluated separately as input ablations.

\begin{table*}[htbp]

\centering
\caption{
\textbf{Policy-input protocol used in the main reward comparison.}
For each task, all reward methods share the same policy observation and PPO configuration.
}
\label{tab:policy-input-settings}

\small
\setlength{\tabcolsep}{3pt}
\renewcommand{\arraystretch}{1.12}

\begin{tabularx}{\textwidth}{@{}
>{\raggedright\arraybackslash}p{1.35cm}
>{\raggedright\arraybackslash\hsize=0.90\hsize}X
>{\raggedright\arraybackslash\hsize=1.15\hsize}X
>{\raggedright\arraybackslash\hsize=0.95\hsize}X
@{}}

\toprule
\textbf{Bench} &
\makecell[l]{\textbf{Benchmark-native}\\\textbf{state $x_t$}} &
\makecell[l]{\textbf{Policy observation}\\\textbf{(actor / critic)}} &
\makecell[l]{\textbf{Visual reward / gate}\\\textbf{observation}} \\
\midrule

\textbf{Meta-World}
&
Fully observable Sawyer state:
current hand/TCP position, gripper opening,
two object-pose slots, previous state, and task goal.

\textit{Dim.:}
$39=18_{\mathrm{cur}}+18_{\mathrm{prev}}+3_{\mathrm{goal}}$.
&
\[
\tilde{o}_t=[x_t;e(s_t)]
\in\mathbb{R}^{39+K+1}.
\]

Across reported tasks, the policy input has
\textbf{41--45 dimensions}.
&
Fixed-camera RGB snapshots or temporal packets are used only by the
visual reward model or \GateVLM{} transition verifier and are never
provided to the policy.
\\

\midrule

\textbf{ManiSkill}
&
Flattened \texttt{obs\_mode=state}:
robot proprioception and end-effector state,
together with task-specific object, goal, and interaction states.

\textit{Dim.:}
$d_{\tau}=31$--$76$ across the 19 tasks.
&
\[
\tilde{o}_t=[x_t;e(s_t)]
\in\mathbb{R}^{d_{\tau}+K+1}.
\]

Across reported tasks, the policy input has
\textbf{36--82 dimensions}.
&
Rendered RGB observations are queried only by the visual reward or
step-verification module; the PPO actor and critic remain state-based.
\\

\midrule

\textbf{SoftGym}
&
Task-native \texttt{key\_point} state:

cloth: four corners + two picker positions (\textbf{18-D});

rope: ten keypoints + two picker positions (\textbf{36-D});

PassWater: cup/retained-water state (\textbf{7-D});

PourWater: two-cup/water state (\textbf{13-D}).
&
All six tasks use $K=2$:
\[
\tilde{o}_t=[x_t;e(s_t)]
\in\mathbb{R}^{d_{\tau}+3}.
\]

Input dimensions are
\textbf{21-D} for cloth,
\textbf{39-D} for rope,
\textbf{10-D} for PassWater, and
\textbf{16-D} for PourWater.
&
$320\times320$ RGB frames and their temporal buffer are consumed only by
the visual reward / gate module and are never provided to the PPO networks.
\\

\bottomrule
\end{tabularx}

\vspace{3pt}
\begin{minipage}{0.985\textwidth}
\end{minipage}

\end{table*}

$e(s_t)\in\{0,1\}^{K+1}$ denotes the one-hot active-step code,
including an additional absorbing/completed state.
For a fixed task, all main-table methods use the same benchmark-native state,
action space, PPO architecture, actor/critic observation tensor, step-code width,
and controller interface.
Reward-model images are never concatenated with the policy input.
Removing the stage code, hiding it from the actor, or replacing the state actor
with an RGB encoder is evaluated separately as an input ablation.

\subsubsection{Observations and Stepwise Reward of Each Task}
\label{app:observations}

We separate the information available to four components: the policy, critic,
\FrameVLM{} reward program, and the live VLM verifier. The policy and critic
receive the benchmark training observation together with the persistent active
stage, whereas the reward program may additionally use task-relevant simulator
state to compute the dense potential and candidate transition predicates. The
live VLM verifier is kept on the visual side of this boundary: simulator state
can schedule a candidate verification, but the verifier itself judges visual
evidence rather than receiving the privileged geometric quantities used by the
reward program. Table~\ref{tab:task_reward_spec} gives the task-level state/API inputs used
for dense-reward construction, together with the stage decomposition and
dense-potential provenance.

\paragraph{MetaWorld.}
The actor receives the standard 39-dimensional MetaWorld state concatenated
with a $(K+1)$-way one-hot code for the Qwen-authorized persistent step; the
critic uses the same observation. The selected PPO runs do not use the native
task reward. Instead, the generated reward programs use task-dependent
geometric quantities such as TCP, object/control, target, and initial poses,
gripper opening, and, where required, object/tool orientation. A state rule
only proposes a transition. The VLM verifier receives the episode-start
reference image, a time-ordered ring of recent RGB frames, the frozen current
camera frame, and the semantic instruction for the active step; metric
distances and the native success flag are not included in the verifier packet.

\paragraph{ManiSkill.}
The selected policies and critics use the simulator state concatenated with the
active-stage one-hot code; the selected runs set the RGB actor input to false.
The reward backend is the generated stepwise rule program. Its task-specific
state fields are listed in Table~\ref{tab:task_reward_spec} and include, for
example, poses, geometric errors, grasp/contact indicators, velocities, and,
where configured for final-step logic, simulator success signals. The VLM step observer is evaluated from
rendered RGB evidence using step-specific visible-completion descriptions,
rather than by directly exposing these reward-program state fields to the
verifier.

\paragraph{SoftGym.}
All six tasks use the benchmark's key-point observation mode. The step wrapper
concatenates the persistent step one-hot code to the base policy observation
and records the step code as available to both actor and critic. Dense rewards
are computed from task-language-compatible simulator APIs, including PyFlex
particle positions, cloth/rope metadata, cup poses, water particles, target
geometry, and geometric containment tests. SoftGym's official
\texttt{normalized\_performance} is excluded from reward and transition
decisions. For Qwen verification, the wrapper renders $320\!\times\!320$ RGB
frames and supplies a reset reference, recent frame ring, frozen current frame,
and the active-step instruction.

\begin{figure}[htbp]
      \centering
      \vspace{0.5em}
      \includegraphics[width=\linewidth]{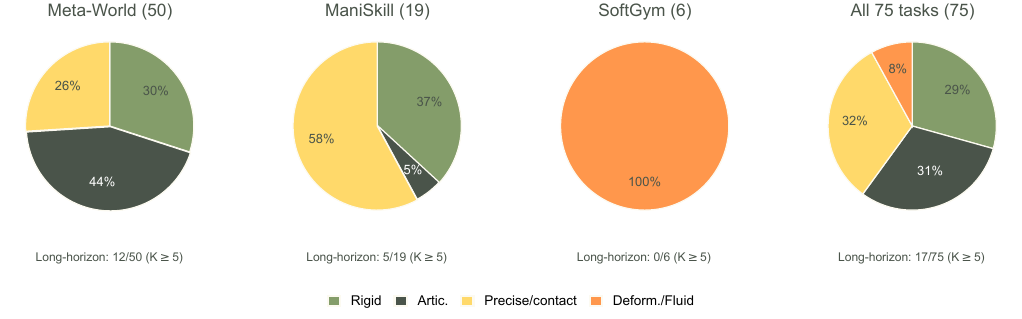}
      \caption{
      \textbf{Task-type distribution across the evaluated benchmarks.}
      Each pie shows the mutually exclusive semantic categories
      \textit{Rigid}, \textit{Artic.}, \textit{Precise/contact}, and
      \textit{Deform./Fluid}. The accompanying
      \textit{Long-horizon} annotation reports tasks with at least five steps
      ($K\geq5$) and may overlap with the four
      semantic categories. See more about task information in Table~\ref{tab:app-all-task-manifest} and~\ref{tab:app-related-work-task-type} and see task completion visualization in Figure~\ref{fig:benchmark_overview}.
      }
      \label{fig:task-type-distribution}
  \end{figure}

Across the full suite, we obtain the step statistics summarized in
Table~\ref{tab:step_statistics_summary}. The absorbing post-completion state is
excluded from $K$, so each task contains exactly $K-1$ ordered transitions. \texttt{StackPyramid-v1} is the only seven-step task. The
absorbing post-completion state is not counted in $K$, and therefore each task
contains exactly $K-1$ ordered transitions.

\begin{table}[htbp]
\centering
\small
\setlength{\tabcolsep}{5pt}
\renewcommand{\arraystretch}{1.0}
\begin{tabular}{lccccc}
\toprule
\textbf{Benchmark} &
\textbf{Tasks} &
\textbf{Steps} &
\textbf{Internal Trans.} &
\textbf{Mean Steps} &
\textbf{Step Range} \\
\midrule
Meta-World & 50 & 172 & 122 & 3.440 & 1--5 \\
ManiSkill  & 19 & 81  & 62  & 4.263 & 3--7 \\
SoftGym    & 6  & 12  & 6   & 2.000 & 2 \\
\midrule
\textbf{Total} & \textbf{75} & \textbf{265} & \textbf{190} &
\textbf{3.533} & \textbf{1--7} \\
\bottomrule
\end{tabular}
\caption{
\textbf{Step-decomposition statistics across the 75 benchmark tasks.}
Internal transitions count boundaries between consecutive non-terminal steps
and exclude transition into the terminal state.
}
\label{tab:step_statistics_summary}
\end{table}


\begin{figure*}[htbp]
\centering
\includegraphics[width=\textwidth]{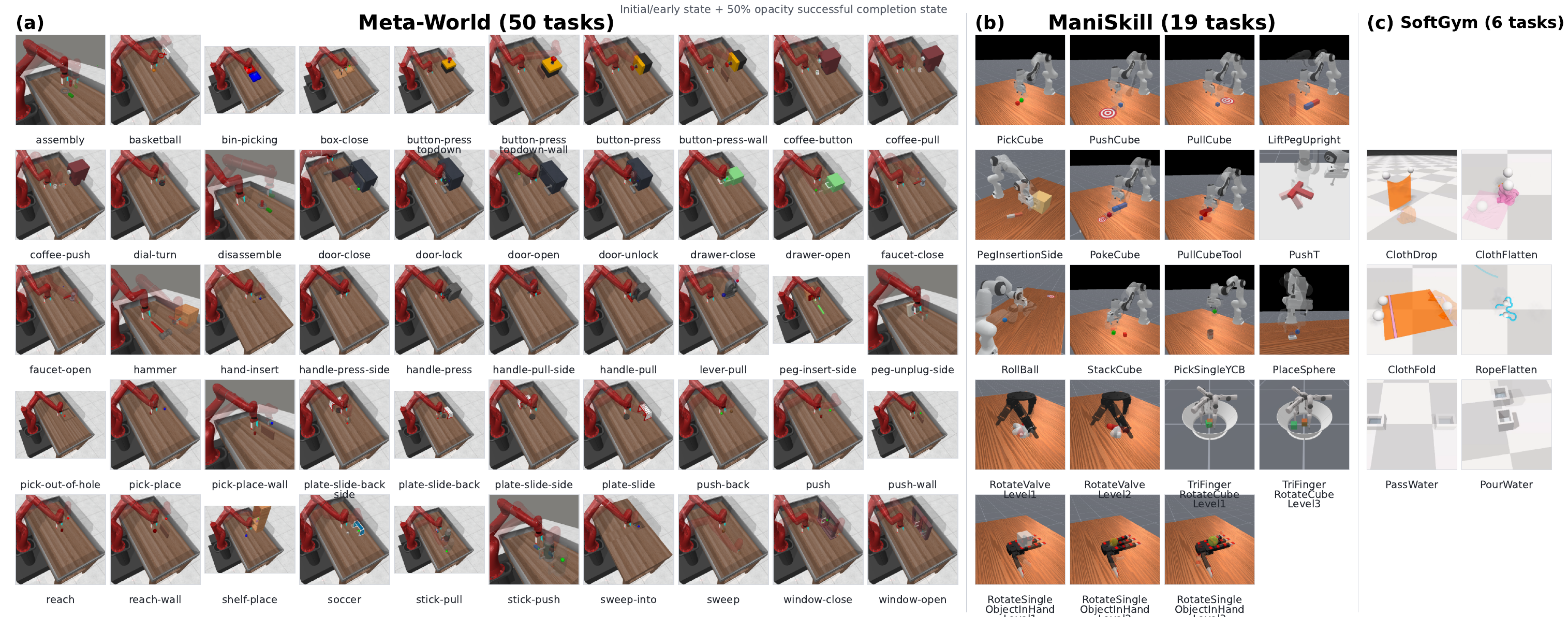}
\caption{
\textbf{Overview of benchmark tasks.}
Task scene demonstrations from MetaWorld, ManiSkill, and SoftGym.
Each overlaid visualization shows both the initial and final task states.
}
\label{fig:benchmark_overview}
\end{figure*}

\subsection{Additional Experimental Results}
\label{app:additional-results}

\subsubsection{Evaluation Protocols}
\label{app:full-results}

For the completed Meta-World \method{} (\GateRule{} and \GateVLM{}) runs,
checkpoints were ranked on independent reset seed 2042 and the selected
checkpoint was evaluated on held-out reset seed 1042.  All 50 tasks have
three independent \textbf{training seeds} (42, 43, and 44).

For each
completed ManiSkill held-out retest or new reward-control run, we select by
\texttt{success\_once} on 20 seed-2042 validation episodes, and test the selected checkpoint on 50
seed-1042 episodes. In SoftGym, the selected \GateRule{} runs use disjoint
validation and test configurations: all six tasks use normalized performance.

\subsubsection{Per-Task Learning Curves}
\label{app:learning-curves}

\begin{figure*}[htbp]
    \centering
    \includegraphics[width=0.9\textwidth]{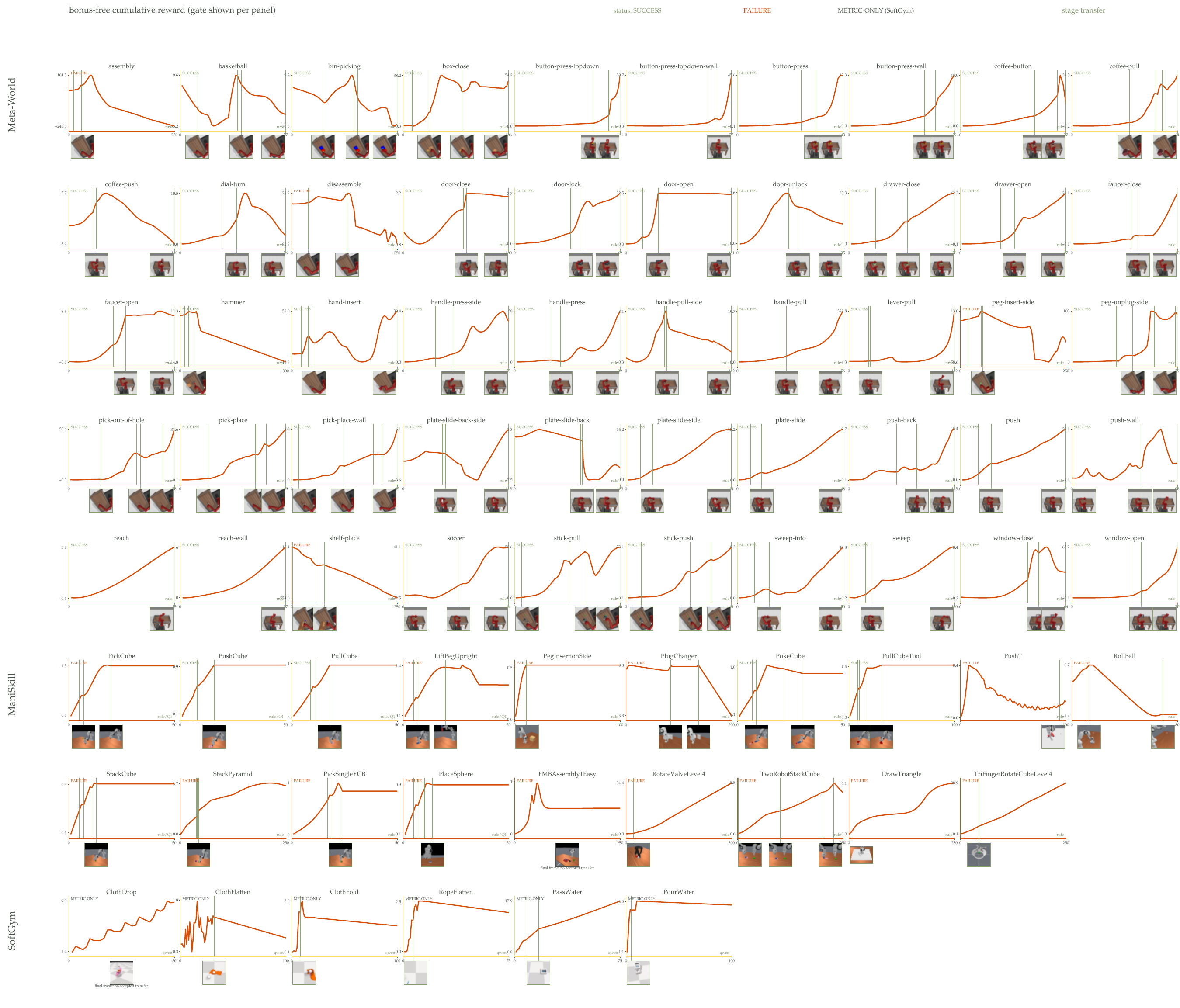}
    \caption{
    \textbf{Illustrative reward traces and step transitions.}
    We show one saved rollout per task for visualization. The examples include both successful and failed trajectories.
    Green and orange annotations indicate success and failure when the
    benchmark exposes a binary success signal, while SoftGym is marked
    metric-only because it has no shared binary success API.
    Vertical lines denote accepted step transitions.
    }
    \label{fig:illustrative-reward-curves}
\end{figure*}

To complement the aggregate results, we visualize representative reward traces over complete task rollouts in Figure~\ref{fig:illustrative-reward-curves}.
These examples illustrate how the learned reward evolves within and across steps under both successful and failed executions.

\begin{figure}[htbp]
\centering
\includegraphics[width=0.7\textwidth]{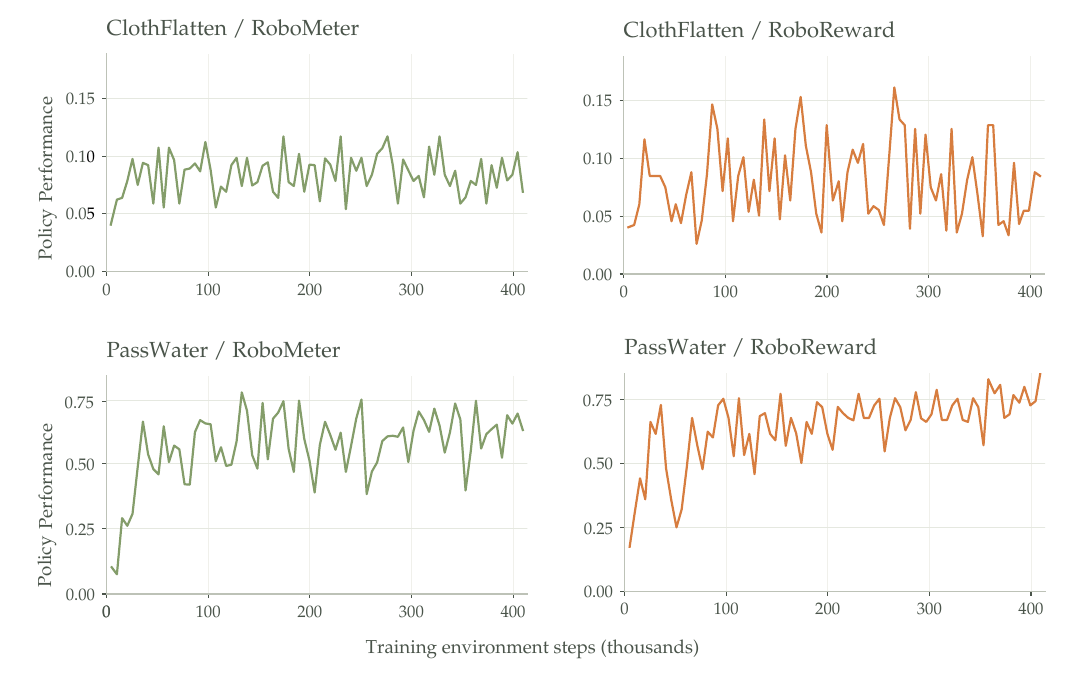}
\caption{Example policy learning curves of visual reward models on SoftGym.}
\label{fig:softgym_progress_learning}
\end{figure}

\paragraph{Learning curves for visual reward baselines.}
Figure~\ref{fig:softgym_progress_learning} provides reference
learning curves for policies trained with the RoboMeter and
RoboReward baselines on two SoftGym tasks. The curves show how
normalized task performance evolves over training environment
steps under each visual reward model, complementing the aggregate
baseline results with a view of policy learning dynamics.

\section{Visual Reward Model Experimental Protocols}
\label{app:baselines}

\subsection{Baseline Selection}
\label{app:baseline-select}

We select baselines that can provide an online visual reward signal under the
same downstream policy-learning interface. Not all related methods are directly
compatible with this setting. For example, RL-VLM-F~\citep{rlvlmf2024} is
tightly coupled to an off-policy PEBBLE/SAC loop and task-specific SoftGym
initialization; ReWiND~\citep{rewind2025} relies on demonstration-based
pretraining and an external success signal; while STDR~\citep{stdr2026} and
MARVL~\citep{marvl2026} require task-specific reward construction from expert
demonstrations or per-stage target images, respectively.

For reproduced visual-reward baselines, we regard the released VLM checkpoint,
prompt, visual input construction, score normalization, and temporal reward
adapter as part of the baseline itself. We therefore follow the backbone and
reward interface released or reported by each method rather than replacing all
baselines with a common VLM.

\begin{table}[htbp]
\centering
\caption{\textbf{Original task coverage of reference-only baselines.}
We list the tasks reported in the original papers and categorize them using
the same primary interaction taxonomy as Table~\ref{tab:q2-main}.}
\label{tab:app-reference-tasks}

\scriptsize
\setlength{\tabcolsep}{3.2pt}
\renewcommand{\arraystretch}{0.92}

\begin{tabularx}{\linewidth}{@{}L{0.95in}cXL{1.25in}@{}}
\toprule
\textbf{Method}
& \textbf{\# Tasks}
& \textbf{Original Tasks}
& \textbf{Type Mix} \\
\midrule

RL-VLM-F~\citep{rlvlmf2024}
& 7
&
\textbf{Meta-World:}
Open Drawer, Soccer, Sweep Into. \newline
\textbf{SoftGym:}
Fold Cloth, Straighten Rope, Pass Water. \newline
\textbf{OpenAI Gym:}
CartPole.
&
Rigid 2; Artic. 1; Deform. 2; Fluid 1;
non-manip. 1.
\\

\midrule

ReWiND~\citep{rewind2025}
& 13
&
\textbf{Meta-World:}
Window Close, Reach Wall, Faucet Close, Coffee Button,
Button Press Wall, Door Lock, Handle Press Side, Sweep Into. \newline
\textbf{Real:}
Separate blue/orange cups, Fold blue towel, Open red trash bin,
Put orange cup on red plate, Put fruit-colored object in box.
&
Rigid 5; Artic. 5; Precise 2; Deform. 1.
\\

\midrule

STDR~\citep{stdr2026}
& 17
&
\textbf{Meta-World:}
Button Press, Window Open, Reach Wall, Plate Slide,
Push, Coffee Push, Stick Push, Pick Place. \newline
\textbf{ManiSkill:}
PullCube, PushCube, PickCube. \newline
\textbf{Franka Kitchen:}
Slide Cabinet, Kettle, Microwave. \newline
\textbf{Real:}
PutPepperInBucket, PutToyOnPlate, ReachAndPullDrawer.
&
Rigid 11; Artic. 4; Precise 1; Tool 1.
\\

\bottomrule
\end{tabularx}
\end{table}

\begin{table}[htbp]
\centering
\caption{\textbf{Original evaluation coverage of reference-only baselines.}
We categorize the policy-learning tasks reported in the original papers using
the same primary interaction taxonomy as Table~\ref{tab:q2-main}. Counts exclude
real-world reward-only evaluation and non-manipulation tasks.}
\label{tab:reference-baseline-coverage}

\scriptsize
\setlength{\tabcolsep}{2.4pt}
\renewcommand{\arraystretch}{0.92}

\begin{tabularx}{\linewidth}{@{}L{1.05in}ccccccccL{1.25in}@{}}
\toprule
\textbf{Method}
& \textbf{Meta-World}
&
\textbf{ManiSkill}
&
\textbf{SoftGym}
& \textbf{Rigid}
& \textbf{Artic.}
& \textbf{Precise}
& \textbf{Tool}
& \textbf{Deform./Fluid}
& \textbf{Original reporting}
\\
\midrule

RL-VLM-F~\citep{rlvlmf2024}
& 3 & 0 & 3
& 2 & 1 & 0 & 0 & 2 / 1
& MW: success rate; others: episode return \\

ReWiND~\citep{rewind2025}
& 8 & 0 & 0
& 2 & 4 & 2 & 0 & 0 / 0
& 79\% IQM SR at 100k steps \\

STDR~\citep{stdr2026}
& 8 & 3 & 0
& 9 & 3 & 1 & 1 & 0 / 0
& Per-task learning curves; MS table reports SR \\

\bottomrule
\end{tabularx}
\end{table}

\subsection{Baseline Implementations}
\label{app:baseline-implementations}

\paragraph{Evaluation Protocol.}
Within each benchmark and task, all reward methods use the same environment
version, reset distribution, policy observation, action and controller
interface, actor--critic architecture, PPO configuration, rollout budget,
episode horizon, training seeds, and checkpoint cadence. Only the reward source
and the visual inputs required by that reward method are changed.

Checkpoints are selected using an independent validation reset set and evaluated
on a held-out reset set using the benchmark-native evaluation metric. For
Meta-World and ManiSkill, we use 20 validation episodes with reset seed 2042
and 50 held-out episodes with reset seed 1042. SoftGym uses disjoint validation
and test configurations and reports its native normalized-performance metric.
Unless otherwise stated, results are aggregated over three independent training
seeds. Missing or incomplete evaluations are reported as unavailable rather
than being converted to zero.

\paragraph{Matched Comparison with Visual Reward Baselines.}
Within each task, we keep the environment version, reset distribution,
action and controller interface, actor--critic architecture, PPO optimization
protocol, rollout budget, random seeds, checkpoint-selection rule, and
evaluation protocol fixed across methods. The intentional differences are
restricted to the reward formulation and any additional input required by a
particular method. Apart from these method-specific inputs, all policy
observations follow the standard observation setting of the corresponding
benchmark.

For global-progress baselines, we follow their released or reported reward
protocols as closely as possible, including their prescribed VLM backbone and
additional model inputs. We treat the VLM choice as part of each reward
method rather than forcing all methods to share a single backbone. In
\method{}, the reward components are constructed offline by \FrameVLM{}
following the shared core formulation, while \GateVLM{} is queried online only
for semantic step-transition verification, rather than for direct prediction
of a global scalar progress reward.

\paragraph{Baseline Reward Interfaces.}
The benchmark-native dense control directly uses the benchmark reward, while
the terminal-only and per-subtask sparse controls retain only terminal or
subtask-completion events, respectively.

For the visual baselines, we preserve their method-specific temporal reward
interfaces. SOLE-R1 uses its released absolute-progress predictor on sparse
visual observations; in the main comparison we use the same fixed
$q=64$, reward scale $0.05$, and no progress clipping across tasks, and
linearly interpolate sparse predictions over the rollout before PPO updates.
RoboMeter uses its reward head on a bounded video history queried every
10 environment steps and holds the latest absolute progress estimate between
queries. RoboReward similarly uses its discrete $1$--$5$ trajectory score,
mapped to $[0,1]$ by $(s-1)/4$, with a $q=10$ video buffer and reward holding.
TOPReward obtains a scalar completion score from VLM token probabilities on
cumulative visual evidence and interpolates sparse $q=10$ predictions over the
rollout. LRM-Progress uses the released absolute-progress endpoint with
$q=10$ and holds the most recent prediction between queries. Visual reward
inputs are never concatenated with the policy observation.

\paragraph{Inference-Side Calibration.}
Visual progress methods expose additional inference-side choices such as query
interval, video history, score scaling or clipping, and interpolation or holding
between sparse predictions. We use released settings whenever available and
otherwise restrict selection to these reward-interface parameters using
validation episodes only; policy architecture and PPO optimization remain
unchanged.

SOLE-R1 is particularly sensitive to this temporal adapter. We therefore also
evaluate task-specific query/scale/clip candidates as a separate diagnostic.
The main Meta-World comparison uses the single frozen
$q=64$ configuration above.

\subsection{Training Cost and Reward-Query Efficiency}
\label{app:training-cost}

\FrameVLM{} is invoked offline during task compilation and therefore introduces
no reward-model inference in the policy rollout. During online learning,
\GateVLM{} is queried only when the transition trigger requests semantic
verification; the verified step state is then retained between queries.

In contrast, global-progress baselines repeatedly rescore visual observations
or video prefixes throughout the rollout. For a fixed-cadence method with
episode horizon $H$ and query interval $q$, the number of reward-model query
events scales approximately as $\lceil H/q\rceil$ per environment episode
before batching. In our main implementations, SOLE-R1 uses $q=64$, while
LRM-Progress, RoboMeter, RoboReward, and TOPReward use $q=10$ according to
their corresponding reward adapters. We preserve these method-specific
cadences rather than artificially matching query counts, since changing the
temporal adapter changes the reward definition itself. All supported visual
queries are batched, and offline \FrameVLM{} compilation is accounted for
separately from online reward-query cost.


\section{Real-to-Sim-to-Real Experiment \ Details}
\label{app:r2s2r}

\subsection{Policy Inputs and Deployment Protocol}
\label{app:r2s2r-inputs}

The generic actor--critic interface is defined in
Section~\ref{sec:method} and Figure~\ref{fig:method-overview}. Here we
specify its Real-to-Sim-to-Real instantiation. The actor receives
deployment-compatible camera-derived perception features, robot
proprioception and controller state, and the persistent step one-hot
$e(z_t)$. The critic additionally receives simulator-only task, contact, and
safety states during PPO training and is discarded at deployment.

For the actor-blind control, we remove only $e(z_t)$ from the actor input:
\begin{equation}
x_{t,\mathrm{w/o}}^{\mathrm{actor}}
=
\left[
f_{\mathrm{vis}}(I_t);
x_t^{\mathrm{robot}}
\right].
\label{eq:r2s2r-actor-without-step}
\end{equation}
The reward program, transition mechanism, critic input, PPO architecture,
training budget, and checkpoint-selection protocol remain unchanged. At
deployment, \GateVLM{} is queried periodically to update $z_t$, and the
resulting step context is supplied only to the step-conditioned actor.

\subsection{Scene Reconstruction and Calibration}
\label{app:r2s2r-scenes}

We reconstruct task-specific digital twins by aligning the robot base,
objects, articulated assets, external-camera calibration, and initial robot
configuration with the corresponding real scenes. Policy training and
checkpoint selection use only the reconstructed simulation. \textbf{No real-robot
demonstrations, pretrained policy or real-world policy updates are used}.

The simulator we include is NVIDIA Isaac Sim with
PhysX. Physics runs at $120$\,Hz, while the policy runs at $10$\,Hz with
12 physics substeps per action. The robot is a UR3 with a Robotiq gripper,
and the four-dimensional action consists of normalized Cartesian
end-effector displacement and an absolute gripper command. End-effector
orientation is fixed by the controller rather than directly predicted by the
actor. The external camera, robot base, and laptop articulation are calibrated
in the reconstructed scene.

The actor uses deployment-compatible robot and controller
state together with a camera-derived screen-handle displacement estimate.
During simulation training, this estimate is perturbed with $2$\,mm visual
position noise. Simulator-only hinge, contact, safety, and success quantities
are restricted to the critic, reward computation, and evaluation.

\begin{figure}[htbp]
    \centering
    \includegraphics[width=0.6\linewidth]{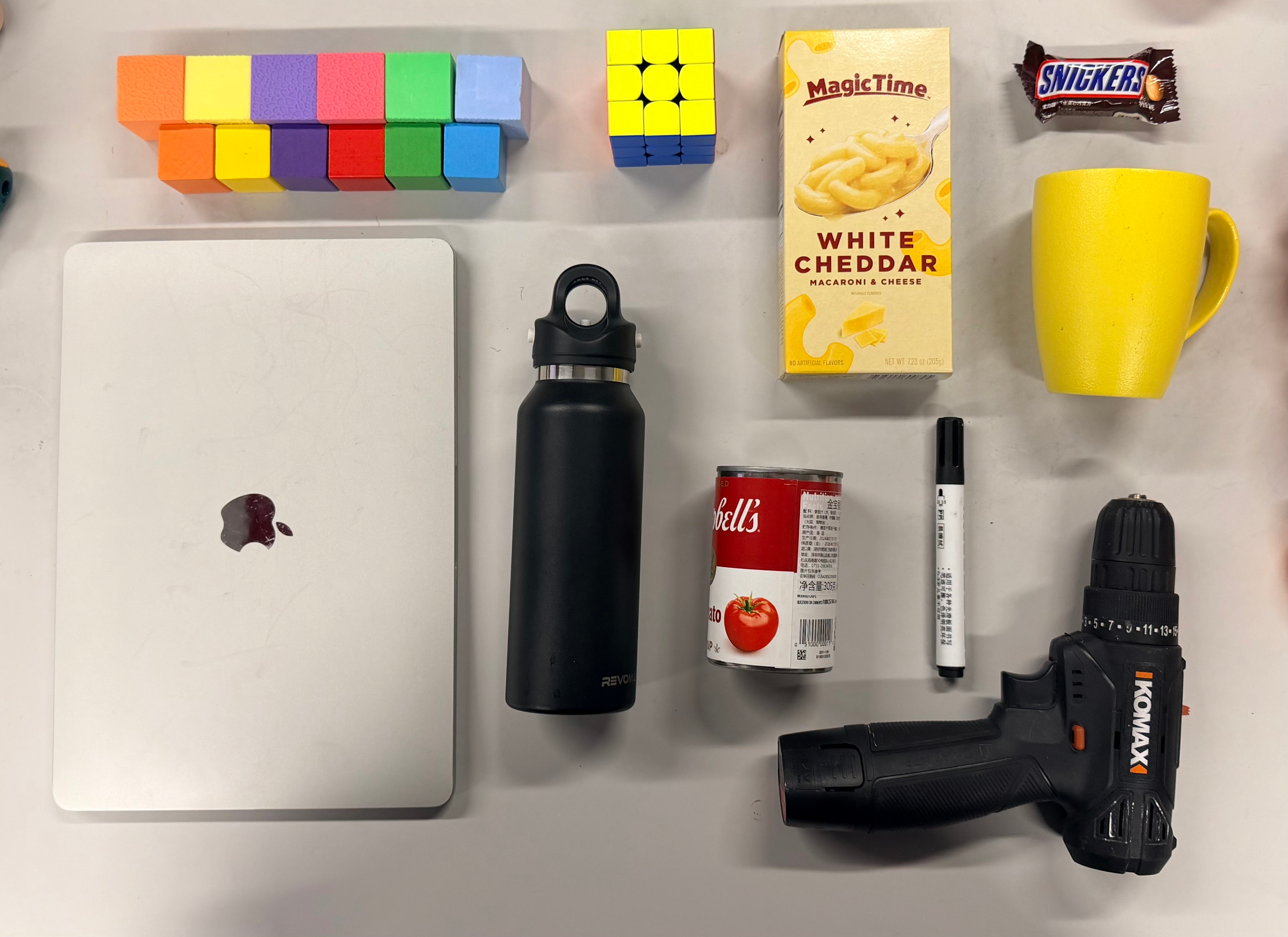}
    \vspace{-0.3em}
    \caption{\textbf{Objects used in our sim-to-real transfer experiments.}}
    \label{fig:real-world-task-objects}
\end{figure}

\subsection{Task and Success Criteria}
\label{app:r2s2r-success}

We evaluate sim-to-real transfer across four real-world manipulation tasks spanning articulated, precise, and rigid-object interactions.
Table~\ref{tab:real-tasks-transfer} summarizes the task definitions and corresponding success criteria used for evaluation.

\begin{table*}[htbp]
\centering
\vspace{-0.5em}

\caption{
\textbf{Real-world task setups for sim-to-real experiments.}
Detailed task definitions and success criteria are provided in Appendix~\ref{app:r2s2r-scenes}.
}
\label{tab:real-tasks-transfer}

\footnotesize
\setlength{\tabcolsep}{6pt}
\renewcommand{\arraystretch}{1.05}

\begin{tabularx}{0.96\textwidth}{
    @{}
    >{\centering\arraybackslash}p{0.07\textwidth}
    >{\centering\arraybackslash}p{0.10\textwidth}
    >{\raggedright\arraybackslash}p{0.34\textwidth}
    >{\raggedright\arraybackslash}X
    @{}
}
\toprule

\textbf{Task}
& \textbf{Type}
& \textbf{Task description}
& \textbf{Success criterion}
\\

\midrule

1.1
& Articulate
& Close the computer.
& The laptop hinge angle is below $10^\circ$.
\\

\midrule

1.2
& Precise
& Place the pen in the cup.
& The longitudinal axis of the pen lies inside the cup.
\\

\midrule

2.1
& Rigid
& Put the cube on the right into the cup.
& The cube is inside the cup.
\\

\midrule

2.2
& Rigid
& Put the rubik near the cup.
& The center of the Rubik's Cube bottom face is within $5\,\mathrm{cm}$ of the center of the cup bottom face.
\\

\bottomrule
\end{tabularx}

\vspace{-0.5em}
\end{table*}

\subsection{Training and Evaluation Protocol}
\label{app:r2s2r-evaluation}

For each task, the step-conditioned and actor-blind variants use matched
actor--critic architectures, reward programs, training budgets, and
checkpoint-selection rules. Policies are trained only in reconstructed
simulation. No demonstration data, real-world reward, or real-world gradient
update is used.

The selected checkpoint for each policy is first evaluated on 100 simulation
episodes. We report its empirical success percentage directly:
\begin{equation}
\widehat{\mathrm{SR}}_{\mathrm{sim}}
=
\frac{N_{\mathrm{success}}}{100}\times100\%.
\label{eq:r2s2r-sim-sr}
\end{equation}
Real-world evaluation contains three blocks of ten trials per task and policy
setting. Let $s_j\in\{0,\ldots,10\}$ be the successful trials in block
$j\in\{1,2,3\}$ and let $r_j=10s_j$ be the corresponding percentage. We
report
\begin{equation}
\overline r
=
\frac{1}{3}
\sum_{j=1}^{3}r_j,
\qquad
\operatorname{StdErr}
=
\frac{1}{\sqrt{3}}
\sqrt{
\frac{
\sum_{j=1}^{3}
(r_j-\overline r)^2
}{
2
}
}.
\label{eq:r2s2r-real-stderr}
\end{equation}
The reported StdErr describes
variation across the three real-world evaluation blocks.

\subsection{Results}
\label{app:r2s2r-time}

Explicit actor step context improves average success by $11.5$ percentage
points in reconstructed simulation and by $8.3$ points in the real world,
with consistent gains across all four tasks. Task~1.2 remains the most
challenging, and the actor without step context fails in all 30 real-world
trials. Both variants exhibit a clear sim-to-real drop, indicating that
explicit task progress improves transfer but does not eliminate errors from
perception, calibration, and physical dynamics. All policies are trained only in reconstructed simulation. Simulation
results are empirical success rates over 100 episodes for a single selected
checkpoint.

\begin{table}[htbp]
\centering
\setlength{\belowcaptionskip}{2pt}
\caption{
\textbf{Simulation and real-world success rates (\%) for direct sim-to-real
transfer.} All policies are trained only in reconstructed simulation.
Real-world values are reported as mean $\pm$ StdErr over three 10-trial runs.
}
\label{tab:q3-real-transfer}

\scriptsize
\setlength{\tabcolsep}{2.8pt}
\renewcommand{\arraystretch}{0.95}

\begin{tabularx}{\linewidth}{
    @{}
    c
    *{4}{>{\centering\arraybackslash}X}
    @{}
}
\toprule
&
\multicolumn{2}{c}{\textbf{Reconstructed Simulation}}
&
\multicolumn{2}{c}{\textbf{Real-World Deployment}}
\\
\cmidrule(lr){2-3}
\cmidrule(lr){4-5}

\textbf{Task}
&
\makecell{\textbf{w/ Actor}\\\textbf{Step Index}}
&
\makecell{\textbf{w/o Actor}\\\textbf{Step Index}}
&
\makecell{\textbf{w/ Actor}\\\textbf{Step Index}}
&
\makecell{\textbf{w/o Actor}\\\textbf{Step Index}}
\\
\midrule

1.1
& \textbf{45.0}
& 33.0
& \textbf{33.3 $\pm$ 3.3}
& 26.7 $\pm$ 3.3
\\

1.2
& \textbf{25.0}
& 13.0
& \textbf{13.3 $\pm$ 3.3}
& 0.0 $\pm$ 0.0
\\

\midrule

2.1
& \textbf{36.0}
& 25.0
& \textbf{23.3 $\pm$ 3.3}
& 16.7 $\pm$ 3.3
\\

2.2
& \textbf{41.0}
& 30.0
& \textbf{30.0 $\pm$ 5.8}
& 23.3 $\pm$ 3.3
\\

\midrule

\textbf{Average}
& \textbf{36.8}
& 25.3
& \textbf{25.0 $\pm$ 3.8}
& 16.7 $\pm$ 2.2
\\

\bottomrule
\end{tabularx}
\end{table}
\section{Bidirectional Task Progress}
\label{app:bi-task-progress}

We evaluate whether the persistent step state learned through clean stepwise
training supports two complementary capabilities: forward composition across
increasingly long task chains, and backward correction after a previously
completed prerequisite becomes invalid. The former trains a separate policy
for each composition length; the latter freezes a clean-trained policy and
changes only its step-index input after an unseen regression. 

\subsection{Forward: Composable Stepwise Rewards}
\label{app:exp-composition}

All composition experiments use the privileged \GateRule{} for atomic-step
and macro-boundary transitions, isolating compositional policy learning factors from
visual-verification errors.

\begin{figure}[htbp]
    \centering
    \includegraphics[width=\linewidth]{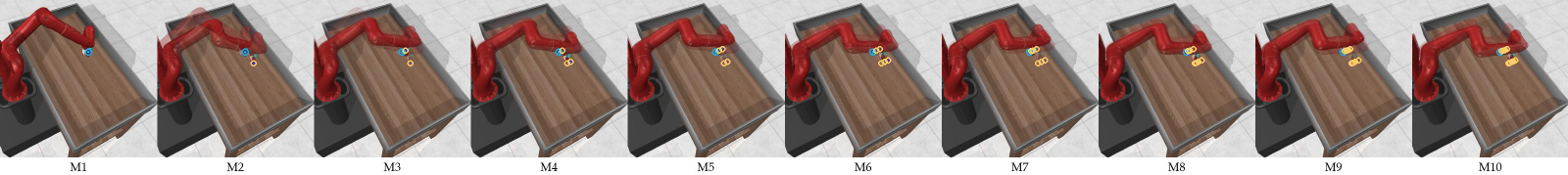}
    \vspace{-0.5em}
    \caption{
    \textbf{Homogeneous composition protocol.}
    $\mathrm{Homo}_n$ moves one object through $n$ distinct targets without
    resetting the robot, object, or scene. Each macro-subtask contains three
    atomic \method{} steps. Panels show $n=1,\ldots,10$.
    }
    \label{fig:compositional-homo}
\end{figure}

\subsubsection{Experiment Setting Formulation}
\label{app:exp-composition-formulation}

\paragraph{Homogeneous composition.}
We use $n$ to denote the number of macro-subtasks and $K$ the number of
atomic \method{} steps. One macro-subtask transfers the object to one target
through
\begin{equation}
P_j
=
[
\text{pregrasp}
\rightarrow
\text{capture and lift}
\rightarrow
\text{place and release}
].
\label{eq:homo-macro}
\end{equation}

A homogeneous composition is

\begin{equation}
\mathrm{Homo}_n
=
P_1\circ P_2\circ\cdots\circ P_n,
\qquad
K=3n.
\label{eq:homo-composition}
\end{equation}
Every $P_j$ uses the same semantic primitive, reward template, and reward
coefficients, but moves the object to a distinct target $g_j$. The complete
reward program is formed by ordered concatenation:
\begin{equation}
\mathcal{R}(\mathrm{Homo}_n)
=
\mathcal{R}(P_1)
\oplus
\cdots
\oplus
\mathcal{R}(P_n),
\label{eq:homo-reward-program}
\end{equation}
where $\oplus$ denotes sequential composition. Appending $P_{n+1}$ does not modify the reward functions or
coefficients of any existing prefix macro tasks.

For a non-terminal global step index $z_t\in\{0,\ldots,3n-1\}$, the active
macro and its local atomic step are
\begin{equation}
m_t
=
1+\left\lfloor\frac{z_t}{3}\right\rfloor,
\qquad
k_t
=
z_t\bmod 3.
\label{eq:macro-local-index}
\end{equation}
Thus, $z_t$ simultaneously selects the active target $g_{m_t}$, the local
three-step reward component indexed by $k_t$, and the one-hot context supplied
to the policy.

\paragraph{Target sequence and physical continuity.}
The experiment uses one Meta-World \texttt{pick-place-v3} scene, one robot,
and one object. In the distinct-target protocol, the target sequence is
\begin{equation}
\begin{aligned}
(g_1,\ldots,g_{10})
=\bigl[
&(-0.08,0.80),\ (0.08,0.80),\
(-0.08,0.84),\ (0.08,0.84),\\
&(-0.08,0.88),\ (0.08,0.88),\
(-0.08,0.82),\ (0.08,0.82),\\
&(-0.08,0.86),\ (0.08,0.86)
\bigr],
\end{aligned}
\label{eq:homo-target-sequence}
\end{equation}
where the coordinates specify the target position in the tabletop plane and
the height follows the benchmark configuration. \textbf{No target repeats within the
reported chain.}

\textbf{The robot, object, and scene are not reset between macros.} After $P_j$
finishes, only the active target and persistent step state are updated:
\begin{equation}
\xi_{\mathrm{start}}^{P_{j+1}}
=
\xi_{\mathrm{terminal}}^{P_j},
\label{eq:continuous-handoff}
\end{equation}
where $\xi$ denotes the full physical simulator configuration. Consequently,
arm pose, placement error, and accumulated control error are preserved across
macro boundaries. A full reset occurs only at the beginning of a new episode.

\paragraph{Fixed-length heterogeneous composition.}
We additionally vary semantic composition while fixing $n=3$ and $K=9$.
\textbf{All modules are instantiated within a single Meta-World
\texttt{pick-place-v3} MuJoCo scene}. The same-scene macro modules are
\begin{equation}
\mathsf{A}
=
\text{pick-and-place},
\qquad
\mathsf{B}
=
\text{object push},
\qquad
\mathsf{C}
=
\text{3 step end-effector waypoint reaching}.
\label{eq:heterogeneous-modules}
\end{equation}
Module $\mathsf{C}$ requires the TCP to reach three successive Cartesian
waypoints, one per atomic step, without object interaction. We compare
\begin{equation}
\mathrm{Homo}_3
=
\mathsf{AAA},
\qquad
\mathrm{Hetero}_3(\mathsf{AAB}),
\qquad
\mathrm{Hetero}_3(\mathsf{ABC}).
\label{eq:heterogeneous-settings}
\end{equation}
All three settings share the same Meta-World scene, four-dimensional action
interface, observation space, policy architecture, step-index width,
$2{,}100$ PPO iterations, and $750$-step episode horizon. The physical state
remains continuous across macro boundaries.

\subsubsection{Training and Evaluation}

\paragraph{Policy input and architecture.}
The compositional wrapper preserves the native Meta-World observation
$o_t^{\mathrm{MW}}\in\mathbb{R}^{39}$ and four-dimensional continuous action
interface. The stepwise actor and critic receive
\begin{equation}
o_t^{\mathrm{actor}}
=
o_t^{\mathrm{critic}}
=
\left[
o_t^{\mathrm{MW}};
\mathbf{e}_{K_{\max}}(z_t)
\right],
\qquad
a_t
=
\left[
\Delta x_t,\Delta y_t,\Delta z_t,g_t
\right]
\in[-1,1]^4.
\label{eq:composition-policy-input}
\end{equation}
The action is the standard four-dimensional Meta-World control vector,
consisting of normalized end-effector Cartesian displacement and gripper
control. \textbf{A single
actor--critic policy controls all macro-subtasks throughout an episode with no
macro-specific policy or skill instantiated}.

The actor and critic are MLPs with ELU activations and hidden widths
$(256,256,128)$. The continuous state features are running-normalized, while
the global step one-hot remains unnormalized. Its width covers all atomic
steps and the terminal state. Within each matched length comparison, the
shorter and longer policies use the same one-hot width.

\paragraph{Training protocol.}
The primary protocol trains a separate policy for each composition length.
\textbf{The nominal training budget and episode horizon scale linearly with $n$}:
\begin{equation}
B(n)=700n
\quad\text{PPO iterations},
\qquad
H_{\mathrm{ep}}(n)=250n
\quad\text{environment steps}.
\label{eq:composition-budget}
\end{equation}
Training uses 128 parallel environments and 64 rollout steps per environment
and PPO iteration. PPO hyperparameters and actor--critic hidden architectures
are otherwise fixed.

The global step one-hot has enough coordinates for every atomic step and the
terminal state. Within the length comparison, the shorter and
longer policies use the same input width.

\paragraph{Strict evaluation criterion.}
Policies are evaluated deterministically on held-out environment seeds. \textbf{An
episode succeeds only when all $3n$ gated atomic steps are completed}:
\begin{equation}
\mathrm{SR}(\mathrm{Homo}_n)
=
\Pr
\left(
z_t=3n
\text{ for some }t\leq H_{\mathrm{ep}}(n)
\right).
\label{eq:homo-success}
\end{equation}
Merely placing the object near the final target without completing the
prescribed gated sequence does not count as success. Benchmark-native target
success is retained only as an auxiliary diagnostic.

For each trained policy, numerators and denominators are pooled across the
predeclared held-out environment seeds before computing SR. The table reports the actual number of completed episodes.

\paragraph{Non-stepwise controlled baseline.}
To test the importance of the complete stepwise interface, we jointly remove
active-step reward routing, the policy step-index input, and transition
bonuses. For the currently active macro, the control sums all three local
dense terms,
\begin{equation}
\widetilde r_t
=
\sum_{k=0}^{2}
D_k
\left(
o_t,o_{t-1};g_{m_t}
\right),
\qquad
\widetilde b_t=0,
\label{eq:joint-nonstepwise-control}
\end{equation}
while the actor and critic receive only the native benchmark observation.
The internal rule tracker is retained solely to \textbf{switch the sequential target
and apply the same strict evaluation criterion}. 

\subsubsection{Experiment Results and Conclusions}

\begin{table*}[htbp]
\centering
\caption{\textbf{Forward-composition result ledger.}}
\label{tab:forward-composition-full}

\scriptsize
\setlength{\tabcolsep}{2.5pt}
\renewcommand{\arraystretch}{0.92}

\resizebox{\textwidth}{!}{%
\begin{tabular}{@{}l*{10}{c}@{}}
\toprule
\textbf{Composition}
&
Homo$_{1}$
&
Homo$_{2}$
&
Homo$_{3}$
&
Homo$_{4}$
&
Homo$_{5}$
&
Homo$_{6}$
&
Homo$_{7}$
&
Homo$_{8}$
&
Homo$_{9}$
&
Homo$_{10}$
\\
\midrule

\textbf{$K$}
& 3 & 6 & 9 & 12 & 15 & 18 & 21 & 24 & 27 & 30
\\

\textbf{Eval. Episodes}
& 1024 & 1024 & 1024 & 1024 & 1024
& 1024 & 1024 & 1024 & 1024 & 1024
\\

\textbf{SR}
& 100.00\%
& 100.00\%
& 99.23\%
& 100.00\%
& 100.00\%
& 98.85\%
& 97.28\%
& 100.00\%
& 93.99\%
& 94.14\%
\\

\bottomrule
\end{tabular}%
}
\end{table*}

The scratch policies remain near ceiling through Homo$_8$, with no
reproducible monotonic degradation in the completed scratch results.
Continuation evaluations show that policies can also complete
Homo$_{10}$ with $K=30$ atomic steps. 

Completed evaluations currently extend to Homo$_{10}$ because of
the available training-time budget and reporting cutoff, \textbf{rather than an
observed failure at the next length}. Future experiments will continue to
search for a reproducible composition breakpoint under matched scratch
training, input width, target geometry, and training seeds.

The non-stepwise control in Equation~\ref{eq:joint-nonstepwise-control} reaches 256/256 strict success on $\mathrm{Homo}_1$, but obtains 0/256 on Homo$_{2}$ and the following compositional length.
Thus, the tested configuration without reward routing, step-index context,
or transition bonuses \textbf{fails on every evaluated composition beyond one macro}.

For the fixed-length semantic heterogenous compositional experiment as reported in Equaton~\ref{eq:heterogeneous-modules}, mean SR over three independently
trained policies is
\begin{equation}
\begin{aligned}
\mathrm{Homo}_3(\mathsf{AAA})
&: 99.23,\\
\mathrm{Hetero}_3(\mathsf{AAB})
&: 85.83 \pm 5.02,\\
\mathrm{Hetero}_3(\mathsf{ABC})
&: 73.33 \pm 5.74.
\end{aligned}
\label{eq:heterogeneous-results}
\end{equation}
The results indicate that semantic diversity in task composition increases
difficulty: replacing repeated instances of the same subtask with heterogeneous
combinations reduces success from $\mathrm{Homo}_3(\mathsf{AAA})$ to
$\mathrm{Hetero}_3(\mathsf{AAB})$ and further to
$\mathrm{Hetero}_3(\mathsf{ABC})$. Nevertheless, the heterogeneous compositions
still achieve substantial success rates, suggesting that the explicit stepwise
reward structure remains effective \textbf{even when successive steps correspond to
different manipulation semantics}.

\subsection{Backward: Regression and Recovery}
\label{app:exp-recovery}

\subsubsection{Experiment Setting Formulation}
\label{app:exp-regression-formulation}

\paragraph{Task source.} We evaluate four tasks from ManiSkill~3:
\texttt{PickCube-v1},
\texttt{PushCube-v1},
\texttt{PullCube-v1}, and
\texttt{PickSingleYCB-v1}. For each task $\tau$, we use an independently trained PPO actor--critic policy learned only from clean forward execution. Following the main policy
interface, the actor and critic receive the native ManiSkill state observation
together with the persistent step one-hot:
\begin{equation}
a_t
\sim
\pi_{\theta_\tau}
\left(
\cdot
\mid
o_t^{(\mathrm{MS})},
e_\tau(z_t)
\right),
\qquad
V_{\psi_\tau}
=
V_{\psi_\tau}
\left(
o_t^{(\mathrm{MS})},
e_\tau(z_t)
\right),
\label{eq:regression-policy-interface}
\end{equation}
where
\begin{equation}
e_\tau(z_t)
\in
\{0,1\}^{K_\tau+1}
\end{equation}
encodes the persistent step state, including the terminal state. Both actor
and critic use three hidden layers of width 256 with Tanh activations. The
actor outputs the native continuous action consumed by the ManiSkill
\texttt{pd\_joint\_delta\_pos} controller, while the critic outputs a scalar
value.

During recovery evaluation,
\begin{equation}
\theta_{\tau,t}
=
\theta_{\tau,\mathrm{clean}}
\qquad
\forall t,
\label{eq:frozen-recovery-policy}
\end{equation}
so the actor receives no gradient update, online finetuning, disturbance
curriculum, recovery demonstration, or recovery-specific reward. Within each
task, the same checkpoint is used before and after the disturbance and for
all compared rollback mechanisms.

\paragraph{Controlled regressions and environment randomization.}
Each episode begins from the native ManiSkill reset distribution, which
randomizes the initial scene configuration, object pose, and task goal.
For \texttt{PickSingleYCB-v1}, the manipulated YCB object and its geometry
also vary across reconfigurations.

The disturbance itself is controlled rather than randomly sampled. For each
task, its type, magnitude, and trigger phase are fixed. The frozen actor first
reaches a designated downstream step through normal execution. Two policy
steps later, at time $t_d$, we apply a \textbf{task-specific disturbance operator}
$\Gamma_\tau$:
\begin{equation}
s_{t_d^{+}}^{\mathrm{sim}}
=
\Gamma_\tau
\left(
s_{t_d^{-}}^{\mathrm{sim}}
\right),
\label{eq:controlled-regression}
\end{equation}
where $s_t^{\mathrm{sim}}$ denotes simulator state used only to construct and
evaluate the controlled regression. The actor and \GateVLM{} are not told
that the intervention occurred.

\begin{table}[htbp]
\centering
\caption{
\textbf{Controlled regression settings.}
The disturbance and evaluator-only recovery step are not provided to
\GateVLM{}.
}
\label{tab:regression-interventions}

\small
\setlength{\tabcolsep}{3.5pt}
\renewcommand{\arraystretch}{0.95}

\begin{tabularx}{\linewidth}{@{}l l X l@{}}
\toprule
\textbf{Task}
&
\textbf{Family}
&
\textbf{Controlled regression}
&
\textbf{RGB reference}
\\
\midrule

\texttt{PickCube-v1}
&
Pick-and-place
&
The grasped cube is returned to the table near the gripper, approximately
$0.05$\,m from its pre-disturbance position, invalidating grasp-and-carry.
&
Step entry
\\

\texttt{PushCube-v1}
&
Nonprehensile push
&
The visible target is displaced by $0.16$\,m, invalidating the previously
satisfied object--target relation.
&
Step entry
\\

\texttt{PullCube-v1}
&
Nonprehensile pull
&
The visible target is displaced by $0.16$\,m, invalidating the previously
satisfied object--target relation.
&
Step entry
\\

\texttt{PickSingleYCB-v1}
&
Object generalization
&
The grasped YCB object is returned to the table near the gripper,
approximately $0.04$\,m from its pre-disturbance position.
&
Refreshed
\\

\bottomrule
\end{tabularx}
\end{table}

\begin{figure}[htbp]
    \centering
    \includegraphics[width=\linewidth]{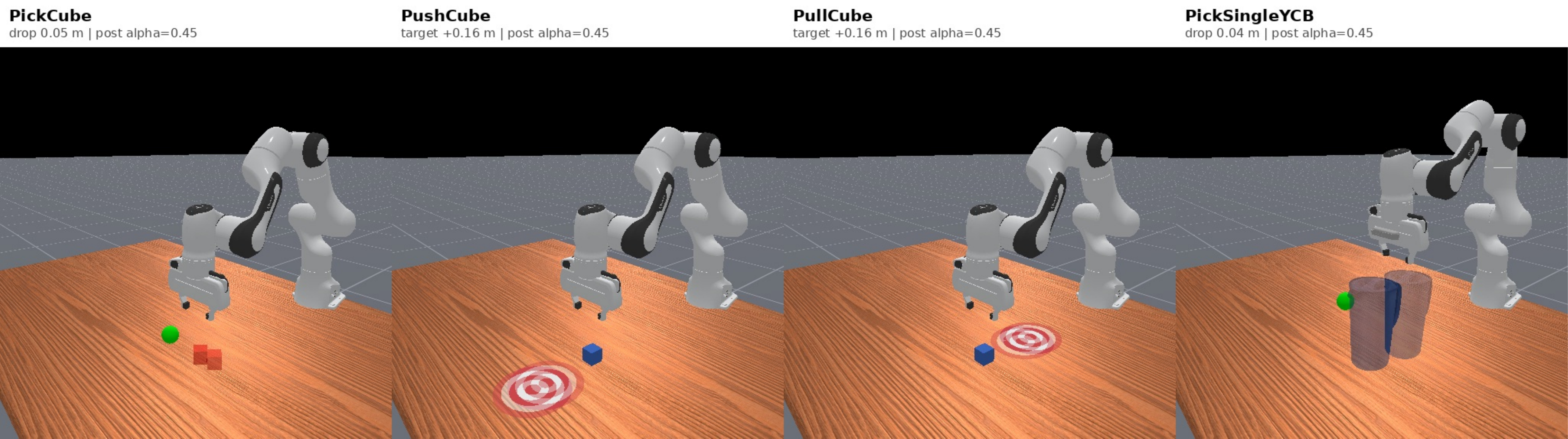}
    \vspace{-0.5em}
    \caption{
    \textbf{Controlled regressions used for recovery evaluation.}
    Each panel overlays the valid pre-regression state and the corresponding
    disturbed state. The overlays are shown only for visualization;
    \GateVLM{} receives the original RGB reference and current frame without
    overlays or information about the intervention metadata.
    }
    \label{fig:regression-disturbances}
    \vspace{-0.8em}
\end{figure}

The first three tasks use the RGB frame saved when the downstream step was
entered. In all cases, the reference is captured before the disturbance, while the relevant prerequisite is still satisfied.

\paragraph{\GateVLM{} backward-step observer.}
Normal forward progress continues to use the original forward step tracker
from the main method. The regression experiment augments this interface with
backward-only visual authority.

Let
\begin{equation}
\mathcal{I}_t^{\mathrm{back}}
=
\left[
I^{\mathrm{ref}},
I_t
\right]
\label{eq:rollback-rgb-packet}
\end{equation}
denote the RGB packet containing the valid reference image and the current
observation. At a fixed query cadence of every
\begin{equation}
q=3
\end{equation}
policy steps, Qwen3.5-9B~\citep{qwen35} predicts the step currently required
by the task:
\begin{equation}
\hat z_t
=
G_{\psi}^{\mathrm{back}}
\left(
\ell_\tau,
z_t^{-},
\left\{
\mathcal{V}_{\tau,i}
\right\}_{i=0}^{z_t^{-}},
\mathcal{I}_t^{\mathrm{back}}
\right),
\label{eq:visual-backward-observer}
\end{equation}
where $\ell_\tau$ is the task description, $z_t^{-}$ is the maintained step
before the visual decision, and $\mathcal{V}_{\tau,i}$ is the
\FrameVLM-generated visual completion description for step $i$.

The query schedule is fixed and is not triggered by the intervention.
The observer receives only the task description, ordered step descriptions,
last maintained step, and the two RGB images. It does not receive simulator
states.

\paragraph{Backward-only step update.}
The visual observer may decrease, but not increase the persistent step
state:
\begin{equation}
z_t^{+}
=
\begin{cases}
\hat z_t,
&
\hat z_t<z_t^{-},
\\[2pt]
z_t^{-},
&
\text{otherwise}.
\end{cases}
\label{eq:backward-only-update}
\end{equation}
Invalid, unchanged, or later-step predictions therefore leave the persistent
state unchanged. After an accepted rollback, the same frozen actor resumes
execution with the corrected step input:
\begin{equation}
a_t
\sim
\pi_{\theta_{\tau,\mathrm{clean}}}
\left(
\cdot
\mid
o_t^{(\mathrm{MS})},
e_\tau(z_t^{+})
\right).
\label{eq:recovery-policy}
\end{equation}
No earlier-step policy is loaded, and no policy network is switched.
Subsequent forward transitions are again handled by the original forward
step tracker.

\paragraph{Compared step trackers.}
All variants use the same task-specific frozen actor, matched environment
reset keys, disturbance operator, action space, recovery horizon, and
official task-success predicate.
\begin{enumerate}
    \item \textbf{Forward-only:}
    the pre-regression step state is retained and is not allowed to move
    backward, even after its prerequisite becomes invalid;

    \item \textbf{\GateVLM{} rollback:}
    the visual observer applies the backward-only update in
    Eq.~\ref{eq:backward-only-update};

    \item \textbf{Oracle rollback:}
    privileged simulator state supplies the evaluator-derived recovery step
    $z^\star$, providing an upper-bound control.
\end{enumerate}

\subsubsection{Evaluation Protocol}
\label{app:exp-regression-evaluation}

An episode is counted as effective only when the clean actor naturally
reaches the designated downstream step and receives the controlled
disturbance. Each task and tracker is evaluated on
\begin{equation}
N=50
\end{equation}
effective episodes using deterministic actor inference and matched reset
keys across Forward-only, \GateVLM{}, and Oracle.

The first three tasks use evaluation seed 12442 and the natural step-entry
RGB reference. \texttt{PickSingleYCB-v1} uses evaluation seed 5042 and the
refreshed-reference protocol. All four rows use a post-disturbance recovery
horizon of
\begin{equation}
H=120
\end{equation}
policy steps and a timely-rollback window of
\begin{equation}
W=10.
\end{equation}

\paragraph{Recovery Success Rate.}
Recovery Success Rate measures whether the original task is completed after
the disturbance and within the controlled recovery horizon:
\begin{equation}
\mathrm{Recovery\ SR}
=
\frac{1}{N}
\sum_{n=1}^{N}
\mathbf{1}
\left[
\exists t
\in
\left(
t_d^{(n)},
t_d^{(n)}+H
\right]
:
\mathrm{Success}_{\tau}
\left(
s_t^{\mathrm{sim},(n)}
\right)
=1
\right].
\label{eq:recovery-success-rate}
\end{equation}
The official task-success predicate is used only for outcome evaluation and
is never provided to the actor or \GateVLM{}.

\paragraph{Timely Rollback Rate.}
For episode $n$, let
\begin{equation}
t_b^{(n)}
=
\inf
\left\{
t\geq t_d^{(n)}
:
z_t<z_{t_d^-}^{(n)}
\right\}
\label{eq:first-backward-time}
\end{equation}
denote the first accepted backward update. Timely Rollback Rate measures
whether any backward correction is applied within $W$ policy steps:
\begin{equation}
\mathrm{TRR}@W
=
\frac{1}{N}
\sum_{n=1}^{N}
\mathbf{1}
\left[
0
\leq
t_b^{(n)}-t_d^{(n)}
\leq
W
\right].
\label{eq:timely-rollback-rate}
\end{equation}

\paragraph{Stage Accuracy.}
Stage Accuracy additionally requires the first correction within the same
window to exactly match the evaluator-derived recovery step:
\begin{equation}
\mathrm{SA}@W
=
\frac{1}{N}
\sum_{n=1}^{N}
\mathbf{1}
\left[
0
\leq
t_b^{(n)}-t_d^{(n)}
\leq
W
\;\land\;
z_{t_b^{(n)}}=z^{\star(n)}
\right].
\label{eq:stage-accuracy}
\end{equation}
Thus, SA penalizes a missing rollback, an insufficient rollback, and an
over-correction to a step earlier than the evaluator-derived recovery step.

\subsubsection{Experiment Results and Conclusions}

\begin{table}[htbp]
\centering
\caption{
\textbf{Controlled regression and zero-shot recovery.}
Each row contains 50 effective disturbed episodes.
}
\label{tab:backward-recovery-full}

\scriptsize
\setlength{\tabcolsep}{3.0pt}
\renewcommand{\arraystretch}{0.95}

\begin{tabularx}{\linewidth}{@{}lXccccc@{}}
\toprule
\textbf{Task}
&
\textbf{Regression}
&
\makecell{\textbf{Forward}\\\textbf{SR}}
&
\makecell{\textbf{\GateVLM{}}\\\textbf{SR}}
&
\makecell{\textbf{Oracle}\\\textbf{SR}}
&
\textbf{TRR@10}
&
\textbf{SA@10}
\\
\midrule

\texttt{PickCube-v1}
&
Object drop, $0.05$\,m
&
0/50
&
\textbf{32/50}
&
32/50
&
50/50
&
49/50
\\

\texttt{PushCube-v1}
&
Target shift, $0.16$\,m
&
2/50
&
\textbf{46/50}
&
47/50
&
50/50
&
49/50
\\

\texttt{PullCube-v1}
&
Target shift, $0.16$\,m
&
1/50
&
\textbf{46/50}
&
45/50
&
50/50
&
32/50
\\

\texttt{PickSingleYCB-v1}
&
Object drop, $0.04$\,m
&
0/50
&
\textbf{28/50}
&
29/50
&
49/50
&
35/50
\\

\midrule
\textbf{Coverage total}
&
\textit{two RGB-reference protocols}
&
\textbf{3/200}
&
\textbf{152/200}
&
\textbf{153/200}
&
\textbf{199/200}
&
\textbf{165/200}
\\
\bottomrule
\end{tabularx}
\end{table}

Across the four reported tasks, visual rollback achieves $152/200$
($76.0\%$) Recovery SR, compared with $3/200$ ($1.5\%$) for Forward-only
execution and $153/200$ ($76.5\%$) for privileged Oracle rollback.
\GateVLM{} applies a backward correction within ten policy steps in
$199/200$ episodes, corresponding to $99.5\%$ TRR@10, and exactly matches
the evaluator-derived recovery step in $165/200$ episodes, corresponding to
$82.5\%$ SA@10.

These results show that recovery behavior can already be latent in an actor
\textbf{trained only on clean stepwise rewards}. Correcting only its persistent
step context can elicit recovery from unseen regressions without updating the
actor parameters or introducing a recovery-specific policy.
\section{Additional Diagnostic Ablations}
\label{app:additional-ablations}  

\subsection{Stepwise Reward Structure}
\label{app:structure}

For the controlled reward-structure study in
Sec.~\ref{sec:exp-stepwise}, we evaluate 10 Meta-World tasks, five relatively
challenging ManiSkill tasks, and three SoftGym tasks. Together, they cover
articulated manipulation, planar transport, grasping and insertion,
contact-rich interaction, fluid transfer, and deformable-object manipulation.
The complete task list is given in Table~\ref{tab:app-q1-tasks}.

The same task subset, step decomposition, and \FrameVLM{}-generated dense reward
components are used across all four combinations of reward structure
(stepwise or all-dense) and policy step-index input (with or without the index).
Within each benchmark, policy architecture, optimization budget, and evaluation
protocol are also fixed. Stepwise variants route only the active dense component
and include the one-time transition bonus, whereas all-dense variants activate
all dense components simultaneously with no transition bonus. The privileged
\GateRule{} tracker is shared across variants only for matched task progression
and evaluation.

  \begin{table}[htbp]
\centering
\caption{\textbf{Task subsets used in the Q1 reward-structure study.} The per-task visualization and configurations can be checked in Figure~\ref{fig:benchmark_overview} and Table~\ref{tab:task_reward_spec}.}
\label{tab:app-q1-tasks}

\small
\setlength{\tabcolsep}{4pt}
\renewcommand{\arraystretch}{1.05}

\begin{tabularx}{\linewidth}{@{}l c X@{}}
\toprule
\textbf{Benchmark} & \textbf{\# Tasks} & \textbf{Tasks} \\
\midrule

Meta-World
& 10
& \texttt{handle-press-v3},
  \texttt{handle-pull-v3},
  \texttt{plate-slide-side-v3},
  \texttt{handle-press-side-v3},
  \texttt{lever-pull-v3},
  \texttt{plate-slide-v3},
  \texttt{sweep-into-v3},
  \texttt{handle-pull-side-v3},
  \texttt{pick-out-of-hole-v3},
  \texttt{sweep-v3} \\

ManiSkill
& 5
& \texttt{PegInsertionSide-v1},
  \texttt{StackCube-v1},
  \texttt{PickSingleYCB-v1},
  \texttt{RollBall-v1},
  \texttt{PokeCube-v1} \\

SoftGym
& 3
& \texttt{PassWater},
  \texttt{PourWater},
  \texttt{RopeFlatten} \\

\bottomrule
\end{tabularx}
\end{table}

\subsubsection{Experiment Setting Formulation}
For a task decomposed into $K$ non-terminal steps, let
$z_t\in\{0,\ldots,K\}$ denote the persistent step state, where $K$ is the
terminal state. We define
\[
A_i(t)=\mathbf{1}[z_t=i],
\qquad
\sum_{i=0}^{K-1}A_i(t)=1
\]
for every non-terminal timestep. Let
$D_i(x_t,x_{t-1})$ denote the \FrameVLM{}-generated dense reward component
for step $i$, and let $P_t$ collect reward terms shared across all variants.

In the \emph{stepwise} formulation, only the dense component associated with
the active step contributes:
\[
r_t^{\mathrm{step}}
=
\sum_{i=0}^{K-1}
A_i(t)\,D_i(x_t,x_{t-1})
+
b_{z_t}\,
\mathbf{1}[\text{accepted transition}]
+
P_t ,
\]
where $b_{z_t}$ is a one-time bonus for an accepted step transition.
After the transition, reward is routed to the new active step and completed-step
dense components are no longer active.

The \emph{all-dense} control removes all reward-side dependence on the current
step:
\[
r_t^{\mathrm{all}}
=
\sum_{i=0}^{K-1}
D_i(x_t,x_{t-1})
+
P_t .
\]
Thus, all \FrameVLM{}-generated dense components remain active simultaneously,
with no step-dependent routing or transition bonus. The same privileged
\GateRule{} tracker is retained only for matched task progression, termination,
and evaluation, and never contributes to the all-dense reward.

Independently, we vary whether the actor and critic receive the current step
index. This yields the $2\times2$ comparison between reward-side stepwise
structure and policy-side step context.

\subsubsection{Experiment Results and Conclusions} We independently vary whether the reward maintains an explicit stepwise state
and whether the policy receives the current step index. All variants use the
same \FrameVLM-generated reward components and matched policy-learning settings.
Values report mean success rate with standard deviation across independent
training runs as in Table~\ref{tab:app-q1-structure}.

\begin{table}[htbp]
\centering
\setlength{\belowcaptionskip}{3pt}
\caption{\textbf{Full results for stepwise reward structure and step-index input.}
}
\label{tab:app-q1-structure}

\scriptsize
\setlength{\tabcolsep}{1.5pt}
\renewcommand{\arraystretch}{0.90}

\begin{tabularx}{\linewidth}{
    @{\hspace{1pt}}
    >{\raggedright\arraybackslash}p{0.54in}
    *{8}{C}
    @{\hspace{1pt}}
}
\toprule
& \multicolumn{4}{c}{\textbf{w/ Step-Index Input}}
& \multicolumn{4}{c}{\textbf{w/o Step-Index Input}} \\
\cmidrule(lr){2-5}\cmidrule(lr){6-9}

\textbf{Reward}
& \mbox{Meta-World} & ManiSkill & SoftGym & Avg.
& \mbox{Meta-World} & ManiSkill & SoftGym & Avg. \\
\midrule

Stepwise
& 92.20$\pm$3.15
& 62.00$\pm$2.67
& 49.12$\pm$1.51
& 67.77
& 46.38$\pm$0.87
& 44.40$\pm$4.87
& 49.83$\pm$1.66
& 46.87 \\

All-dense
& 62.85$\pm$3.18
& 21.60$\pm$2.35
& 44.00$\pm$5.12
& 42.82
& 52.98$\pm$5.21
& 20.00$\pm$5.33
& 47.61$\pm$3.55
& 40.19 \\

\bottomrule
\end{tabularx}
\end{table}

\subsection{\GateVLM{} Trigger}
\label{app:transition-variants}

\subsubsection{Experiment Setting Formulation}

\paragraph{Controlled interface.}
We use the same 18-task subset as the Q1 reward-structure study, listed in
Table~\ref{tab:app-q1-tasks}. For every task, the step decomposition,
\FrameVLM{}-generated reward program, actor--critic architecture, policy
observations, PPO budget, and evaluation protocol are fixed across invocation
variants.

For benchmark $b$, the actor and critic receive the benchmark observation
together with the persistent step context:
\begin{equation}
a_t
\sim
\pi_\theta
\left(
\cdot
\mid
o_t^{(b)},\mathbf{e}(z_t)
\right),
\qquad
V_\psi
=
V_\psi
\left(
o_t^{(b)},\mathbf{e}(z_t)
\right).
\label{eq:invocation-policy-interface}
\end{equation}
The RGB observations used by \GateVLM{} are not provided to these
state-based policies. At an invoked timestep, the verifier receives the same
inputs as in the main formulation:
\begin{equation}
v_t
=
\mathbf{1}
\left[
\GateVLM{}
\left(
\ell,
z_t,
\mathcal{V}_{z_t},
\mathcal{I}_t
\right)
=
\texttt{success}
\right],
\label{eq:invocation-gate-decision}
\end{equation}
where
\begin{equation}
\mathcal{I}_t
=
\left[
I^{\mathrm{start}},
I_t^{\mathrm{history}},
I_t
\right]
\label{eq:invocation-visual-packet}
\end{equation}
is the benchmark-specific RGB packet. Outputs indicating
\texttt{failure}, \texttt{unknown}, or an unparsable response yield
$v_t=0$. All variants use Qwen3.5-9B as \GateVLM{}.

The accepted transition remains
\begin{equation}
e_t
=
\chi_t v_t,
\qquad
z_{t+1}
=
z_t+e_t.
\label{eq:invocation-transition}
\end{equation}
Thus, this ablation changes only the invocation trigger $\chi_t$; the visual
decision $v_t$, persistent state update, routed dense rewards, transition
bonus, and policy interface remain unchanged.

\paragraph{Periodic invocation.}
To isolate the effect of query timing, we replace the default
task-conditioned invocation schedule with a task-state-independent periodic
trigger. For benchmark-specific query interval $F_b$,
\begin{equation}
\chi_t^{\mathrm{freq}}(F_b)
=
\mathbf{1}[z_t<K]\,
\mathbf{1}[t\bmod F_b=0].
\label{eq:periodic-invocation}
\end{equation}

\paragraph{Benchmark-specific cadence ranges.}
A policy step does not represent the same temporal granularity across the
three benchmarks, which use different native control interfaces, episode
timescales, simulation environment and PPO rollout protocols. We therefore define \textbf{fast,
intermediate, and slow} invocation regimes separately within each benchmark,
rather than applying one absolute query interval to all environments.
Specifically, we use intervals $\{16,64,128\}$ for Meta-World and
$\{8,16,32\}$ for ManiSkill and SoftGym. These settings are intended to span
a practical range of relative query cadences within each benchmark and absolute
interval values are not compared across benchmarks.

We evaluate three query cadences per benchmark:

\begin{equation}
\begin{aligned}
\mathcal{F}_{\mathrm{MW}}
&=
\{16,64,128\},\\
\mathcal{F}_{\mathrm{MS}}
&=
\{8,16,32\},\\
\mathcal{F}_{\mathrm{SG}}
&=
\{8,16,32\},
\end{aligned}
\label{eq:invocation-intervals}
\end{equation}

corresponding to the fast, intermediate, and slow settings. Under periodic invocation,
\begin{equation}
e_t^{(F_b)}
=
\chi_t^{\mathrm{freq}}(F_b)v_t,
\qquad
z_{t+1}
=
z_t+e_t^{(F_b)}.
\label{eq:periodic-transition}
\end{equation}
Once a transition is accepted, the next step-local dense reward becomes
active, the one-time transition bonus is issued, and the updated step
one-hot is supplied to the actor and critic. Therefore, $F_b$ changes only
when verification is requested.

\paragraph{Verification-error asymmetry.}
The two types of verification error have different effects on the persistent
reward state. A false rejection leaves $z_t$ unchanged, so the current dense
reward and policy context remain active and completion can be reconsidered at
a later query. A false acceptance instead advances $z_t$, activates the next
dense reward, issues the transition bonus, and changes the policy context.

More frequent invocation reduces the delay before a valid transition can be
recognized, but also creates more opportunities for premature acceptance.
Periodic cadence therefore trades off delayed progression against erroneous
advancement.

\paragraph{Training and evaluation.}
Each task is trained independently. All cadence variants use the same reset
distribution, reward functions, transition bonuses, policy architecture,
PPO hyperparameters, training budget, checkpoint rule, and evaluation
protocol. We report benchmark-average success rates with standard deviations
over three independently trained policies. Official benchmark success
predicates are used only for evaluation.

\subsubsection{Experiment Results and Conclusions}

Periodic invocation supports policy learning across all three benchmarks,
but no single cadence performs best universally. On Meta-World, average
success increases from $12.73\pm1.46\%$ at interval 16 to
$23.60\pm2.16\%$ at interval 64 and $31.40\pm3.34\%$ at interval 128.
On ManiSkill, the intermediate interval of 16 performs best at
$43.67\pm3.18\%$, compared with $31.25\pm3.66\%$ at interval 8 and
$38.33\pm1.20\%$ at interval 32. SoftGym is comparatively insensitive,
with average success ranging from $27.11\%$ to $28.56\%$ across the three
intervals.

These results show that \method{} does not require task-conditioned query
timing to remain operational, although the appropriate cadence depends on
the benchmark. Excessively sparse queries may delay valid transitions,
whereas excessively frequent queries repeatedly expose the persistent step
state to verification errors. The weak fast-query results on Meta-World and
ManiSkill are consistent with premature advancement being more damaging than
delayed advancement. Because this ablation does not directly label individual
verification errors, we treat this explanation as a mechanistic
interpretation rather than a measured error decomposition.

\subsection{Policy Context for Stepwise Reward Learning}
\label{app:policy-context}

\subsubsection{Experiment Setting Formulation}

\paragraph{Controlled comparison.}
We use the same five ManiSkill tasks:
\texttt{StackCube-v1},
\texttt{PushCube-v1},
\texttt{PullCube-v1},
\texttt{PlaceSphere-v1}, and
\texttt{PickCube-v1} across all variants. we fix the generated dense rewards, persistent stage
machine, transition authority, PPO implementation, training budget,
checkpoint-selection rule, and held-out evaluation protocol. The only
experimental variable is the information exposed to the actor and critic.
The stage tracker continues to route rewards even when neither policy network
receives the stage index.

Let $o_t^\pi$ and $o_t^V$ denote the benchmark state observations available
to the actor and critic, respectively, and let $e(z_t)$ be the active-stage
one-hot. The full state-based policy used in Sec.~\ref{sec:exp-stepwise} has
\begin{equation}
x_t^\pi
=
\left[o_t^\pi;e(z_t)\right],
\qquad
x_t^V
=
\left[o_t^V;e(z_t)\right].
\label{eq:full-stage-policy-input}
\end{equation}
We evaluate four additional input variants:
\begin{align}
\text{Actor-blind stage:}\qquad
&
x_t^\pi=o_t^\pi,
&
x_t^V=\left[o_t^V;e(z_t)\right],
\label{eq:actor-blind-stage}
\\
\text{No policy-stage input:}\qquad
&
x_t^\pi=o_t^\pi,
&
x_t^V=o_t^V.
\label{eq:no-stage-policy}
\end{align}
\paragraph{Visual actor variants.}
For the visual policies, the current RGB image is resized to
$224\times224$ and encoded using a frozen DINOv2 ViT-S/14:
\begin{equation}
h_t
=
f_{\mathrm{DINO}}
\left(
I_t
\right)
\in
\mathbb{R}^{384}.
\label{eq:dino-policy-feature}
\end{equation}
No gradient is propagated into the encoder. Let $p_t$ denote robot-only
proprioception. The two asymmetric visual variants use
\begin{align}
\text{Visual actor with stage:}\qquad
&
x_t^\pi
=
\left[h_t;p_t;e(z_t)\right],
&
x_t^V
=
\left[o_t^V;e(z_t)\right],
\label{eq:rgb-stage-policy}
\\
\text{Visual actor without stage:}\qquad
&
x_t^\pi
=
\left[h_t;p_t\right],
&
x_t^V
=
\left[o_t^V;e(z_t)\right].
\label{eq:rgb-blind-policy}
\end{align}
Thus, the visual actor receives neither object poses, goal coordinates, nor
the benchmark privileged state. The second variant additionally removes the
stage index from the actor, while retaining a privileged stage-aware critic
only during training. We refer to these as visual policies rather than
RGB-only policies because the actor also receives robot proprioception.
The actor images are policy observations and are distinct from the RGB
packets supplied to \GateVLM{} for transition verification.

\paragraph{Training and evaluation.}
Each variant is trained independently with three seeds on all five tasks.
Checkpoints are selected using the same fixed validation protocol and
evaluated on held-out test episodes. We report the mean test success rate
over the resulting 15 task--seed cells.

\subsubsection{Experiment Results and Conclusions}

The stage-blind actor with a stage-aware critic achieves
$67.20\pm10.83\%$ mean success, compared with
$58.67\pm2.49\%$ when the step index is removed from both networks. Thus,
even when the actor does not directly observe task progress, exposing the
persistent step state to the critic provides a measurable optimization
benefit.

The visual actor with and without explicit step context obtains
$37.20\pm1.06\%$ and $37.07\pm0.96\%$, respectively. Their nearly identical
performance indicates that, under the current frozen-DINO representation,
\textbf{visual object and goal grounding} is the dominant limitation and adding the
symbolic step index alone does not resolve it. Because all reward and
optimization settings remain fixed, this decrease reflects the stricter
visual policy input rather than a change in the reward method.

\textbf{The stage-blind visual variant nevertheless demonstrates a deployment-
compatible asymmetric interface}: the deployed actor requires only frozen
visual features and robot proprioception, while privileged state and stage
information are confined to the training-time critic.

\subsection{Sensitivity to Gate Errors}
\label{app:gate-error}

\paragraph{Task group and controlled variables.}
We use three staged Meta-World tasks:
\texttt{handle-pull-side-v3},
\texttt{lever-pull-v3}, and
\texttt{sweep-into-v3}.
Across all operating points, we fix the task decomposition, step-local dense
rewards, transition bonuses, persistent step machine, actor--critic
architecture, PPO hyperparameters, training budget, checkpoint-selection
rule, and evaluation protocol. Only the gate error probabilities vary.

\paragraph{Synthetic gate errors.}
To isolate the causal effects of verification errors from model-specific VLM
behavior, we replace \GateVLM{} with a synthetic gate. At a scheduled query,
let $g_t^\star\in\{0,1\}$ denote the privileged completion label of the
active step and let $v_t\in\{0,1\}$ denote gate acceptance. We define

\begin{equation}
\Pr(v_t=1\mid g_t^\star=0)=\alpha,
\qquad
\Pr(v_t=0\mid g_t^\star=1)=\beta,
\label{eq:synthetic-gate-errors}
\end{equation}

where $\alpha$ is the false-acceptance rate (FAR) and $\beta$ is the
false-rejection rate (FRR). The transition rule remains unchanged:

\begin{equation}
e_t=\chi_t v_t,
\qquad
z_{t+1}=z_t+e_t.
\label{eq:noisy-gate-transition}
\end{equation}

A false acceptance therefore advances the persistent step state, issues the
normal transition bonus, and activates the next step reward. A false
rejection leaves the current step active, allowing a later retry.

\paragraph{Query schedule and operating-point grid.}
All settings use the same Hybrid-K64 schedule: native candidate events invoke
the gate, while an additional shadow query is issued every 64 environment
steps during an active step; coincident events are counted once. FAR and FRR
are therefore conditional rates over gate opportunities rather than
per-environment-step probabilities.

We evaluate

\begin{equation}
\alpha
\in
\{0,0.05,0.15,0.30,0.50,0.75,1.00\},
\qquad
\beta
\in
\{0,0.05,0.15,0.30\},
\label{eq:far-frr-grid}
\end{equation}

forming the $4\times7$ grid in Figure~\ref{fig:far-frr}. Each operating point
contains three independently trained policies per task, giving
$28\times3\times3=252$ task--seed runs. Checkpoints are selected using the
fixed validation protocol and evaluated on 50 held-out episodes.

\paragraph{Results.}
At $(\mathrm{FAR},\mathrm{FRR})=(0,0)$, mean success is $94.0\%$.
Increasing FAR to $0.30$ while fixing FRR at zero reduces success to
$21.8\%$, whereas increasing FRR to $0.30$ while fixing FAR at zero retains
$66.0\%$ success. The surface is not strictly monotonic at every individual
cell because of task and training-seed variability, but the overall
degradation is substantially steeper along the FAR axis. These results show
that premature step advancement is considerably more damaging than delayed
advancement under the tested protocol.

\paragraph{Real-VLM operating points.}
We additionally evaluate five Qwen-family checkpoints under the same
Hybrid-K64 query schedule, prompt, camera, and parser used for policy
training. For each task and model, FAR and FRR are measured on a fixed packet
containing 200 operational positive and 200 operational negative
opportunities. The matched panel uses \texttt{sweep-into-v3} and
\texttt{lever-pull-v3}; \texttt{handle-pull-side-v3} is retained only as an
auxiliary diagnostic because an equally complete large-model policy panel is
not available for that task.

\begin{table*}[htbp]
\centering
\caption{
\textbf{Real \GateVLM{} operating points and downstream policy success.}
All models use the same query and parsing protocol. SR reports the seed-42
held-out policy result; additional completed seeds are discussed in text.
}
\label{tab:real-vlm-gate-errors}

\scriptsize
\setlength{\tabcolsep}{4.2pt}
\renewcommand{\arraystretch}{0.92}

\begin{tabular}{@{}lccc@{\hspace{14pt}}ccc@{}}
\toprule
&
\multicolumn{3}{c}{\textbf{\texttt{sweep-into-v3}}}
&
\multicolumn{3}{c}{\textbf{\texttt{lever-pull-v3}}}
\\
\cmidrule(lr){2-4}
\cmidrule(lr){5-7}

\textbf{Model}
&
\textbf{FAR}
&
\textbf{FRR}
&
\textbf{SR}
&
\textbf{FAR}
&
\textbf{FRR}
&
\textbf{SR}
\\
\midrule

Qwen3-VL-4B
& 0.695 & 0.270 & 0.00
& 0.815 & 0.165 & 0.04
\\

Qwen3-VL-8B
& 0.540 & 0.380 & 0.44
& 0.555 & 0.465 & 0.14
\\

Qwen3.5-9B
& 0.795 & 0.195 & 0.26
& 0.830 & 0.080 & 0.00
\\

Qwen3.6-27B
& 0.895 & 0.245 & 0.58
& 0.385 & 0.095 & 0.00
\\

Qwen3-VL-32B
& 0.740 & 0.280 & 0.00
& 0.615 & 0.240 & 0.00
\\

\bottomrule
\end{tabular}
\end{table*}

\section{Full Related Work}
\label{app:full-related-work}

\subsection{Full Related Work Table}
\label{app:full-related-work}

\begin{table}[htbp]
\centering
\caption{
\textbf{Comparison of robot reward-learning and structured reward-design approaches.}
We compare how prior methods represent task progress, obtain reward supervision,
verify progress transitions, and use the resulting rewards for policy learning.
}
\label{tab:related_work}

\small
\setlength{\tabcolsep}{4pt}
\renewcommand{\arraystretch}{1.05}

\begin{tabularx}{\linewidth}{@{}
L{0.16\linewidth}
X
X
X
L{0.16\linewidth}
@{}}

\toprule
\textbf{Method} &
\makecell[l]{\textbf{Reward Structure}} &
\makecell[l]{\textbf{Reward Supervision}} &
\makecell[l]{\textbf{Transition}\\\textbf{Verification}} &
\makecell[l]{\textbf{Learning}} \\
\midrule

\multicolumn{5}{@{}l}{\textit{Visual progress rewards for robot learning}}\\[-1pt]
\midrule

LIV~\citep{liv2023} &
global value / $\Delta V$ &
temporal order + goal/outcome &
-- &
planning / LCBC \\
\cmidrule(lr){1-5}

ReWiND~\citep{rewind2025} &
global scalar progress &
target demos + temporal/rewind labels &
external success signal &
online RL \\
\cmidrule(lr){1-5}

SARM~\citep{sarm2025} &
stage-aware scalar progress &
target demos + stage boundaries &
stage re-estimation &
offline FT \\
\cmidrule(lr){1-5}

RoboReward~\citep{roboreward2026} &
global ordinal progress &
human scores + temporal labels &
terminal scoring &
online RL \\
\cmidrule(lr){1-5}

TOPReward~\citep{topreward2026} &
global completion value &
frozen VLM token likelihood &
-- &
offline FT \\
\cmidrule(lr){1-5}

Large Reward Models~\citep{lrm2026} &
global multi-head progress &
multi-source video + weak temporal labels &
completion head &
PPO / filtered BC \\
\cmidrule(lr){1-5}

RoboMeter~\citep{robometer2026} &
persistent stage + global progress &
temporal + preference labels &
success classifier &
online RL \\
\cmidrule(lr){1-5}

SOLE-R1~\citep{soler12026} &
global scalar progress &
sim progress + temporal/reversal labels &
-- &
online RL \\
\cmidrule(lr){1-5}

ProcVLM~\citep{procvlm2026} &
stage-aware scalar progress &
successful videos + procedure labels &
stage re-estimation &
offline FT \\
\cmidrule(lr){1-5}

STDR~\citep{stdr2026} &
stage-aware scalar progress &
expert videos + stage labels &
learned stage classifier &
online RL \\
\cmidrule(lr){1-5}

WVM~\citep{wvm2026} &
global distributional value &
temporal + rewind + human labels &
-- &
offline FT \\

\midrule
\multicolumn{5}{@{}l}{\textit{Programmatic and structured reward design}}\\[-1pt]
\midrule

Reward Machines~\citep{rewardmachines2018} &
persistent automaton reward &
manual automaton + event labels &
event predicate &
online RL \\
\cmidrule(lr){1-5}

Text2Reward~\citep{text2reward2024} &
flat executable reward &
task text + environment state/API &
-- &
online RL \\
\cmidrule(lr){1-5}

Eureka~\citep{eureka2024} &
flat executable reward &
task text + environment source + fitness &
reward search &
online RL \\
\cmidrule(lr){1-5}

Code-as-Reward~\citep{codeasreward2024} &
persistent subtask reward &
VLM decomposition + trajectory verification &
programmatic completion predicate &
online RL \\
\cmidrule(lr){1-5}

PRM4RL~\citep{prm4rl2025} &
stage-aware global potential &
task text + simulator state/API &
privileged-state stage inference &
online RL \\
\cmidrule(lr){1-5}

ARM-FM~\citep{armfm2026} &
persistent automaton reward &
language + environment API + FM reasoning &
event-labeling function &
compositional RL \\

\midrule

\textbf{RoboStep (ours)} &
\textbf{persistent step-local reward} &
\textbf{task/API + pretrained VLM; no reward demos} &
\textbf{semantic gate + reversible rollback} &
\textbf{online RL} \\

\bottomrule
\end{tabularx}
\end{table}

\subsection{Full Related Work Task Table}
\label{app:full-related-work-task}

\normalsize
\setlength{\tabcolsep}{3.2pt}
\renewcommand{\arraystretch}{1.0}
\setlength{\LTcapwidth}{\textwidth}
\small
\begin{xltabular}{\textwidth}{@{}L{0.92in}cX@{}}

\caption{\normalsize\textbf{Original task coverage of related work.}
Task names are copied from the original papers whenever they are explicitly
listed. Simulation and real-world evaluations are marked separately; a
benchmark count is not treated as a downstream policy task unless the paper
uses it that way.}
\label{tab:app-all-task-manifest}\\

\toprule
\textbf{Method} & \textbf{\# Tasks} & \textbf{Original Tasks (simulation / real)} \\
\midrule
\endfirsthead

\toprule
\textbf{Method} & \textbf{\# Tasks} & \textbf{Original Tasks (simulation / real)} \\
\midrule
\endhead

\midrule
\multicolumn{3}{r@{}}{\emph{Continued on next page}}\\
\endfoot

\bottomrule
\endlastfoot

LIV~\citep{liv2023}
& 20
& \textbf{Sim (11):} MetaWorld---close drawer, open drawer, turn faucet right, turn faucet left, move black mug right, move the white mug left; FrankaKitchen---open microwave, slide cabinet, open left door, turn on stove, switch on light.\newline
\textbf{Real (9):} RealRobot fruit--container assignments: \{apple, pineapple, pear\}$\times$\{tray, black pot, green pot\}. \\
\midrule

Code-as-Reward~\citep{codeasreward2024}
& 8
& \textbf{Sim (8):} MiniGrid---DoorKey, Unlock, DoorKey8x8; PandasGym---PandasPush, PandasSlide; CLIPort---Separating-Piles, Put-Blocks-in-Bowls, Packing-Shapes.\newline
\textbf{Real:} --. \\
\midrule

RL--VLM--F~\citep{rlvlmf2024}
& 7
& \textbf{Sim (7):} OpenAI Gym---Cart Pole; MetaWorld---Open Drawer, Sweep Into, Soccer; SoftGym---Fold Cloth, Straighten Rope, Pass Water.\newline
\textbf{Real:} --. \\
\midrule

EUREKA~\citep{eureka2024}
& 29
& \textbf{Sim (29):} IsaacGym---Cartpole, Quadcopter, FrankaCabinet, Anymal, BallBalance, Ant, AllegroHand, Humanoid, ShadowHand; Dexterity---Over, DoorCloseInward, DoorCloseOutward, DoorOpenInward, DoorOpenOutward, Scissors, SwingCup, Switch, Kettle, LiftUnderarm, Pen, BottleCap, CatchAbreast, CatchOver2Underarm, CatchUnderarm, ReOrientation, GraspAndPlace, BlockStack, PushBlock, TwoCatchUnderarm.\newline
\textbf{Real:} --. \\
\midrule

FuRL~\citep{furl2024}
& 10
& \textbf{Sim (10):} MetaWorld MT10---button-press-topdown-v2, door-open-v2, drawer-close-v2, drawer-open-v2, peg-insert-side-v2, pick-place-v2, push-v2, reach-v2, window-close-v2, window-open-v2.\newline
\textbf{Real:} --. \\
\midrule

VLAC~\citep{zhanggeneralist}
& 4
& \textbf{Sim:} --.\newline
\textbf{Real (4):} Desktop Sweep Disposal, Pick and Place Bowl, Unfold Mat, Rice Scooping and Transfer. \\
\midrule

GVL~\citep{gvl2025}
& 6 sim + 7 real
& \textbf{Sim (6):} single\_insertion, double\_insertion, mug\_on\_plate, bowl\_on\_rack, plate\_on\_rack, glass\_on\_rack.\newline
\textbf{Real (7 downstream):} bowl-in-rack, banana-handover, close-laptop, open-drawer, remove-gears, pen-handover, fold-dress.\newline
Additional benchmark scale: ALOHA-250 and more than 300 real-world tasks across OXE datasets. \\
\midrule

PRM4RL~\citep{prm4rl2025}
& 15
& \textbf{Sim (15):} MetaWorld---button-press, door-close, drawer-close, faucet-open, window-open, sweep-into, pick-place, pick-out-of-hole, sweep, push; ManiSkill---LiftCube, PullCube, PushCube, PickCube, LiftPegUpright.\newline
\textbf{Real:} --. \\
\midrule

ReWiND~\citep{rewind2025}
& 13
& \textbf{Sim (8):} MetaWorld---Window Close, Reach Wall, Faucet Close, Coffee Button, Button Press Wall, Door Lock, Handle Press Side, Sweep Into.\newline
\textbf{Real (5):} Separate blue/orange cups, Fold blue towel, Open red trash bin, Put orange cup on red plate, Put fruit-colored object in box. \\
\midrule

SARM~\citep{sarm2025}
& 3
& \textbf{Sim (1):} MuJoCo cube-picking.\newline
\textbf{Real (2):} T-shirt folding, unloading dishes from a rack. \\
\midrule

RoboReward~\citep{roboreward2026}
& 5 downstream
& \textbf{Sim (3):} Robomimic---Lift, Can, Square.\newline
\textbf{Real (2):} Pick-and-place monkey, Open drawer.\newline
RoboRewardBench/OXE is retained as benchmark data, not expanded into a finite task list. \\
\midrule

TOPReward~\citep{topreward2026}
& 130 bench. + 6 downstream
& \textbf{Sim:} --.\newline
\textbf{Real (6 downstream):} Place toy car in box, Stack red cube on green cube, Put pen into cup, Place doll in box, Pick up cube, Put cube in cup.\newline
ManiRewardBench contains 130 unique real-world manipulation tasks. \\
\midrule

Large Reward Models~\citep{lrm2026}
& 1 named + sim NR
& \textbf{Sim:} ManiSkill3 benchmark; task names are not enumerated in the paper text.\newline
\textbf{Real (1):} Pick-and-place toy giraffe into a target bowl. \\
\midrule

ARM--FM~\citep{armfm2026}
& 10 named + 1000 generated
& \textbf{Sim:} MiniGrid---DoorKey, UnlockToUnlock, BlockedUnlockPickup, KeyCorridorS3R3; Craftium---mine a diamond; MetaWorld---Assembly, Bin-Picking, Pick-Place, Shelf-Place, Stick-Push; XLand--MiniGrid---1,000 generated tasks.\newline
\textbf{Real:} --. \\
\midrule

RoboMeter~\citep{robometer2026}
& 8 named downstream
& \textbf{Sim (2):} LIBERO-90 Task 28 (close the top drawer), Task 33 (close the microwave).\newline
\textbf{Real (6):} Put the bowl on the table; put the corn in the pot and then put the lid on the pot; Put the bread in the oven; Put the red bowl on the blue plate; Put the marker in the pen cup; Put the red cup on the purple coaster.\newline
The full RBM-1M/RBM-EVAL data span 21 embodiments and are not a finite named-task list. \\
\midrule

ARM~\citep{arm2026}
& 1
& \textbf{Sim:} --.\newline
\textbf{Real (1):} bimanual towel-folding. \\
\midrule

SOLE--R1~\citep{soler12026}
& 39 named$^\dagger$
& \textbf{Sim (33):} MetaWorld---faucet-open, faucet-close, window-open, window-close, plate-slide (puck in net), drawer-close, drawer-open, door-close, door-open, button-press, handle-pull, soccer, lever-pull, coffee-button, handle-press, coffee-push; RoboSuite---Lift, Wipe, Door, PickPlaceCan; LIBERO---Close microwave, Open microwave, Turn on stove, Turn off stove, Close top drawer, Open bottom drawer; ManiSkill---PickPanda, PushCube, PullCube, Mobile-OpenCabDrawer, PickSO100, PickWidowXAI, PickSingleYCB.\newline
\textbf{Real (6):} Pick can, Touch strawberry in clutter, Push can to cube, Close drawer, Open drawer, Insert pipe. \\
\midrule

ProcVLM~\citep{procvlm2026}
& 10 sim + 1 real
& \textbf{Sim (10):} LIBERO-10.\newline
\textbf{Real (1):} Stack Bowls. \\
\midrule

MARVL~\citep{marvl2026}
& 8
& \textbf{Sim (8):} MetaWorld---Button press (topdown), Door open, Drawer close, Drawer open, Push, Reach, Window close, Window open.\newline
\textbf{Real:} --. Panda-Gym is used for domain-shift/cross-embodiment evaluation, but the paper does not enumerate a separate task list. \\
\midrule

SARM2~\citep{sarm22026}
& 10 bench. + 2 downstream
& \textbf{Sim:} --.\newline
\textbf{Real benchmark (10):} Pick and place plates into bin, Pick and place plates into dish rack, Fold the t-shirt, Fold the shorts, Pull plugs off the socket, Clean whiteboard with whiteboard eraser, Set dinner table, Sweep paper scraps with broom, Coil and wrap headphones, Put away an umbrella.\newline
\textbf{Real downstream (2):} Folding Shorts, Cleaning Whiteboard. \\
\midrule

STDR~\citep{stdr2026}
& 17
& \textbf{Sim (14):} MetaWorld---Button Press, Window Open, Reach Wall, Plate Slide, Push, Coffee Push, Stick Push, Pick Place; ManiSkill---PullCube, PushCube, PickCube; Franka Kitchen---Slide Cabinet, Kettle, Microwave.\newline
\textbf{Real (3):} PutPepperInBucket, PutToyOnPlate, ReachAndPullDrawer. \\
\midrule

World Value Models~\citep{wvm2026}
& 15 bench. + 6 downstream
& \textbf{Sim benchmark (6):} RoboSuite---lift, pick \& place bread, pick \& place can, pick \& place cereal, pick \& place milk, stack.\newline
\textbf{Real benchmark (9):} AgileX---carrot off plate, carrot on plate, mickey box, sausage pot; ARX---flip bottles, open box, split cups, stack bowls, stack cups.\newline
\textbf{Downstream:} Sim---Stack cubes, Put cereal in tray, Put milk in tray; Real---Put mouse in box, Put shrimp in pot, Put marker in holder. \\
\midrule

RoboStep (ours)
& 75 sim + 4 real
& \textbf{Sim:} Demonstrated in Table~\ref{tab:task_reward_spec}.\newline
\textbf{Real:} Demonstrated in Table~\ref{fig:real-tasks-transfer}. \\

\end{xltabular}

\vspace{-1mm}
\noindent{$^\dagger$ The SOLE--R1 paper lists 39 named tasks in its
task table, while one figure caption states 40.}
\normalsize

Reward-model-only evaluations on fixed
trajectories (e.g., progress/ranking accuracy or failure detection) are excluded.
Each task is assigned to one primary physical-interaction category.
Long-horizon is an orthogonal attribute and counts explicitly identified
long-horizon task families; difficulty variants and atomic sub-stages are not
counted separately. `0' denotes that no such task is reported, `NR' denotes
insufficient task-level information in the original paper for a defensible
count, and `N/A' denotes a reported robotic task suite that cannot be
meaningfully partitioned using this manipulation taxonomy. Aggregate benchmark names are not expanded into task-level totals unless
the benchmark provides an explicit task list. For SOLE--R1, the main
text and Fig.~3 report 40 tasks, whereas Appendix Table~3 lists 39; we follow
Fig.~3, which additionally includes \texttt{handle-press-side}.


\normalsize
\setlength{\tabcolsep}{2.2pt}
\renewcommand{\arraystretch}{1.0}
\newcolumntype{C}{>{\centering\arraybackslash}p{0.55in}}
\newcolumntype{H}{>{\centering\arraybackslash}p{0.85in}}
\setlength{\LTleft}{0pt}
\setlength{\LTright}{0pt}
\setlength{\LTcapwidth}{\textwidth}

\small
\begin{xltabular}{\textwidth}{@{}L{0.98in}*{6}{C}H@{}}
\caption{\normalsize\textbf{Task-type coverage of reported robot-learning tasks.}
Counts include only downstream robot policy/control experiments in which the
proposed method is used for policy training, policy refinement, planning, or
closed-loop execution. }
\label{tab:app-related-work-task-type}\\
\toprule
\textbf{Method} &
\textbf{Sim} &
\textbf{Real} &
\textbf{Rigid} &
\textbf{Artic.} &
\makecell{\textbf{Precise/}\\\textbf{contact}} &
\makecell{\textbf{Deform./}\\\textbf{Fluid}} &
\makecell{\textbf{Long-}\\\textbf{horizon}} \\
\midrule
\endfirsthead

\toprule
\textbf{Method} &
\textbf{Sim} &
\textbf{Real} &
\textbf{Rigid} &
\textbf{Artic.} &
\makecell{\textbf{Precise/}\\\textbf{contact}} &
\makecell{\textbf{Deform./}\\\textbf{Fluid}} &
\makecell{\textbf{Long-}\\\textbf{horizon}} \\
\midrule
\endhead

\midrule
\multicolumn{8}{r@{}}{\emph{Continued on next page}}\\
\endfoot

\bottomrule
\endlastfoot

LIV~\citep{liv2023}
& 11 & 9 & 11 & 9 & 0 & 0 & 3 \\
\midrule

Code-as-Reward~\citep{codeasreward2024}
& 2 & 0 & 2 & 0 & 0 & 0 & 0 \\
\midrule

RL--VLM--F~\citep{rlvlmf2024}
& 6 & 0 & 2 & 1 & 0 & 3 & 0 \\
\midrule

EUREKA~\citep{eureka2024}
& 29 & 0 & N/A & N/A & N/A & N/A & 0 \\
\midrule

FuRL~\citep{furl2024}
& 10 & 0 & 3 & 5 & 2 & 0 & 0 \\
\midrule

VLAC~\citep{zhanggeneralist}
& 0 & 4 & 1 & 0 & 1 & 2 & 0 \\
\midrule

GVL~\citep{gvl2025}
& 6 & 7 & 7 & 2 & 3 & 1 & 0 \\
\midrule

PRM4RL~\citep{prm4rl2025}
& 15 & 0 & 8 & 5 & 2 & 0 & 2 \\
\midrule

ReWiND~\citep{rewind2025}
& 8 & 5 & 5 & 5 & 2 & 1 & 0 \\
\midrule

SARM~\citep{sarm2025}
& 1 & 1 & 1 & 0 & 0 & 1 & 1 \\
\midrule

RoboReward~\citep{roboreward2026}
& 3 & 2 & 3 & 1 & 1 & 0 & 0 \\
\midrule

TOPReward~\citep{topreward2026}
& 0 & 6 & 3 & 0 & 3 & 0 & 0 \\
\midrule

Large Reward Models~\citep{lrm2026}
& NR & 1 & NR & NR & NR & NR & NR \\
\midrule

ARM--FM~\citep{armfm2026}
& 5 & 0 & 3 & 0 & 2 & 0 & 0 \\
\midrule

Phys2Real~\citep{phys2real2026}
& 2 families & 2 families & 2 & 0 & 2 & 0 & 0 \\
\midrule

RoboMeter~\citep{robometer2026}
& 2 & 7 & 5 & 3 & 1 & 0 & 1 \\
\midrule

ARM~\citep{arm2026}
& 0 & 1 & 0 & 0 & 0 & 1 & 1 \\
\midrule

SOLE--R1~\citep{soler12026}
& 34 & 6 & 13 & 22 & 5 & 0 & 0 \\
\midrule

ProcVLM~\citep{procvlm2026}
& 10 & 1 & 7 & 3 & 1 & 0 & 10 \\
\midrule

MARVL~\citep{marvl2026}
& 10 & 0 & 3 & 5 & 2 & 0 & 0 \\
\midrule

SARM2~\citep{sarm22026}
& 0 & 2 & 0 & 0 & 1 & 1 & 2 \\
\midrule

STDR~\citep{stdr2026}
& 14 & 0 & 6 & 3 & 5 & 0 & NR \\
\midrule

World Value Models~\citep{wvm2026}
& 3 & 3 & 4 & 0 & 2 & 0 & 0 \\
\midrule

CoRe~\citep{core2026}
& 10 & 5 & 2 & 2 & 6 & 5 & 0 \\
\midrule

\textbf{RoboStep (ours)}
  & 75 & 4 & 22 & 23 & 24 & 6 & 17 \\

\end{xltabular}
\normalsize

\begingroup
\scriptsize
\setlength{\tabcolsep}{1.8pt}
\renewcommand{\arraystretch}{1.00}
\setlength{\LTleft}{0pt}
\setlength{\LTright}{0pt}
\setlength{\LTcapwidth}{\textwidth}
\newcolumntype{T}{>{\raggedright\arraybackslash\hspace{0pt}}p{0.92in}}
\newcolumntype{I}{>{\raggedright\arraybackslash\hspace{0pt}}p{1.38in}}
\newcolumntype{K}{>{\centering\arraybackslash}p{0.22in}}
\newcolumntype{S}{>{\raggedright\arraybackslash\hspace{0pt}\hsize=.82\hsize}X}
\newcolumntype{R}{>{\raggedright\arraybackslash\hspace{0pt}\hsize=1.18\hsize}X}
\newcommand{\taskwithtype}[2]{%
  \parbox[t]{\linewidth}{%
    \raggedright
    #1\par
    \vspace{-1pt}
    {\scriptsize\textit{#2}}%
  }%
}

\begin{xltabular}{\textwidth}{@{}T I K S R@{}}
\caption{\textbf{Per-task stepwise reward specification.} Frame-VLM state/API inputs, ordered stages, and dense reward components. \textbf{\textit{LH}} is short for long-horizon.}
\label{tab:task_reward_spec}\\
\toprule
\textbf{Task} & \textbf{Frame-VLM state/API} & \textbf{$K$} &
\textbf{Ordered stages} & \textbf{Dense components}\\
\midrule
\endfirsthead

\multicolumn{5}{c}{\tablename\ \thetable\ -- continued}\\
\toprule
\textbf{Task} & \textbf{Frame-VLM state/API} & \textbf{$K$} &
\textbf{Ordered stages} & \textbf{Dense components}\\
\midrule
\endhead

\midrule
\multicolumn{5}{r}{Continued on next page}\\
\endfoot

\bottomrule
\endlastfoot
\multicolumn{5}{@{}l}{\textbf{MetaWorld}} \\
\midrule
\taskwithtype{\texttt{assembly-\allowbreak{}v3}}{Precise\textsuperscript{LH}} & TCP, object/target/initial poses, gripper, optional orientation & 5 & \textbf{1} \texttt{approach\_\allowbreak{}object}; \textbf{2} \texttt{capture\_\allowbreak{}object}; \textbf{3} \texttt{fixture\_\allowbreak{}clearance}; \textbf{4} \texttt{transport\_\allowbreak{}or\_\allowbreak{}extract}; \textbf{5} \texttt{verify\_\allowbreak{}task\_\allowbreak{}outcome} & \textbf{1} \texttt{tcp\_\allowbreak{}to\_\allowbreak{}pregrasp}, \texttt{gripper\_\allowbreak{}opening}; \textbf{2} \texttt{tcp\_\allowbreak{}to\_\allowbreak{}grasp}, \texttt{held}, \texttt{capture\_\allowbreak{}progress}; \textbf{3} \texttt{tcp\_\allowbreak{}to\_\allowbreak{}vertical}, \texttt{clearance\_\allowbreak{}progress}, \texttt{tcp\_\allowbreak{}to\_\allowbreak{}grasp}, \texttt{held}; \textbf{4} \texttt{tcp\_\allowbreak{}to\_\allowbreak{}transport}, \texttt{object\_\allowbreak{}transport\_\allowbreak{}to\_\allowbreak{}final}, \texttt{clearance\_\allowbreak{}progress}, \texttt{tcp\_\allowbreak{}to\_\allowbreak{}grasp}, \texttt{held}, \texttt{object\_\allowbreak{}to\_\allowbreak{}final}; \textbf{5} \texttt{object\_\allowbreak{}to\_\allowbreak{}final}, \texttt{fine\_\allowbreak{}object\_\allowbreak{}to\_\allowbreak{}final}, \texttt{tcp\_\allowbreak{}to\_\allowbreak{}grasp}, \texttt{held} \\[0.3pt]
\taskwithtype{\texttt{basketball-\allowbreak{}v3}}{Precise\textsuperscript{LH}} & TCP, object/target/initial poses, gripper, optional orientation & 5 & \textbf{1} \texttt{approach\_\allowbreak{}object}; \textbf{2} \texttt{capture\_\allowbreak{}object}; \textbf{3} \texttt{fixture\_\allowbreak{}clearance}; \textbf{4} \texttt{transport\_\allowbreak{}or\_\allowbreak{}extract}; \textbf{5} \texttt{verify\_\allowbreak{}task\_\allowbreak{}outcome} & \textbf{1} \texttt{tcp\_\allowbreak{}to\_\allowbreak{}pregrasp}, \texttt{gripper\_\allowbreak{}opening}; \textbf{2} \texttt{tcp\_\allowbreak{}to\_\allowbreak{}grasp}, \texttt{held}, \texttt{capture\_\allowbreak{}progress}; \textbf{3} \texttt{tcp\_\allowbreak{}to\_\allowbreak{}vertical}, \texttt{clearance\_\allowbreak{}progress}, \texttt{tcp\_\allowbreak{}to\_\allowbreak{}grasp}, \texttt{held}; \textbf{4} \texttt{tcp\_\allowbreak{}to\_\allowbreak{}transport}, \texttt{object\_\allowbreak{}transport\_\allowbreak{}to\_\allowbreak{}final}, \texttt{clearance\_\allowbreak{}progress}, \texttt{tcp\_\allowbreak{}to\_\allowbreak{}grasp}, \texttt{held}, \texttt{object\_\allowbreak{}to\_\allowbreak{}final}; \textbf{5} \texttt{object\_\allowbreak{}to\_\allowbreak{}final}, \texttt{fine\_\allowbreak{}object\_\allowbreak{}to\_\allowbreak{}final}, \texttt{tcp\_\allowbreak{}to\_\allowbreak{}grasp}, \texttt{held} \\[0.3pt]
\taskwithtype{\texttt{bin-\allowbreak{}picking-\allowbreak{}v3}}{Precise\textsuperscript{LH}} & TCP, object/target/initial poses, gripper, optional orientation & 5 & \textbf{1} \texttt{approach\_\allowbreak{}object}; \textbf{2} \texttt{capture\_\allowbreak{}object}; \textbf{3} \texttt{fixture\_\allowbreak{}clearance}; \textbf{4} \texttt{transport\_\allowbreak{}or\_\allowbreak{}extract}; \textbf{5} \texttt{verify\_\allowbreak{}task\_\allowbreak{}outcome} & \textbf{1} \texttt{tcp\_\allowbreak{}to\_\allowbreak{}pregrasp}, \texttt{gripper\_\allowbreak{}opening}; \textbf{2} \texttt{tcp\_\allowbreak{}to\_\allowbreak{}grasp}, \texttt{held}, \texttt{capture\_\allowbreak{}progress}; \textbf{3} \texttt{tcp\_\allowbreak{}to\_\allowbreak{}vertical}, \texttt{clearance\_\allowbreak{}progress}, \texttt{tcp\_\allowbreak{}to\_\allowbreak{}grasp}, \texttt{held}; \textbf{4} \texttt{tcp\_\allowbreak{}to\_\allowbreak{}transport}, \texttt{object\_\allowbreak{}transport\_\allowbreak{}to\_\allowbreak{}final}, \texttt{clearance\_\allowbreak{}progress}, \texttt{tcp\_\allowbreak{}to\_\allowbreak{}grasp}, \texttt{held}, \texttt{object\_\allowbreak{}to\_\allowbreak{}final}; \textbf{5} \texttt{object\_\allowbreak{}to\_\allowbreak{}final}, \texttt{fine\_\allowbreak{}object\_\allowbreak{}to\_\allowbreak{}final}, \texttt{tcp\_\allowbreak{}to\_\allowbreak{}grasp}, \texttt{held} \\[0.3pt]
\taskwithtype{\texttt{box-\allowbreak{}close-\allowbreak{}v3}}{Artic.\textsuperscript{LH}} & TCP, object/target/initial poses, gripper, optional orientation & 5 & \textbf{1} \texttt{approach\_\allowbreak{}object}; \textbf{2} \texttt{capture\_\allowbreak{}object}; \textbf{3} \texttt{fixture\_\allowbreak{}clearance}; \textbf{4} \texttt{transport\_\allowbreak{}or\_\allowbreak{}extract}; \textbf{5} \texttt{verify\_\allowbreak{}task\_\allowbreak{}outcome} & \textbf{1} \texttt{tcp\_\allowbreak{}to\_\allowbreak{}pregrasp}, \texttt{gripper\_\allowbreak{}opening}; \textbf{2} \texttt{tcp\_\allowbreak{}to\_\allowbreak{}grasp}, \texttt{held}, \texttt{capture\_\allowbreak{}progress}; \textbf{3} \texttt{tcp\_\allowbreak{}to\_\allowbreak{}vertical}, \texttt{clearance\_\allowbreak{}progress}, \texttt{tcp\_\allowbreak{}to\_\allowbreak{}grasp}, \texttt{held}; \textbf{4} \texttt{tcp\_\allowbreak{}to\_\allowbreak{}transport}, \texttt{object\_\allowbreak{}transport\_\allowbreak{}to\_\allowbreak{}final}, \texttt{clearance\_\allowbreak{}progress}, \texttt{tcp\_\allowbreak{}to\_\allowbreak{}grasp}, \texttt{held}, \texttt{object\_\allowbreak{}to\_\allowbreak{}final}; \textbf{5} \texttt{object\_\allowbreak{}to\_\allowbreak{}final}, \texttt{fine\_\allowbreak{}object\_\allowbreak{}to\_\allowbreak{}final}, \texttt{tcp\_\allowbreak{}to\_\allowbreak{}grasp}, \texttt{held} \\[0.3pt]
\taskwithtype{\texttt{button-\allowbreak{}press-\allowbreak{}topdown-\allowbreak{}v3}}{Artic.} & TCP, control/handle, target pose & 3 & \textbf{1} \texttt{approach\_\allowbreak{}control}; \textbf{2} \texttt{contact\_\allowbreak{}control}; \textbf{3} \texttt{verify\_\allowbreak{}task\_\allowbreak{}outcome} & \textbf{1} \texttt{prepress\_\allowbreak{}distance}; \textbf{2} \texttt{tcp\_\allowbreak{}to\_\allowbreak{}contact\_\allowbreak{}point}; \textbf{3} \texttt{contact\_\allowbreak{}distance}, \texttt{press\_\allowbreak{}or\_\allowbreak{}target\_\allowbreak{}distance}, \texttt{tcp\_\allowbreak{}press\_\allowbreak{}axis\_\allowbreak{}distance} \\[0.3pt]
\taskwithtype{\texttt{button-\allowbreak{}press-\allowbreak{}topdown-\allowbreak{}wall-\allowbreak{}v3}}{Artic.} & TCP, control/handle, target pose & 3 & \textbf{1} \texttt{approach\_\allowbreak{}control}; \textbf{2} \texttt{contact\_\allowbreak{}control}; \textbf{3} \texttt{verify\_\allowbreak{}task\_\allowbreak{}outcome} & \textbf{1} \texttt{prepress\_\allowbreak{}distance}; \textbf{2} \texttt{tcp\_\allowbreak{}to\_\allowbreak{}contact\_\allowbreak{}point}; \textbf{3} \texttt{contact\_\allowbreak{}distance}, \texttt{press\_\allowbreak{}or\_\allowbreak{}target\_\allowbreak{}distance}, \texttt{tcp\_\allowbreak{}press\_\allowbreak{}axis\_\allowbreak{}distance} \\[0.3pt]
\taskwithtype{\texttt{button-\allowbreak{}press-\allowbreak{}v3}}{Artic.} & TCP, control/handle, target pose & 3 & \textbf{1} \texttt{approach\_\allowbreak{}control}; \textbf{2} \texttt{contact\_\allowbreak{}control}; \textbf{3} \texttt{verify\_\allowbreak{}task\_\allowbreak{}outcome} & \textbf{1} \texttt{prepress\_\allowbreak{}distance}; \textbf{2} \texttt{tcp\_\allowbreak{}to\_\allowbreak{}contact\_\allowbreak{}point}; \textbf{3} \texttt{contact\_\allowbreak{}distance}, \texttt{press\_\allowbreak{}or\_\allowbreak{}target\_\allowbreak{}distance}, \texttt{tcp\_\allowbreak{}press\_\allowbreak{}axis\_\allowbreak{}distance} \\[0.3pt]
\taskwithtype{\texttt{button-\allowbreak{}press-\allowbreak{}wall-\allowbreak{}v3}}{Artic.} & TCP, control/handle, target pose & 3 & \textbf{1} \texttt{approach\_\allowbreak{}control}; \textbf{2} \texttt{contact\_\allowbreak{}control}; \textbf{3} \texttt{verify\_\allowbreak{}task\_\allowbreak{}outcome} & \textbf{1} \texttt{prepress\_\allowbreak{}distance}; \textbf{2} \texttt{tcp\_\allowbreak{}to\_\allowbreak{}contact\_\allowbreak{}point}; \textbf{3} \texttt{contact\_\allowbreak{}distance}, \texttt{press\_\allowbreak{}or\_\allowbreak{}target\_\allowbreak{}distance}, \texttt{tcp\_\allowbreak{}press\_\allowbreak{}axis\_\allowbreak{}distance} \\[0.3pt]
\taskwithtype{\texttt{coffee-\allowbreak{}button-\allowbreak{}v3}}{Artic.} & TCP, control/handle, target pose & 3 & \textbf{1} \texttt{approach\_\allowbreak{}control}; \textbf{2} \texttt{contact\_\allowbreak{}control}; \textbf{3} \texttt{verify\_\allowbreak{}task\_\allowbreak{}outcome} & \textbf{1} \texttt{prepress\_\allowbreak{}distance}; \textbf{2} \texttt{tcp\_\allowbreak{}to\_\allowbreak{}contact\_\allowbreak{}point}; \textbf{3} \texttt{contact\_\allowbreak{}distance}, \texttt{press\_\allowbreak{}or\_\allowbreak{}target\_\allowbreak{}distance}, \texttt{tcp\_\allowbreak{}press\_\allowbreak{}axis\_\allowbreak{}distance} \\[0.3pt]
\taskwithtype{\texttt{coffee-\allowbreak{}pull-\allowbreak{}v3}}{Rigid\textsuperscript{LH}} & TCP, object/target/initial poses, gripper, optional orientation & 5 & \textbf{1} \texttt{approach\_\allowbreak{}object}; \textbf{2} \texttt{capture\_\allowbreak{}object}; \textbf{3} \texttt{fixture\_\allowbreak{}clearance}; \textbf{4} \texttt{transport\_\allowbreak{}or\_\allowbreak{}extract}; \textbf{5} \texttt{verify\_\allowbreak{}task\_\allowbreak{}outcome} & \textbf{1} \texttt{tcp\_\allowbreak{}to\_\allowbreak{}pregrasp}, \texttt{gripper\_\allowbreak{}opening}; \textbf{2} \texttt{tcp\_\allowbreak{}to\_\allowbreak{}grasp}, \texttt{held}, \texttt{capture\_\allowbreak{}progress}; \textbf{3} \texttt{tcp\_\allowbreak{}to\_\allowbreak{}vertical}, \texttt{clearance\_\allowbreak{}progress}, \texttt{tcp\_\allowbreak{}to\_\allowbreak{}grasp}, \texttt{held}; \textbf{4} \texttt{tcp\_\allowbreak{}to\_\allowbreak{}transport}, \texttt{object\_\allowbreak{}transport\_\allowbreak{}to\_\allowbreak{}final}, \texttt{clearance\_\allowbreak{}progress}, \texttt{tcp\_\allowbreak{}to\_\allowbreak{}grasp}, \texttt{held}, \texttt{object\_\allowbreak{}to\_\allowbreak{}final}; \textbf{5} \texttt{object\_\allowbreak{}to\_\allowbreak{}final}, \texttt{fine\_\allowbreak{}object\_\allowbreak{}to\_\allowbreak{}final}, \texttt{tcp\_\allowbreak{}to\_\allowbreak{}grasp}, \texttt{held} \\[0.3pt]
\taskwithtype{\texttt{coffee-\allowbreak{}push-\allowbreak{}v3}}{Rigid} & TCP, object/target poses, gripper & 3 & \textbf{1} \texttt{approach\_\allowbreak{}object}; \textbf{2} \texttt{contact\_\allowbreak{}object}; \textbf{3} \texttt{verify\_\allowbreak{}task\_\allowbreak{}outcome} & \textbf{1} \texttt{prepush\_\allowbreak{}distance}; \textbf{2} \texttt{contact\_\allowbreak{}distance}, \texttt{tcp\_\allowbreak{}obj\_\allowbreak{}distance}, \texttt{gripper\_\allowbreak{}opening}; \textbf{3} \texttt{obj\_\allowbreak{}target\_\allowbreak{}distance}, \texttt{target\_\allowbreak{}contact\_\allowbreak{}distance}, \texttt{tcp\_\allowbreak{}obj\_\allowbreak{}distance}, \texttt{gripper\_\allowbreak{}opening} \\[0.3pt]
\taskwithtype{\texttt{dial-\allowbreak{}turn-\allowbreak{}v3}}{Artic.} & TCP, control/handle, target pose & 3 & \textbf{1} \texttt{approach\_\allowbreak{}control}; \textbf{2} \texttt{contact\_\allowbreak{}control}; \textbf{3} \texttt{verify\_\allowbreak{}task\_\allowbreak{}outcome} & \textbf{1} \texttt{prepress\_\allowbreak{}distance}; \textbf{2} \texttt{tcp\_\allowbreak{}to\_\allowbreak{}contact\_\allowbreak{}point}; \textbf{3} \texttt{contact\_\allowbreak{}distance}, \texttt{press\_\allowbreak{}or\_\allowbreak{}target\_\allowbreak{}distance}, \texttt{tcp\_\allowbreak{}press\_\allowbreak{}axis\_\allowbreak{}distance} \\[0.3pt]
\taskwithtype{\texttt{disassemble-\allowbreak{}v3}}{Precise\textsuperscript{LH}} & TCP, object/target/initial poses, gripper, optional orientation & 5 & \textbf{1} \texttt{approach\_\allowbreak{}object}; \textbf{2} \texttt{capture\_\allowbreak{}object}; \textbf{3} \texttt{fixture\_\allowbreak{}clearance}; \textbf{4} \texttt{transport\_\allowbreak{}or\_\allowbreak{}extract}; \textbf{5} \texttt{verify\_\allowbreak{}task\_\allowbreak{}outcome} & \textbf{1} \texttt{tcp\_\allowbreak{}to\_\allowbreak{}pregrasp}, \texttt{gripper\_\allowbreak{}opening}; \textbf{2} \texttt{tcp\_\allowbreak{}to\_\allowbreak{}grasp}, \texttt{held}, \texttt{capture\_\allowbreak{}progress}; \textbf{3} \texttt{tcp\_\allowbreak{}to\_\allowbreak{}vertical}, \texttt{clearance\_\allowbreak{}progress}, \texttt{tcp\_\allowbreak{}to\_\allowbreak{}grasp}, \texttt{held}; \textbf{4} \texttt{tcp\_\allowbreak{}to\_\allowbreak{}transport}, \texttt{object\_\allowbreak{}transport\_\allowbreak{}to\_\allowbreak{}final}, \texttt{clearance\_\allowbreak{}progress}, \texttt{tcp\_\allowbreak{}to\_\allowbreak{}grasp}, \texttt{held}, \texttt{object\_\allowbreak{}to\_\allowbreak{}final}; \textbf{5} \texttt{object\_\allowbreak{}to\_\allowbreak{}final}, \texttt{fine\_\allowbreak{}object\_\allowbreak{}to\_\allowbreak{}final}, \texttt{tcp\_\allowbreak{}to\_\allowbreak{}grasp}, \texttt{held} \\[0.3pt]
\taskwithtype{\texttt{door-\allowbreak{}close-\allowbreak{}v3}}{Artic.} & TCP, control/handle, target pose & 3 & \textbf{1} \texttt{approach\_\allowbreak{}control}; \textbf{2} \texttt{contact\_\allowbreak{}control}; \textbf{3} \texttt{verify\_\allowbreak{}task\_\allowbreak{}outcome} & \textbf{1} \texttt{prepress\_\allowbreak{}distance}; \textbf{2} \texttt{tcp\_\allowbreak{}to\_\allowbreak{}contact\_\allowbreak{}point}; \textbf{3} \texttt{contact\_\allowbreak{}distance}, \texttt{press\_\allowbreak{}or\_\allowbreak{}target\_\allowbreak{}distance}, \texttt{tcp\_\allowbreak{}press\_\allowbreak{}axis\_\allowbreak{}distance} \\[0.3pt]
\taskwithtype{\texttt{door-\allowbreak{}lock-\allowbreak{}v3}}{Artic.} & TCP, control/handle, target pose & 3 & \textbf{1} \texttt{approach\_\allowbreak{}control}; \textbf{2} \texttt{contact\_\allowbreak{}control}; \textbf{3} \texttt{verify\_\allowbreak{}task\_\allowbreak{}outcome} & \textbf{1} \texttt{prepress\_\allowbreak{}distance}; \textbf{2} \texttt{tcp\_\allowbreak{}to\_\allowbreak{}contact\_\allowbreak{}point}; \textbf{3} \texttt{contact\_\allowbreak{}distance}, \texttt{press\_\allowbreak{}or\_\allowbreak{}target\_\allowbreak{}distance}, \texttt{tcp\_\allowbreak{}press\_\allowbreak{}axis\_\allowbreak{}distance} \\[0.3pt]
\taskwithtype{\texttt{door-\allowbreak{}open-\allowbreak{}v3}}{Artic.} & TCP, control/handle, target pose & 3 & \textbf{1} \texttt{approach\_\allowbreak{}control}; \textbf{2} \texttt{contact\_\allowbreak{}control}; \textbf{3} \texttt{verify\_\allowbreak{}task\_\allowbreak{}outcome} & \textbf{1} \texttt{prepress\_\allowbreak{}distance}; \textbf{2} \texttt{tcp\_\allowbreak{}to\_\allowbreak{}contact\_\allowbreak{}point}; \textbf{3} \texttt{contact\_\allowbreak{}distance}, \texttt{press\_\allowbreak{}or\_\allowbreak{}target\_\allowbreak{}distance}, \texttt{tcp\_\allowbreak{}press\_\allowbreak{}axis\_\allowbreak{}distance} \\[0.3pt]
\taskwithtype{\texttt{door-\allowbreak{}unlock-\allowbreak{}v3}}{Artic.} & TCP, control/handle, target pose & 3 & \textbf{1} \texttt{approach\_\allowbreak{}lock}; \textbf{2} \texttt{contact\_\allowbreak{}lock}; \textbf{3} \texttt{unlock\_\allowbreak{}articulation} & \textbf{1} \texttt{prepress\_\allowbreak{}distance}; \textbf{2} \texttt{tcp\_\allowbreak{}to\_\allowbreak{}contact\_\allowbreak{}point}; \textbf{3} \texttt{contact\_\allowbreak{}distance}, \texttt{press\_\allowbreak{}or\_\allowbreak{}target\_\allowbreak{}distance}, \texttt{tcp\_\allowbreak{}press\_\allowbreak{}axis\_\allowbreak{}distance} \\[0.3pt]
\taskwithtype{\texttt{drawer-\allowbreak{}close-\allowbreak{}v3}}{Artic.} & TCP, handle/target poses & 3 & \textbf{1} \texttt{approach\_\allowbreak{}handle}; \textbf{2} \texttt{contact\_\allowbreak{}handle}; \textbf{3} \texttt{close\_\allowbreak{}drawer} & \textbf{1} \texttt{align\_\allowbreak{}xy\_\allowbreak{}distance}, \texttt{tcp\_\allowbreak{}above\_\allowbreak{}handle}; \textbf{2} \texttt{hook\_\allowbreak{}distance}, \texttt{tcp\_\allowbreak{}above\_\allowbreak{}handle}; \textbf{3} \texttt{open\_\allowbreak{}distance}, \texttt{pull\_\allowbreak{}waypoint\_\allowbreak{}distance} \\[0.3pt]
\taskwithtype{\texttt{drawer-\allowbreak{}open-\allowbreak{}v3}}{Artic.} & TCP, handle/target poses & 3 & \textbf{1} \texttt{align\_\allowbreak{}above\_\allowbreak{}handle}; \textbf{2} \texttt{hook\_\allowbreak{}handle}; \textbf{3} \texttt{open\_\allowbreak{}drawer} & \textbf{1} \texttt{align\_\allowbreak{}xy\_\allowbreak{}distance}, \texttt{tcp\_\allowbreak{}above\_\allowbreak{}handle}; \textbf{2} \texttt{hook\_\allowbreak{}distance}, \texttt{tcp\_\allowbreak{}above\_\allowbreak{}handle}; \textbf{3} \texttt{open\_\allowbreak{}distance}, \texttt{pull\_\allowbreak{}waypoint\_\allowbreak{}distance} \\[0.3pt]
\taskwithtype{\texttt{faucet-\allowbreak{}close-\allowbreak{}v3}}{Artic.} & TCP, control/handle, target pose & 3 & \textbf{1} \texttt{approach\_\allowbreak{}handle}; \textbf{2} \texttt{contact\_\allowbreak{}handle}; \textbf{3} \texttt{close\_\allowbreak{}faucet} & \textbf{1} \texttt{prepress\_\allowbreak{}distance}; \textbf{2} \texttt{tcp\_\allowbreak{}to\_\allowbreak{}contact\_\allowbreak{}point}; \textbf{3} \texttt{contact\_\allowbreak{}distance}, \texttt{press\_\allowbreak{}or\_\allowbreak{}target\_\allowbreak{}distance}, \texttt{tcp\_\allowbreak{}press\_\allowbreak{}axis\_\allowbreak{}distance} \\[0.3pt]
\taskwithtype{\texttt{faucet-\allowbreak{}open-\allowbreak{}v3}}{Artic.} & TCP, control/handle, target pose & 3 & \textbf{1} \texttt{approach\_\allowbreak{}handle}; \textbf{2} \texttt{contact\_\allowbreak{}handle}; \textbf{3} \texttt{open\_\allowbreak{}faucet} & \textbf{1} \texttt{prepress\_\allowbreak{}distance}; \textbf{2} \texttt{tcp\_\allowbreak{}to\_\allowbreak{}contact\_\allowbreak{}point}; \textbf{3} \texttt{contact\_\allowbreak{}distance}, \texttt{press\_\allowbreak{}or\_\allowbreak{}target\_\allowbreak{}distance}, \texttt{tcp\_\allowbreak{}press\_\allowbreak{}axis\_\allowbreak{}distance} \\[0.3pt]
\taskwithtype{\texttt{hammer-\allowbreak{}v3}}{Tool} & TCP, object/target/initial poses, gripper & 3 & \textbf{1} \texttt{approach\_\allowbreak{}object}; \textbf{2} \texttt{capture\_\allowbreak{}object}; \textbf{3} \texttt{verify\_\allowbreak{}task\_\allowbreak{}outcome} & \textbf{1} \texttt{tcp\_\allowbreak{}to\_\allowbreak{}pregrasp}, \texttt{gripper\_\allowbreak{}opening}; \textbf{2} \texttt{tcp\_\allowbreak{}to\_\allowbreak{}obj}, \texttt{held}, \texttt{capture\_\allowbreak{}progress}; \textbf{3} \texttt{obj\_\allowbreak{}to\_\allowbreak{}target}, \texttt{tcp\_\allowbreak{}to\_\allowbreak{}goal}, \texttt{tcp\_\allowbreak{}to\_\allowbreak{}obj}, \texttt{gripper\_\allowbreak{}closed}, \texttt{held} \\[0.3pt]
\taskwithtype{\texttt{hand-\allowbreak{}insert-\allowbreak{}v3}}{Precise\textsuperscript{LH}} & TCP, object/target/initial poses, gripper, optional orientation & 5 & \textbf{1} \texttt{approach\_\allowbreak{}object}; \textbf{2} \texttt{capture\_\allowbreak{}object}; \textbf{3} \texttt{fixture\_\allowbreak{}clearance}; \textbf{4} \texttt{transport\_\allowbreak{}or\_\allowbreak{}extract}; \textbf{5} \texttt{verify\_\allowbreak{}task\_\allowbreak{}outcome} & \textbf{1} \texttt{tcp\_\allowbreak{}to\_\allowbreak{}pregrasp}, \texttt{gripper\_\allowbreak{}opening}; \textbf{2} \texttt{tcp\_\allowbreak{}to\_\allowbreak{}grasp}, \texttt{held}, \texttt{capture\_\allowbreak{}progress}; \textbf{3} \texttt{tcp\_\allowbreak{}to\_\allowbreak{}vertical}, \texttt{clearance\_\allowbreak{}progress}, \texttt{tcp\_\allowbreak{}to\_\allowbreak{}grasp}, \texttt{held}; \textbf{4} \texttt{tcp\_\allowbreak{}to\_\allowbreak{}transport}, \texttt{object\_\allowbreak{}transport\_\allowbreak{}to\_\allowbreak{}final}, \texttt{clearance\_\allowbreak{}progress}, \texttt{tcp\_\allowbreak{}to\_\allowbreak{}grasp}, \texttt{held}, \texttt{object\_\allowbreak{}to\_\allowbreak{}final}; \textbf{5} \texttt{object\_\allowbreak{}to\_\allowbreak{}final}, \texttt{fine\_\allowbreak{}object\_\allowbreak{}to\_\allowbreak{}final}, \texttt{tcp\_\allowbreak{}to\_\allowbreak{}grasp}, \texttt{held} \\[0.3pt]
\taskwithtype{\texttt{handle-\allowbreak{}press-\allowbreak{}side-\allowbreak{}v3}}{Artic.} & TCP, control/handle, target pose & 3 & \textbf{1} \texttt{approach\_\allowbreak{}handle}; \textbf{2} \texttt{contact\_\allowbreak{}handle}; \textbf{3} \texttt{press\_\allowbreak{}handle} & \textbf{1} \texttt{prepress\_\allowbreak{}distance}; \textbf{2} \texttt{tcp\_\allowbreak{}to\_\allowbreak{}contact\_\allowbreak{}point}; \textbf{3} \texttt{contact\_\allowbreak{}distance}, \texttt{press\_\allowbreak{}or\_\allowbreak{}target\_\allowbreak{}distance}, \texttt{tcp\_\allowbreak{}press\_\allowbreak{}axis\_\allowbreak{}distance} \\[0.3pt]
\taskwithtype{\texttt{handle-\allowbreak{}press-\allowbreak{}v3}}{Artic.} & TCP, control/handle, target pose & 3 & \textbf{1} \texttt{approach\_\allowbreak{}control}; \textbf{2} \texttt{contact\_\allowbreak{}control}; \textbf{3} \texttt{verify\_\allowbreak{}task\_\allowbreak{}outcome} & \textbf{1} \texttt{prepress\_\allowbreak{}distance}; \textbf{2} \texttt{tcp\_\allowbreak{}to\_\allowbreak{}contact\_\allowbreak{}point}; \textbf{3} \texttt{contact\_\allowbreak{}distance}, \texttt{press\_\allowbreak{}or\_\allowbreak{}target\_\allowbreak{}distance}, \texttt{tcp\_\allowbreak{}press\_\allowbreak{}axis\_\allowbreak{}distance} \\[0.3pt]
\taskwithtype{\texttt{handle-\allowbreak{}pull-\allowbreak{}side-\allowbreak{}v3}}{Artic.} & TCP, control/handle, target pose & 3 & \textbf{1} \texttt{approach\_\allowbreak{}control}; \textbf{2} \texttt{contact\_\allowbreak{}control}; \textbf{3} \texttt{verify\_\allowbreak{}task\_\allowbreak{}outcome} & \textbf{1} \texttt{prepress\_\allowbreak{}distance}; \textbf{2} \texttt{tcp\_\allowbreak{}to\_\allowbreak{}contact\_\allowbreak{}point}; \textbf{3} \texttt{contact\_\allowbreak{}distance}, \texttt{press\_\allowbreak{}or\_\allowbreak{}target\_\allowbreak{}distance}, \texttt{tcp\_\allowbreak{}press\_\allowbreak{}axis\_\allowbreak{}distance} \\[0.3pt]
\taskwithtype{\texttt{handle-\allowbreak{}pull-\allowbreak{}v3}}{Artic.} & TCP, control/handle, target pose & 3 & \textbf{1} \texttt{approach\_\allowbreak{}handle}; \textbf{2} \texttt{contact\_\allowbreak{}handle}; \textbf{3} \texttt{pull\_\allowbreak{}handle} & \textbf{1} \texttt{prepress\_\allowbreak{}distance}; \textbf{2} \texttt{tcp\_\allowbreak{}to\_\allowbreak{}contact\_\allowbreak{}point}; \textbf{3} \texttt{contact\_\allowbreak{}distance}, \texttt{press\_\allowbreak{}or\_\allowbreak{}target\_\allowbreak{}distance}, \texttt{tcp\_\allowbreak{}press\_\allowbreak{}axis\_\allowbreak{}distance} \\[0.3pt]
\taskwithtype{\texttt{lever-\allowbreak{}pull-\allowbreak{}v3}}{Artic.} & TCP, control/handle, target pose & 3 & \textbf{1} \texttt{approach\_\allowbreak{}control}; \textbf{2} \texttt{contact\_\allowbreak{}control}; \textbf{3} \texttt{verify\_\allowbreak{}task\_\allowbreak{}outcome} & \textbf{1} \texttt{prepress\_\allowbreak{}distance}; \textbf{2} \texttt{tcp\_\allowbreak{}to\_\allowbreak{}contact\_\allowbreak{}point}; \textbf{3} \texttt{contact\_\allowbreak{}distance}, \texttt{press\_\allowbreak{}or\_\allowbreak{}target\_\allowbreak{}distance}, \texttt{tcp\_\allowbreak{}press\_\allowbreak{}axis\_\allowbreak{}distance} \\[0.3pt]
\taskwithtype{\texttt{peg-\allowbreak{}insert-\allowbreak{}side-\allowbreak{}v3}}{Precise\textsuperscript{LH}} & TCP, object/target/initial poses, gripper, optional orientation & 5 & \textbf{1} \texttt{approach\_\allowbreak{}object}; \textbf{2} \texttt{capture\_\allowbreak{}object}; \textbf{3} \texttt{fixture\_\allowbreak{}clearance}; \textbf{4} \texttt{transport\_\allowbreak{}or\_\allowbreak{}extract}; \textbf{5} \texttt{verify\_\allowbreak{}task\_\allowbreak{}outcome} & \textbf{1} \texttt{tcp\_\allowbreak{}to\_\allowbreak{}pregrasp}, \texttt{gripper\_\allowbreak{}opening}; \textbf{2} \texttt{tcp\_\allowbreak{}to\_\allowbreak{}grasp}, \texttt{held}, \texttt{capture\_\allowbreak{}progress}; \textbf{3} \texttt{tcp\_\allowbreak{}to\_\allowbreak{}vertical}, \texttt{clearance\_\allowbreak{}progress}, \texttt{tcp\_\allowbreak{}to\_\allowbreak{}grasp}, \texttt{held}; \textbf{4} \texttt{tcp\_\allowbreak{}to\_\allowbreak{}transport}, \texttt{object\_\allowbreak{}transport\_\allowbreak{}to\_\allowbreak{}final}, \texttt{clearance\_\allowbreak{}progress}, \texttt{tcp\_\allowbreak{}to\_\allowbreak{}grasp}, \texttt{held}, \texttt{object\_\allowbreak{}to\_\allowbreak{}final}; \textbf{5} \texttt{object\_\allowbreak{}to\_\allowbreak{}final}, \texttt{fine\_\allowbreak{}object\_\allowbreak{}to\_\allowbreak{}final}, \texttt{tcp\_\allowbreak{}to\_\allowbreak{}grasp}, \texttt{held} \\[0.3pt]
\taskwithtype{\texttt{peg-\allowbreak{}unplug-\allowbreak{}side-\allowbreak{}v3}}{Precise\textsuperscript{LH}} & TCP, object/target/initial poses, gripper, optional orientation & 5 & \textbf{1} \texttt{approach\_\allowbreak{}object}; \textbf{2} \texttt{capture\_\allowbreak{}object}; \textbf{3} \texttt{fixture\_\allowbreak{}clearance}; \textbf{4} \texttt{transport\_\allowbreak{}or\_\allowbreak{}extract}; \textbf{5} \texttt{verify\_\allowbreak{}task\_\allowbreak{}outcome} & \textbf{1} \texttt{tcp\_\allowbreak{}to\_\allowbreak{}pregrasp}, \texttt{gripper\_\allowbreak{}opening}; \textbf{2} \texttt{tcp\_\allowbreak{}to\_\allowbreak{}grasp}, \texttt{held}, \texttt{capture\_\allowbreak{}progress}; \textbf{3} \texttt{tcp\_\allowbreak{}to\_\allowbreak{}vertical}, \texttt{clearance\_\allowbreak{}progress}, \texttt{tcp\_\allowbreak{}to\_\allowbreak{}grasp}, \texttt{held}; \textbf{4} \texttt{tcp\_\allowbreak{}to\_\allowbreak{}transport}, \texttt{object\_\allowbreak{}transport\_\allowbreak{}to\_\allowbreak{}final}, \texttt{clearance\_\allowbreak{}progress}, \texttt{tcp\_\allowbreak{}to\_\allowbreak{}grasp}, \texttt{held}, \texttt{object\_\allowbreak{}to\_\allowbreak{}final}; \textbf{5} \texttt{object\_\allowbreak{}to\_\allowbreak{}final}, \texttt{fine\_\allowbreak{}object\_\allowbreak{}to\_\allowbreak{}final}, \texttt{tcp\_\allowbreak{}to\_\allowbreak{}grasp}, \texttt{held} \\[0.3pt]
\taskwithtype{\texttt{pick-\allowbreak{}out-\allowbreak{}of-\allowbreak{}hole-\allowbreak{}v3}}{Precise} & TCP, object/target/initial poses, gripper & 3 & \textbf{1} \texttt{approach\_\allowbreak{}object}; \textbf{2} \texttt{capture\_\allowbreak{}object}; \textbf{3} \texttt{verify\_\allowbreak{}task\_\allowbreak{}outcome} & \textbf{1} \texttt{tcp\_\allowbreak{}to\_\allowbreak{}pregrasp}, \texttt{gripper\_\allowbreak{}opening}; \textbf{2} \texttt{tcp\_\allowbreak{}to\_\allowbreak{}obj}, \texttt{held}, \texttt{capture\_\allowbreak{}progress}; \textbf{3} \texttt{obj\_\allowbreak{}to\_\allowbreak{}target}, \texttt{tcp\_\allowbreak{}to\_\allowbreak{}goal}, \texttt{tcp\_\allowbreak{}to\_\allowbreak{}obj}, \texttt{gripper\_\allowbreak{}closed}, \texttt{held} \\[0.3pt]
\taskwithtype{\texttt{pick-\allowbreak{}place-\allowbreak{}v3}}{Rigid\textsuperscript{LH}} & TCP, object/target/initial poses, gripper, optional orientation & 5 & \textbf{1} \texttt{approach\_\allowbreak{}object}; \textbf{2} \texttt{capture\_\allowbreak{}object}; \textbf{3} \texttt{fixture\_\allowbreak{}clearance}; \textbf{4} \texttt{transport\_\allowbreak{}or\_\allowbreak{}extract}; \textbf{5} \texttt{verify\_\allowbreak{}task\_\allowbreak{}outcome} & \textbf{1} \texttt{tcp\_\allowbreak{}to\_\allowbreak{}pregrasp}, \texttt{gripper\_\allowbreak{}opening}; \textbf{2} \texttt{tcp\_\allowbreak{}to\_\allowbreak{}grasp}, \texttt{held}, \texttt{capture\_\allowbreak{}progress}; \textbf{3} \texttt{tcp\_\allowbreak{}to\_\allowbreak{}vertical}, \texttt{clearance\_\allowbreak{}progress}, \texttt{tcp\_\allowbreak{}to\_\allowbreak{}grasp}, \texttt{held}; \textbf{4} \texttt{tcp\_\allowbreak{}to\_\allowbreak{}transport}, \texttt{object\_\allowbreak{}transport\_\allowbreak{}to\_\allowbreak{}final}, \texttt{clearance\_\allowbreak{}progress}, \texttt{tcp\_\allowbreak{}to\_\allowbreak{}grasp}, \texttt{held}, \texttt{object\_\allowbreak{}to\_\allowbreak{}final}; \textbf{5} \texttt{object\_\allowbreak{}to\_\allowbreak{}final}, \texttt{fine\_\allowbreak{}object\_\allowbreak{}to\_\allowbreak{}final}, \texttt{tcp\_\allowbreak{}to\_\allowbreak{}grasp}, \texttt{held} \\[0.3pt]
\taskwithtype{\texttt{pick-\allowbreak{}place-\allowbreak{}wall-\allowbreak{}v3}}{Precise} & TCP, object/target/initial poses, gripper & 3 & \textbf{1} \texttt{approach\_\allowbreak{}object}; \textbf{2} \texttt{capture\_\allowbreak{}object}; \textbf{3} \texttt{verify\_\allowbreak{}task\_\allowbreak{}outcome} & \textbf{1} \texttt{tcp\_\allowbreak{}to\_\allowbreak{}pregrasp}, \texttt{gripper\_\allowbreak{}opening}; \textbf{2} \texttt{tcp\_\allowbreak{}to\_\allowbreak{}obj}, \texttt{held}, \texttt{capture\_\allowbreak{}progress}; \textbf{3} \texttt{obj\_\allowbreak{}to\_\allowbreak{}target}, \texttt{tcp\_\allowbreak{}to\_\allowbreak{}goal}, \texttt{tcp\_\allowbreak{}to\_\allowbreak{}obj}, \texttt{gripper\_\allowbreak{}closed}, \texttt{held} \\[0.3pt]
\taskwithtype{\texttt{plate-\allowbreak{}slide-\allowbreak{}back-\allowbreak{}side-\allowbreak{}v3}}{Rigid} & TCP, object/target poses, gripper & 3 & \textbf{1} \texttt{approach\_\allowbreak{}object}; \textbf{2} \texttt{contact\_\allowbreak{}object}; \textbf{3} \texttt{verify\_\allowbreak{}task\_\allowbreak{}outcome} & \textbf{1} \texttt{prepush\_\allowbreak{}distance}; \textbf{2} \texttt{contact\_\allowbreak{}distance}, \texttt{tcp\_\allowbreak{}obj\_\allowbreak{}distance}, \texttt{gripper\_\allowbreak{}opening}; \textbf{3} \texttt{obj\_\allowbreak{}target\_\allowbreak{}distance}, \texttt{target\_\allowbreak{}contact\_\allowbreak{}distance}, \texttt{tcp\_\allowbreak{}obj\_\allowbreak{}distance}, \texttt{gripper\_\allowbreak{}opening} \\[0.3pt]
\taskwithtype{\texttt{plate-\allowbreak{}slide-\allowbreak{}back-\allowbreak{}v3}}{Rigid} & TCP, object/target poses, gripper & 3 & \textbf{1} \texttt{approach\_\allowbreak{}object}; \textbf{2} \texttt{contact\_\allowbreak{}object}; \textbf{3} \texttt{verify\_\allowbreak{}task\_\allowbreak{}outcome} & \textbf{1} \texttt{prepush\_\allowbreak{}distance}; \textbf{2} \texttt{contact\_\allowbreak{}distance}, \texttt{tcp\_\allowbreak{}obj\_\allowbreak{}distance}, \texttt{gripper\_\allowbreak{}opening}; \textbf{3} \texttt{obj\_\allowbreak{}target\_\allowbreak{}distance}, \texttt{target\_\allowbreak{}contact\_\allowbreak{}distance}, \texttt{tcp\_\allowbreak{}obj\_\allowbreak{}distance}, \texttt{gripper\_\allowbreak{}opening} \\[0.3pt]
\taskwithtype{\texttt{plate-\allowbreak{}slide-\allowbreak{}side-\allowbreak{}v3}}{Rigid} & TCP, object/target poses, gripper & 3 & \textbf{1} \texttt{approach\_\allowbreak{}object}; \textbf{2} \texttt{contact\_\allowbreak{}object}; \textbf{3} \texttt{verify\_\allowbreak{}task\_\allowbreak{}outcome} & \textbf{1} \texttt{prepush\_\allowbreak{}distance}; \textbf{2} \texttt{contact\_\allowbreak{}distance}, \texttt{tcp\_\allowbreak{}obj\_\allowbreak{}distance}, \texttt{gripper\_\allowbreak{}opening}; \textbf{3} \texttt{obj\_\allowbreak{}target\_\allowbreak{}distance}, \texttt{target\_\allowbreak{}contact\_\allowbreak{}distance}, \texttt{tcp\_\allowbreak{}obj\_\allowbreak{}distance}, \texttt{gripper\_\allowbreak{}opening} \\[0.3pt]
\taskwithtype{\texttt{plate-\allowbreak{}slide-\allowbreak{}v3}}{Rigid} & TCP, object/target poses, gripper & 3 & \textbf{1} \texttt{approach\_\allowbreak{}object}; \textbf{2} \texttt{contact\_\allowbreak{}object}; \textbf{3} \texttt{verify\_\allowbreak{}task\_\allowbreak{}outcome} & \textbf{1} \texttt{prepush\_\allowbreak{}distance}; \textbf{2} \texttt{contact\_\allowbreak{}distance}, \texttt{tcp\_\allowbreak{}obj\_\allowbreak{}distance}, \texttt{gripper\_\allowbreak{}opening}; \textbf{3} \texttt{obj\_\allowbreak{}target\_\allowbreak{}distance}, \texttt{target\_\allowbreak{}contact\_\allowbreak{}distance}, \texttt{tcp\_\allowbreak{}obj\_\allowbreak{}distance}, \texttt{gripper\_\allowbreak{}opening} \\[0.3pt]
\taskwithtype{\texttt{push-\allowbreak{}back-\allowbreak{}v3}}{Rigid} & TCP, object/target poses, gripper & 3 & \textbf{1} \texttt{approach\_\allowbreak{}object}; \textbf{2} \texttt{contact\_\allowbreak{}object}; \textbf{3} \texttt{verify\_\allowbreak{}task\_\allowbreak{}outcome} & \textbf{1} \texttt{prepush\_\allowbreak{}distance}; \textbf{2} \texttt{contact\_\allowbreak{}distance}, \texttt{tcp\_\allowbreak{}obj\_\allowbreak{}distance}, \texttt{gripper\_\allowbreak{}opening}; \textbf{3} \texttt{obj\_\allowbreak{}target\_\allowbreak{}distance}, \texttt{target\_\allowbreak{}contact\_\allowbreak{}distance}, \texttt{tcp\_\allowbreak{}obj\_\allowbreak{}distance}, \texttt{gripper\_\allowbreak{}opening} \\[0.3pt]
\taskwithtype{\texttt{push-\allowbreak{}v3}}{Rigid} & TCP, object/target poses, gripper & 3 & \textbf{1} \texttt{approach\_\allowbreak{}object}; \textbf{2} \texttt{contact\_\allowbreak{}object}; \textbf{3} \texttt{verify\_\allowbreak{}task\_\allowbreak{}outcome} & \textbf{1} \texttt{prepush\_\allowbreak{}distance}; \textbf{2} \texttt{contact\_\allowbreak{}distance}, \texttt{tcp\_\allowbreak{}obj\_\allowbreak{}distance}, \texttt{gripper\_\allowbreak{}opening}; \textbf{3} \texttt{obj\_\allowbreak{}target\_\allowbreak{}distance}, \texttt{target\_\allowbreak{}contact\_\allowbreak{}distance}, \texttt{tcp\_\allowbreak{}obj\_\allowbreak{}distance}, \texttt{gripper\_\allowbreak{}opening} \\[0.3pt]
\taskwithtype{\texttt{push-\allowbreak{}wall-\allowbreak{}v3}}{Rigid\textsuperscript{LH}} & TCP, object/target poses, gripper & 5 & \textbf{1} \texttt{approach\_\allowbreak{}object}; \textbf{2} \texttt{reach\_\allowbreak{}left\_\allowbreak{}clearance}; \textbf{3} \texttt{reacquire\_\allowbreak{}front\_\allowbreak{}push\_\allowbreak{}contact}; \textbf{4} \texttt{clear\_\allowbreak{}wall\_\allowbreak{}front}; \textbf{5} \texttt{verify\_\allowbreak{}task\_\allowbreak{}outcome} & \textbf{1} \texttt{prepush\_\allowbreak{}distance}; \textbf{2} \texttt{waypoint\_\allowbreak{}distance}, \texttt{tcp\_\allowbreak{}obj\_\allowbreak{}distance}; \textbf{3} \texttt{path\_\allowbreak{}contact\_\allowbreak{}distance}, \texttt{tcp\_\allowbreak{}obj\_\allowbreak{}distance}, \texttt{path\_\allowbreak{}progress}; \textbf{4} \texttt{waypoint\_\allowbreak{}progress}, \texttt{path\_\allowbreak{}contact\_\allowbreak{}distance}, \texttt{gripper\_\allowbreak{}opening}; \textbf{5} \texttt{obj\_\allowbreak{}target\_\allowbreak{}distance}, \texttt{target\_\allowbreak{}contact\_\allowbreak{}distance}, \texttt{tcp\_\allowbreak{}obj\_\allowbreak{}distance}, \texttt{gripper\_\allowbreak{}opening} \\[0.3pt]
\taskwithtype{\texttt{reach-\allowbreak{}v3}}{Rigid} & TCP--target distance & 1 & \textbf{1} \texttt{approach\_\allowbreak{}target} & \textbf{1} \texttt{tcp\_\allowbreak{}to\_\allowbreak{}target} \\[0.3pt]
\taskwithtype{\texttt{reach-\allowbreak{}wall-\allowbreak{}v3}}{Rigid} & TCP--target distance & 1 & \textbf{1} \texttt{approach\_\allowbreak{}target} & \textbf{1} \texttt{tcp\_\allowbreak{}to\_\allowbreak{}target} \\[0.3pt]
\taskwithtype{\texttt{shelf-\allowbreak{}place-\allowbreak{}v3}}{Precise\textsuperscript{LH}} & TCP, object/target/initial poses, gripper, optional orientation & 5 & \textbf{1} \texttt{approach\_\allowbreak{}object}; \textbf{2} \texttt{capture\_\allowbreak{}object}; \textbf{3} \texttt{fixture\_\allowbreak{}clearance}; \textbf{4} \texttt{transport\_\allowbreak{}or\_\allowbreak{}extract}; \textbf{5} \texttt{verify\_\allowbreak{}task\_\allowbreak{}outcome} & \textbf{1} \texttt{tcp\_\allowbreak{}to\_\allowbreak{}pregrasp}, \texttt{gripper\_\allowbreak{}opening}; \textbf{2} \texttt{tcp\_\allowbreak{}to\_\allowbreak{}grasp}, \texttt{held}, \texttt{capture\_\allowbreak{}progress}; \textbf{3} \texttt{tcp\_\allowbreak{}to\_\allowbreak{}vertical}, \texttt{clearance\_\allowbreak{}progress}, \texttt{tcp\_\allowbreak{}to\_\allowbreak{}grasp}, \texttt{held}; \textbf{4} \texttt{tcp\_\allowbreak{}to\_\allowbreak{}transport}, \texttt{object\_\allowbreak{}transport\_\allowbreak{}to\_\allowbreak{}final}, \texttt{clearance\_\allowbreak{}progress}, \texttt{tcp\_\allowbreak{}to\_\allowbreak{}grasp}, \texttt{held}, \texttt{object\_\allowbreak{}to\_\allowbreak{}final}; \textbf{5} \texttt{object\_\allowbreak{}to\_\allowbreak{}final}, \texttt{fine\_\allowbreak{}object\_\allowbreak{}to\_\allowbreak{}final}, \texttt{tcp\_\allowbreak{}to\_\allowbreak{}grasp}, \texttt{held} \\[0.3pt]
\taskwithtype{\texttt{soccer-\allowbreak{}v3}}{Rigid} & TCP, object/target poses, gripper & 3 & \textbf{1} \texttt{approach\_\allowbreak{}object}; \textbf{2} \texttt{contact\_\allowbreak{}object}; \textbf{3} \texttt{verify\_\allowbreak{}task\_\allowbreak{}outcome} & \textbf{1} \texttt{prepush\_\allowbreak{}distance}; \textbf{2} \texttt{contact\_\allowbreak{}distance}, \texttt{tcp\_\allowbreak{}obj\_\allowbreak{}distance}, \texttt{gripper\_\allowbreak{}opening}; \textbf{3} \texttt{obj\_\allowbreak{}target\_\allowbreak{}distance}, \texttt{target\_\allowbreak{}contact\_\allowbreak{}distance}, \texttt{tcp\_\allowbreak{}obj\_\allowbreak{}distance}, \texttt{gripper\_\allowbreak{}opening} \\[0.3pt]
\taskwithtype{\texttt{stick-\allowbreak{}pull-\allowbreak{}v3}}{Tool} & TCP, tool/secondary/target poses, gripper, optional orientation & 4 & \textbf{1} \texttt{approach\_\allowbreak{}tool}; \textbf{2} \texttt{capture\_\allowbreak{}and\_\allowbreak{}lift\_\allowbreak{}tool}; \textbf{3} \texttt{tool\_\allowbreak{}endpoint\_\allowbreak{}at\_\allowbreak{}secondary\_\allowbreak{}object}; \textbf{4} \texttt{secondary\_\allowbreak{}object\_\allowbreak{}at\_\allowbreak{}goal} & \textbf{1} \texttt{tcp\_\allowbreak{}to\_\allowbreak{}pregrasp}, \texttt{gripper\_\allowbreak{}opening}; \textbf{2} \texttt{tcp\_\allowbreak{}to\_\allowbreak{}grasp}, \texttt{held}, \texttt{tool\_\allowbreak{}lift}; \textbf{3} \texttt{effect\_\allowbreak{}to\_\allowbreak{}secondary}, \texttt{alignment\_\allowbreak{}distance}, \texttt{gripper\_\allowbreak{}closed}, \texttt{held}, \texttt{tcp\_\allowbreak{}to\_\allowbreak{}grasp}; \textbf{4} \texttt{secondary\_\allowbreak{}to\_\allowbreak{}target}, \texttt{effect\_\allowbreak{}to\_\allowbreak{}secondary}, \texttt{held} \\[0.3pt]
\taskwithtype{\texttt{stick-\allowbreak{}push-\allowbreak{}v3}}{Tool} & TCP, tool/secondary/target poses, gripper, optional orientation & 4 & \textbf{1} \texttt{approach\_\allowbreak{}tool}; \textbf{2} \texttt{capture\_\allowbreak{}and\_\allowbreak{}lift\_\allowbreak{}tool}; \textbf{3} \texttt{tool\_\allowbreak{}endpoint\_\allowbreak{}at\_\allowbreak{}secondary\_\allowbreak{}object}; \textbf{4} \texttt{secondary\_\allowbreak{}object\_\allowbreak{}at\_\allowbreak{}goal} & \textbf{1} \texttt{tcp\_\allowbreak{}to\_\allowbreak{}pregrasp}, \texttt{gripper\_\allowbreak{}opening}; \textbf{2} \texttt{tcp\_\allowbreak{}to\_\allowbreak{}grasp}, \texttt{held}, \texttt{tool\_\allowbreak{}lift}; \textbf{3} \texttt{effect\_\allowbreak{}to\_\allowbreak{}secondary}, \texttt{alignment\_\allowbreak{}distance}, \texttt{gripper\_\allowbreak{}closed}, \texttt{held}, \texttt{tcp\_\allowbreak{}to\_\allowbreak{}grasp}; \textbf{4} \texttt{secondary\_\allowbreak{}to\_\allowbreak{}target}, \texttt{effect\_\allowbreak{}to\_\allowbreak{}secondary}, \texttt{held} \\[0.3pt]
\taskwithtype{\texttt{sweep-\allowbreak{}into-\allowbreak{}v3}}{Rigid} & TCP, object/target poses, gripper & 3 & \textbf{1} \texttt{approach\_\allowbreak{}object}; \textbf{2} \texttt{contact\_\allowbreak{}object}; \textbf{3} \texttt{verify\_\allowbreak{}task\_\allowbreak{}outcome} & \textbf{1} \texttt{prepush\_\allowbreak{}distance}; \textbf{2} \texttt{contact\_\allowbreak{}distance}, \texttt{tcp\_\allowbreak{}obj\_\allowbreak{}distance}, \texttt{gripper\_\allowbreak{}opening}; \textbf{3} \texttt{obj\_\allowbreak{}target\_\allowbreak{}distance}, \texttt{target\_\allowbreak{}contact\_\allowbreak{}distance}, \texttt{tcp\_\allowbreak{}obj\_\allowbreak{}distance}, \texttt{gripper\_\allowbreak{}opening} \\[0.3pt]
\taskwithtype{\texttt{sweep-\allowbreak{}v3}}{Rigid} & TCP, object/target poses, gripper & 3 & \textbf{1} \texttt{approach\_\allowbreak{}object}; \textbf{2} \texttt{contact\_\allowbreak{}object}; \textbf{3} \texttt{verify\_\allowbreak{}task\_\allowbreak{}outcome} & \textbf{1} \texttt{prepush\_\allowbreak{}distance}; \textbf{2} \texttt{contact\_\allowbreak{}distance}, \texttt{tcp\_\allowbreak{}obj\_\allowbreak{}distance}, \texttt{gripper\_\allowbreak{}opening}; \textbf{3} \texttt{obj\_\allowbreak{}target\_\allowbreak{}distance}, \texttt{target\_\allowbreak{}contact\_\allowbreak{}distance}, \texttt{tcp\_\allowbreak{}obj\_\allowbreak{}distance}, \texttt{gripper\_\allowbreak{}opening} \\[0.3pt]
\taskwithtype{\texttt{window-\allowbreak{}close-\allowbreak{}v3}}{Artic.} & TCP, control/handle, target pose & 3 & \textbf{1} \texttt{approach\_\allowbreak{}control}; \textbf{2} \texttt{contact\_\allowbreak{}control}; \textbf{3} \texttt{verify\_\allowbreak{}task\_\allowbreak{}outcome} & \textbf{1} \texttt{prepress\_\allowbreak{}distance}; \textbf{2} \texttt{tcp\_\allowbreak{}to\_\allowbreak{}contact\_\allowbreak{}point}; \textbf{3} \texttt{contact\_\allowbreak{}distance}, \texttt{press\_\allowbreak{}or\_\allowbreak{}target\_\allowbreak{}distance}, \texttt{tcp\_\allowbreak{}press\_\allowbreak{}axis\_\allowbreak{}distance} \\[0.3pt]
\taskwithtype{\texttt{window-\allowbreak{}open-\allowbreak{}v3}}{Artic.} & TCP, control/handle, target pose & 3 & \textbf{1} \texttt{approach\_\allowbreak{}control}; \textbf{2} \texttt{contact\_\allowbreak{}control}; \textbf{3} \texttt{verify\_\allowbreak{}task\_\allowbreak{}outcome} & \textbf{1} \texttt{prepress\_\allowbreak{}distance}; \textbf{2} \texttt{tcp\_\allowbreak{}to\_\allowbreak{}contact\_\allowbreak{}point}; \textbf{3} \texttt{contact\_\allowbreak{}distance}, \texttt{press\_\allowbreak{}or\_\allowbreak{}target\_\allowbreak{}distance}, \texttt{tcp\_\allowbreak{}press\_\allowbreak{}axis\_\allowbreak{}distance} \\[0.3pt]
\addlinespace[1.5pt]
\multicolumn{5}{@{}l}{\textbf{ManiSkill}} \\
\midrule
\taskwithtype{\texttt{Pick\allowbreak{}Cube-\allowbreak{}v1}}{Rigid} & TCP, cube/goal poses, grasp & 3 & \textbf{1} \texttt{reach\_\allowbreak{}object}; \textbf{2} \texttt{establish\_\allowbreak{}grasp}; \textbf{3} \texttt{transport\_\allowbreak{}and\_\allowbreak{}settle} & \textbf{1} \texttt{tcp\_\allowbreak{}obj}; \textbf{2} \texttt{grasped}; \textbf{3} \texttt{obj\_\allowbreak{}goal} \\[0.3pt]
\taskwithtype{\texttt{Push\allowbreak{}Cube-\allowbreak{}v1}}{Rigid} & TCP, cube/goal poses & 3 & \textbf{1} \texttt{reach\_\allowbreak{}pre\_\allowbreak{}push\_\allowbreak{}pose}; \textbf{2} \texttt{establish\_\allowbreak{}contact}; \textbf{3} \texttt{push\_\allowbreak{}to\_\allowbreak{}goal} & \textbf{1} \texttt{tcp\_\allowbreak{}push}; \textbf{2} \texttt{tcp\_\allowbreak{}push}; \textbf{3} \texttt{obj\_\allowbreak{}goal} \\[0.3pt]
\taskwithtype{\texttt{Pull\allowbreak{}Cube-\allowbreak{}v1}}{Rigid} & TCP, cube/goal poses & 3 & \textbf{1} \texttt{reach\_\allowbreak{}pull\_\allowbreak{}pose}; \textbf{2} \texttt{establish\_\allowbreak{}contact}; \textbf{3} \texttt{pull\_\allowbreak{}to\_\allowbreak{}goal} & \textbf{1} \texttt{tcp\_\allowbreak{}pull}; \textbf{2} \texttt{tcp\_\allowbreak{}pull}; \textbf{3} \texttt{obj\_\allowbreak{}goal} \\[0.3pt]
\taskwithtype{\texttt{Lift\allowbreak{}Peg\allowbreak{}Upright-\allowbreak{}v1}}{Precise} & TCP, peg pose, grasp, upright/height & 4 & \textbf{1} \texttt{reach\_\allowbreak{}peg}; \textbf{2} \texttt{grasp\_\allowbreak{}peg}; \textbf{3} \texttt{reorient\_\allowbreak{}upright}; \textbf{4} \texttt{settle} & \textbf{1} \texttt{tcp\_\allowbreak{}peg}; \textbf{2} \texttt{grasped}; \textbf{3} \texttt{upright}; \textbf{4} \texttt{upright}, \texttt{z\_\allowbreak{}error} \\[0.3pt]
\taskwithtype{\texttt{Peg\allowbreak{}Insertion\allowbreak{}Side-\allowbreak{}v1}}{Precise} & TCP, peg/hole poses, grasp & 4 & \textbf{1} \texttt{reach\_\allowbreak{}peg}; \textbf{2} \texttt{grasp\_\allowbreak{}peg}; \textbf{3} \texttt{align\_\allowbreak{}with\_\allowbreak{}hole}; \textbf{4} \texttt{insert} & \textbf{1} \texttt{tcp\_\allowbreak{}peg}; \textbf{2} \texttt{grasped}; \textbf{3} \texttt{align\_\allowbreak{}error}; \textbf{4} \texttt{hole\_\allowbreak{}error} \\[0.3pt]
\taskwithtype{\texttt{Plug\allowbreak{}Charger-\allowbreak{}v1}}{Precise\textsuperscript{LH}} & TCP, charger/goal poses, grasp, orientation & 5 & \textbf{1} \texttt{reach\_\allowbreak{}charger}; \textbf{2} \texttt{grasp\_\allowbreak{}charger}; \textbf{3} \texttt{orient}; \textbf{4} \texttt{align\_\allowbreak{}with\_\allowbreak{}receptacle}; \textbf{5} \texttt{insert} & \textbf{1} \texttt{tcp\_\allowbreak{}obj}; \textbf{2} \texttt{grasped}; \textbf{3} \texttt{orientation\_\allowbreak{}error}; \textbf{4} \texttt{obj\_\allowbreak{}goal}, \texttt{orientation\_\allowbreak{}error}; \textbf{5} \texttt{obj\_\allowbreak{}goal} \\[0.3pt]
\taskwithtype{\texttt{Poke\allowbreak{}Cube-\allowbreak{}v1}}{Tool} & TCP, peg/cube/goal poses, grasp, orientation & 4 & \textbf{1} \texttt{reach\_\allowbreak{}tool}; \textbf{2} \texttt{grasp\_\allowbreak{}tool}; \textbf{3} \texttt{align\_\allowbreak{}tool\_\allowbreak{}with\_\allowbreak{}cube}; \textbf{4} \texttt{poke\_\allowbreak{}to\_\allowbreak{}goal} & \textbf{1} \texttt{tcp\_\allowbreak{}peg}; \textbf{2} \texttt{grasped}; \textbf{3} \texttt{peg\_\allowbreak{}angle}, \texttt{head\_\allowbreak{}cube\_\allowbreak{}distance}; \textbf{4} \texttt{cube\_\allowbreak{}goal} \\[0.3pt]
\taskwithtype{\texttt{Pull\allowbreak{}Cube\allowbreak{}Tool-\allowbreak{}v1}}{Tool} & TCP, tool/cube poses, grasp, hook geometry & 4 & \textbf{1} \texttt{reach\_\allowbreak{}tool}; \textbf{2} \texttt{grasp\_\allowbreak{}tool}; \textbf{3} \texttt{hook\_\allowbreak{}cube}; \textbf{4} \texttt{pull\_\allowbreak{}to\_\allowbreak{}goal} & \textbf{1} \texttt{tcp\_\allowbreak{}tool}; \textbf{2} \texttt{grasped}; \textbf{3} \texttt{hook\_\allowbreak{}error}; \textbf{4} \texttt{cube\_\allowbreak{}base\_\allowbreak{}distance} \\[0.3pt]
\taskwithtype{\texttt{Push\allowbreak{}T-\allowbreak{}v1}}{Precise} & TCP, T/goal poses, orientation & 4 & \textbf{1} \texttt{reach\_\allowbreak{}contact\_\allowbreak{}pose}; \textbf{2} \texttt{translate\_\allowbreak{}object}; \textbf{3} \texttt{rotate\_\allowbreak{}and\_\allowbreak{}align}; \textbf{4} \texttt{settle} & \textbf{1} \texttt{tcp\_\allowbreak{}obj}; \textbf{2} \texttt{xy\_\allowbreak{}error}; \textbf{3} \texttt{angle\_\allowbreak{}error}; \textbf{4} \texttt{xy\_\allowbreak{}error}, \texttt{angle\_\allowbreak{}error} \\[0.3pt]
\taskwithtype{\texttt{Roll\allowbreak{}Ball-\allowbreak{}v1}}{Rigid} & TCP, ball/goal pose, ball velocity & 4 & \textbf{1} \texttt{reach\_\allowbreak{}ball}; \textbf{2} \texttt{establish\_\allowbreak{}contact}; \textbf{3} \texttt{roll\_\allowbreak{}to\_\allowbreak{}goal}; \textbf{4} \texttt{settle} & \textbf{1} \texttt{tcp\_\allowbreak{}hit}; \textbf{2} \texttt{tcp\_\allowbreak{}ball}; \textbf{3} \texttt{ball\_\allowbreak{}goal}; \textbf{4} \texttt{ball\_\allowbreak{}speed} \\[0.3pt]
\taskwithtype{\texttt{Stack\allowbreak{}Cube-\allowbreak{}v1}}{Rigid\textsuperscript{LH}} & TCP, cube poses, grasp, stack geometry, success & 5 & \textbf{1} \texttt{reach\_\allowbreak{}cube}; \textbf{2} \texttt{grasp\_\allowbreak{}cube}; \textbf{3} \texttt{align\_\allowbreak{}above\_\allowbreak{}base}; \textbf{4} \texttt{place}; \textbf{5} \texttt{release} & \textbf{1} \texttt{tcp\_\allowbreak{}obj}; \textbf{2} \texttt{grasped}; \textbf{3} \texttt{goal\_\allowbreak{}error}; \textbf{4} \texttt{goal\_\allowbreak{}error}; \textbf{5} \texttt{success} \\[0.3pt]
\taskwithtype{\texttt{Stack\allowbreak{}Pyramid-\allowbreak{}v1}}{Precise\textsuperscript{LH}} & TCP, three cube poses, grasp/static flags, success & 7 & \textbf{1} \texttt{grasp\_\allowbreak{}red}; \textbf{2} \texttt{place\_\allowbreak{}red\_\allowbreak{}beside\_\allowbreak{}green}; \textbf{3} \texttt{release\_\allowbreak{}red}; \textbf{4} \texttt{grasp\_\allowbreak{}blue}; \textbf{5} \texttt{align\_\allowbreak{}blue}; \textbf{6} \texttt{stack\_\allowbreak{}blue}; \textbf{7} \texttt{release\_\allowbreak{}blue} & \textbf{1} \texttt{tcp\_\allowbreak{}red}, \texttt{grasp\_\allowbreak{}red}; \textbf{2} \texttt{red\_\allowbreak{}green\_\allowbreak{}distance}; \textbf{3} \texttt{beside}, \texttt{release\_\allowbreak{}red}; \textbf{4} \texttt{tcp\_\allowbreak{}blue}, \texttt{grasp\_\allowbreak{}blue}; \textbf{5} \texttt{blue\_\allowbreak{}goal}; \textbf{6} \texttt{stacked}; \textbf{7} \texttt{success} \\[0.3pt]
\taskwithtype{\texttt{Pick\allowbreak{}Single\allowbreak{}YCB-\allowbreak{}v1}}{Rigid} & TCP, object/goal poses, grasp, placed/static & 4 & \textbf{1} \texttt{reach\_\allowbreak{}object}; \textbf{2} \texttt{establish\_\allowbreak{}grasp}; \textbf{3} \texttt{transport\_\allowbreak{}to\_\allowbreak{}goal}; \textbf{4} \texttt{settle} & \textbf{1} \texttt{tcp\_\allowbreak{}obj}; \textbf{2} \texttt{grasped}; \textbf{3} \texttt{goal}; \textbf{4} \texttt{placed}, \texttt{static} \\[0.3pt]
\taskwithtype{\texttt{Place\allowbreak{}Sphere-\allowbreak{}v1}}{Rigid} & TCP, sphere/bin poses, grasp & 4 & \textbf{1} \texttt{reach\_\allowbreak{}sphere}; \textbf{2} \texttt{grasp\_\allowbreak{}sphere}; \textbf{3} \texttt{align\_\allowbreak{}with\_\allowbreak{}bin}; \textbf{4} \texttt{place\_\allowbreak{}and\_\allowbreak{}release} & \textbf{1} \texttt{tcp\_\allowbreak{}obj}; \textbf{2} \texttt{grasped}; \textbf{3} \texttt{goal}; \textbf{4} \texttt{on\_\allowbreak{}bin}, \texttt{released} \\[0.3pt]
\taskwithtype{\texttt{FMBAssembly1\allowbreak{}Easy-\allowbreak{}v1}}{Precise\textsuperscript{LH}} & TCP, bridge/fixture/goal poses, grasp, orientation & 6 & \textbf{1} \texttt{reach\_\allowbreak{}bridge}; \textbf{2} \texttt{grasp\_\allowbreak{}bridge}; \textbf{3} \texttt{reorient\_\allowbreak{}with\_\allowbreak{}fixture}; \textbf{4} \texttt{regrasp\_\allowbreak{}bridge}; \textbf{5} \texttt{align\_\allowbreak{}with\_\allowbreak{}board}; \textbf{6} \texttt{place\_\allowbreak{}bridge} & \textbf{1} \texttt{tcp\_\allowbreak{}bridge}; \textbf{2} \texttt{grasped}; \textbf{3} \texttt{fixture\_\allowbreak{}error}, \texttt{goal\_\allowbreak{}angle}; \textbf{4} \texttt{grasped}; \textbf{5} \texttt{goal\_\allowbreak{}pos}, \texttt{goal\_\allowbreak{}angle}; \textbf{6} \texttt{goal\_\allowbreak{}pos} \\[0.3pt]
\taskwithtype{\texttt{Rotate\allowbreak{}Valve\allowbreak{}Level4-\allowbreak{}v1}}{Artic.} & finger tips, valve pose/qpos/qvel, turn direction, success & 4 & \textbf{1} \texttt{reach\_\allowbreak{}valve\_\allowbreak{}spokes}; \textbf{2} \texttt{establish\_\allowbreak{}turning\_\allowbreak{}motion}; \textbf{3} \texttt{rotate\_\allowbreak{}full\_\allowbreak{}turn}; \textbf{4} \texttt{settle\_\allowbreak{}success} & \textbf{1} \texttt{contact\_\allowbreak{}error}; \textbf{2} \texttt{contact\_\allowbreak{}error}, \texttt{positive\_\allowbreak{}motion}; \textbf{3} \texttt{rotation\_\allowbreak{}progress}; \textbf{4} \texttt{success}, \texttt{valve\_\allowbreak{}speed} \\[0.3pt]
\taskwithtype{\texttt{Two\allowbreak{}Robot\allowbreak{}Stack\allowbreak{}Cube-\allowbreak{}v1}}{Precise\textsuperscript{LH}} & robot TCPs, cube/goal poses, grasp, success & 5 & \textbf{1} \texttt{reach\_\allowbreak{}assigned\_\allowbreak{}cubes}; \textbf{2} \texttt{grasp\_\allowbreak{}top\_\allowbreak{}cube}; \textbf{3} \texttt{place\_\allowbreak{}bottom\_\allowbreak{}cube}; \textbf{4} \texttt{stack\_\allowbreak{}top\_\allowbreak{}cube}; \textbf{5} \texttt{release\_\allowbreak{}both} & \textbf{1} \texttt{reach}; \textbf{2} \texttt{cube\allowbreak{}A\_\allowbreak{}grasped}; \textbf{3} \texttt{cube\allowbreak{}B\_\allowbreak{}goal}, \texttt{cube\allowbreak{}A\_\allowbreak{}grasped}; \textbf{4} \texttt{cube\allowbreak{}A\_\allowbreak{}top\_\allowbreak{}error}; \textbf{5} \texttt{success} \\[0.3pt]
\taskwithtype{\texttt{Draw\allowbreak{}Triangle-\allowbreak{}v1}}{Tool} & TCP, canvas/vertices, path coverage, success & 4 & \textbf{1} \texttt{reach\_\allowbreak{}canvas\_\allowbreak{}start}; \textbf{2} \texttt{trace\_\allowbreak{}first\_\allowbreak{}side}; \textbf{3} \texttt{trace\_\allowbreak{}remaining\_\allowbreak{}sides}; \textbf{4} \texttt{complete\_\allowbreak{}triangle} & \textbf{1} \texttt{reach}; \textbf{2} \texttt{path\_\allowbreak{}coverage}; \textbf{3} \texttt{path\_\allowbreak{}coverage}; \textbf{4} \texttt{covered} \\[0.3pt]
\taskwithtype{\texttt{Tri\allowbreak{}Finger\allowbreak{}Rotate\allowbreak{}Cube\allowbreak{}Level4-\allowbreak{}v1}}{Precise} & finger tips, cube/goal pose, position/orientation error, success & 4 & \textbf{1} \texttt{reach\_\allowbreak{}cube\_\allowbreak{}with\_\allowbreak{}fingers}; \textbf{2} \texttt{lift\_\allowbreak{}and\_\allowbreak{}translate\_\allowbreak{}cube}; \textbf{3} \texttt{match\_\allowbreak{}goal\_\allowbreak{}orientation}; \textbf{4} \texttt{settle\_\allowbreak{}at\_\allowbreak{}goal} & \textbf{1} \texttt{tip\_\allowbreak{}to\_\allowbreak{}cube}; \textbf{2} \texttt{position\_\allowbreak{}progress}; \textbf{3} \texttt{pose\_\allowbreak{}progress}; \textbf{4} \texttt{success} \\[0.3pt]
\addlinespace[1.5pt]
\multicolumn{5}{@{}l}{\textbf{SoftGym}} \\
\midrule
\taskwithtype{\texttt{Cloth\allowbreak{}Drop}}{Deform.} & particles, cloth size/radius, target layout & 2 & \textbf{1} \texttt{approach\_\allowbreak{}release}; \textbf{2} \texttt{flat\_\allowbreak{}completion} & \textbf{1} \texttt{target\_\allowbreak{}progress}, \texttt{target\_\allowbreak{}alignment}, \texttt{low\_\allowbreak{}height}, \texttt{spread}; \textbf{2} \texttt{target\_\allowbreak{}progress}, \texttt{target\_\allowbreak{}alignment}, \texttt{flatness}, \texttt{spread} \\[0.3pt]
\taskwithtype{\texttt{Cloth\allowbreak{}Flatten}}{Deform.} & particles, cloth size/radius & 2 & \textbf{1} \texttt{spread\_\allowbreak{}progress}; \textbf{2} \texttt{flat\_\allowbreak{}completion} & \textbf{1} \texttt{spread}, \texttt{flatness}; \textbf{2} \texttt{spread}, \texttt{flatness} \\[0.3pt]
\taskwithtype{\texttt{Cloth\allowbreak{}Fold}}{Deform.} & particles, fold groups, initial group distance & 2 & \textbf{1} \texttt{fold\_\allowbreak{}progress}; \textbf{2} \texttt{fold\_\allowbreak{}completion} & \textbf{1} \texttt{coverage\_\allowbreak{}A\_\allowbreak{}to\_\allowbreak{}B}, \texttt{coverage\_\allowbreak{}B\_\allowbreak{}to\_\allowbreak{}A}; \textbf{2} \texttt{coverage\_\allowbreak{}A\_\allowbreak{}to\_\allowbreak{}B}, \texttt{coverage\_\allowbreak{}B\_\allowbreak{}to\_\allowbreak{}A} \\[0.3pt]
\taskwithtype{\texttt{Rope\allowbreak{}Flatten}}{Deform.} & ordered particles, rope length & 2 & \textbf{1} \texttt{stretch\_\allowbreak{}progress}; \textbf{2} \texttt{straight\_\allowbreak{}completion} & \textbf{1} \texttt{extension}, \texttt{straightness}; \textbf{2} \texttt{straightness}, \texttt{extension}, \texttt{on\_\allowbreak{}plane} \\[0.3pt]
\taskwithtype{\texttt{Pass\allowbreak{}Water}}{Fluid} & cup/target x, water particles, in-glass test, glass geometry & 2 & \textbf{1} \texttt{transport\_\allowbreak{}progress}; \textbf{2} \texttt{water\_\allowbreak{}delivery} & \textbf{1} \texttt{transport\_\allowbreak{}progress}, \texttt{retained\_\allowbreak{}water}; \textbf{2} \texttt{target\_\allowbreak{}closeness}, \texttt{spill\_\allowbreak{}penalty} \\[0.3pt]
\taskwithtype{\texttt{Pour\allowbreak{}Water}}{Fluid} & source/target cup geometry, water particles, in-glass test & 2 & \textbf{1} \texttt{pour\_\allowbreak{}progress}; \textbf{2} \texttt{target\_\allowbreak{}fill\_\allowbreak{}completion} & \textbf{1} \texttt{cup\_\allowbreak{}alignment}, \texttt{tilt}, \texttt{early\_\allowbreak{}transfer}, \texttt{retained\_\allowbreak{}water}; \textbf{2} \texttt{target\_\allowbreak{}fill}, \texttt{retained\_\allowbreak{}water}, \texttt{source\_\allowbreak{}emptying} \\[0.3pt]
\end{xltabular}
\endgroup

\stopcontents[appendix]

\end{document}